\documentclass[preprint,review,12pt]{elsarticle}

\usepackage{amssymb}

\usepackage[breaklinks]{hyperref}
\usepackage[dvipsnames]{xcolor}
\usepackage{xcolor}
\usepackage{epsfig}
\usepackage{graphicx}
\usepackage{amsmath}
\usepackage[nodisplayskipstretch]{setspace}
\usepackage{amssymb}
\usepackage{multirow}
\usepackage{adjustbox}
\usepackage{csquotes}
\usepackage{subcaption}
\usepackage{comment,balance}
\usepackage{array, caption, tabularx,  ragged2e,  booktabs}
\usepackage{stackengine}
\usepackage{mathtools}
\usepackage{duckuments}
\usepackage{setspace}
\usepackage{bbm}
\usepackage[bb=dsserif]{mathalpha}
\usepackage{algpseudocode}
\usepackage{algorithm}
\usepackage{rotating}
\usepackage{makecell}
\usepackage{placeins}
\usepackage{afterpage}
\usepackage{anyfontsize}
\algnewcommand\And{\textbf{and} }
\newcommand{\mycomment}[1]{}
\newcolumntype{?}[1]{!{\vrule width #1}}

\journal{Signal Processing: Image Communication}

\begin{document}
\begin{frontmatter}



\title{Text in the Dark: Extremely Low-Light Text Image Enhancement}
\author[chalmers]{Che-Tsung Lin\fnref{equal}}
\author[um]{Chun Chet Ng\fnref{equal}}
\author[um]{Zhi Qin Tan}
\author[um]{Wan Jun Nah}
\author[adelaide]{Xinyu Wang}
\author[um]{Jie Long Kew}
\author[nthu]{Pohao Hsu}
\author[nthu]{Shang Hong Lai}
\author[um]{Chee Seng Chan\corref{cor1}}
\ead{cs.chan@um.edu.my}
\author[chalmers]{Christopher Zach}

\fntext[equal]{These authors contributed equally to this work.}
\cortext[cor1]{Corresponding Author.}

\affiliation[chalmers]{organization={Chalmers University of Technology},
            state={Gothenburg},
            country={Sweden}}
            
\affiliation[um]{organization={Universiti Malaya},
            state={Kuala Lumpur},
            country={Malaysia}}
            
\affiliation[nthu]{organization={National Tsing Hua University},
            state={Hsinchu},
            country={Taiwan}}

\affiliation[adelaide]{organization={The University of Adelaide},
            state={Adelaide},
            country={Australia}}


\begin{abstract}
Text extraction in extremely low-light images is challenging. Although existing low-light image enhancement methods can enhance images as pre-processing before text extraction, they do not focus on scene text. Further research is also hindered by the lack of extremely low-light text datasets. Thus, we propose a novel extremely low-light image enhancement framework with an edge-aware attention module to focus on scene text regions. Our method is trained with text detection and edge reconstruction losses to emphasize low-level scene text features. Additionally, we present a Supervised Deep Curve Estimation model to synthesize extremely low-light images based on the public ICDAR15 (IC15) dataset. We also labeled texts in the extremely low-light See In the Dark (SID) and ordinary LOw-Light (LOL) datasets to benchmark extremely low-light scene text tasks. Extensive experiments prove our model outperforms state-of-the-art methods on all datasets. Code and dataset will be released publicly at \href{https://github.com/chunchet-ng/Text-in-the-Dark}{https://github.com/chunchet-ng/Text-in-the-Dark}.
\end{abstract}

\begin{highlights}
\item We present a new method to enhance low-light images, especially scene text regions.

\item We developed a novel Supervised-DCE model to synthesize extremely low-light images.

\item We create 3 new low-light text datasets SID-Sony-Text, SID-Fuji-Text, and LOL-Text.

\item Our new datasets assess enhanced low-light images with scene text extraction tasks.

\item Our method achieves the best results on all datasets quantitatively \& qualitatively.
\end{highlights}

\end{frontmatter}

\setlength{\abovedisplayskip}{7pt}
\setlength{\belowdisplayskip}{7pt}

\section{Introduction}
\label{sec:intro}

\begin{figure*}[!ht]
\centering
\begin{subfigure}{.24\linewidth}
    \centering
    \includegraphics[keepaspectratio=true, width=\textwidth]{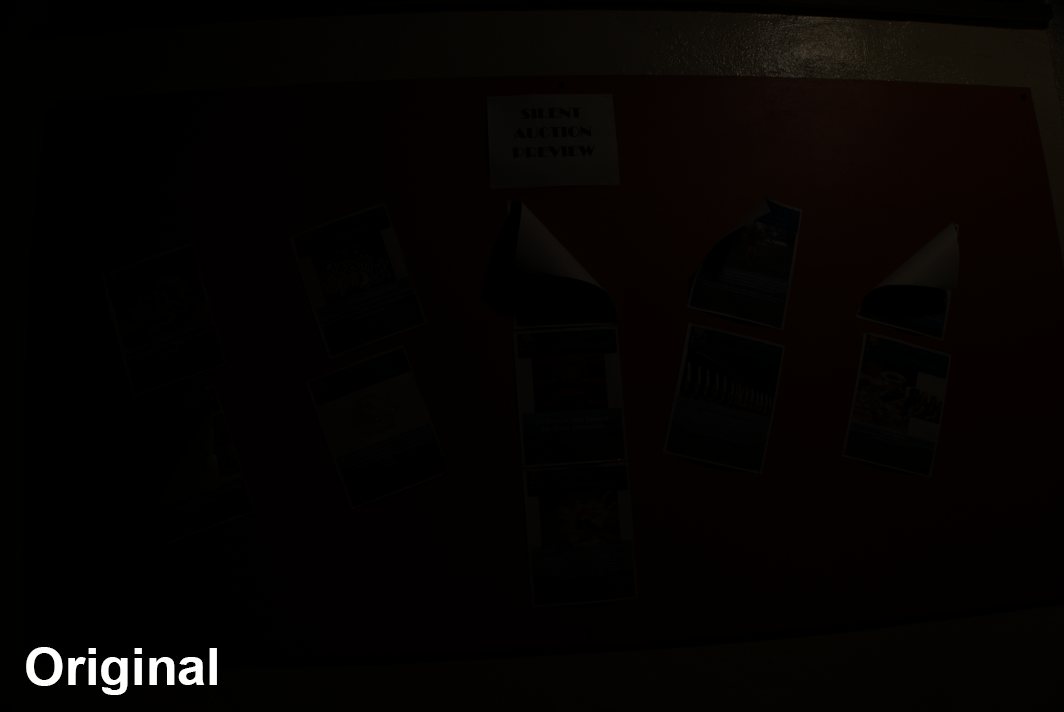}
    \includegraphics[keepaspectratio=true, width=\textwidth]{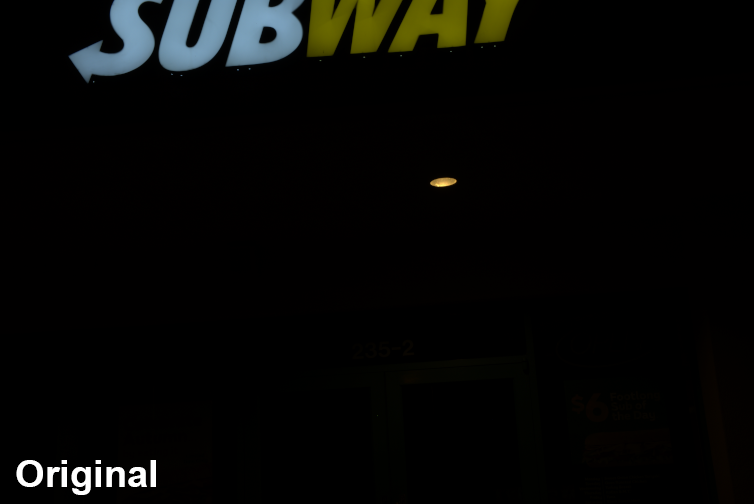}
    \includegraphics[height=2cm, width=\textwidth]
    {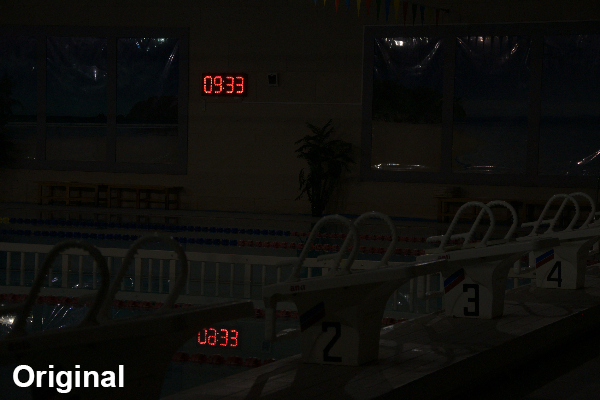}
    \caption{}
\end{subfigure}
\begin{subfigure}{.24\linewidth}
    \centering
    \includegraphics[keepaspectratio=true, width=\textwidth]{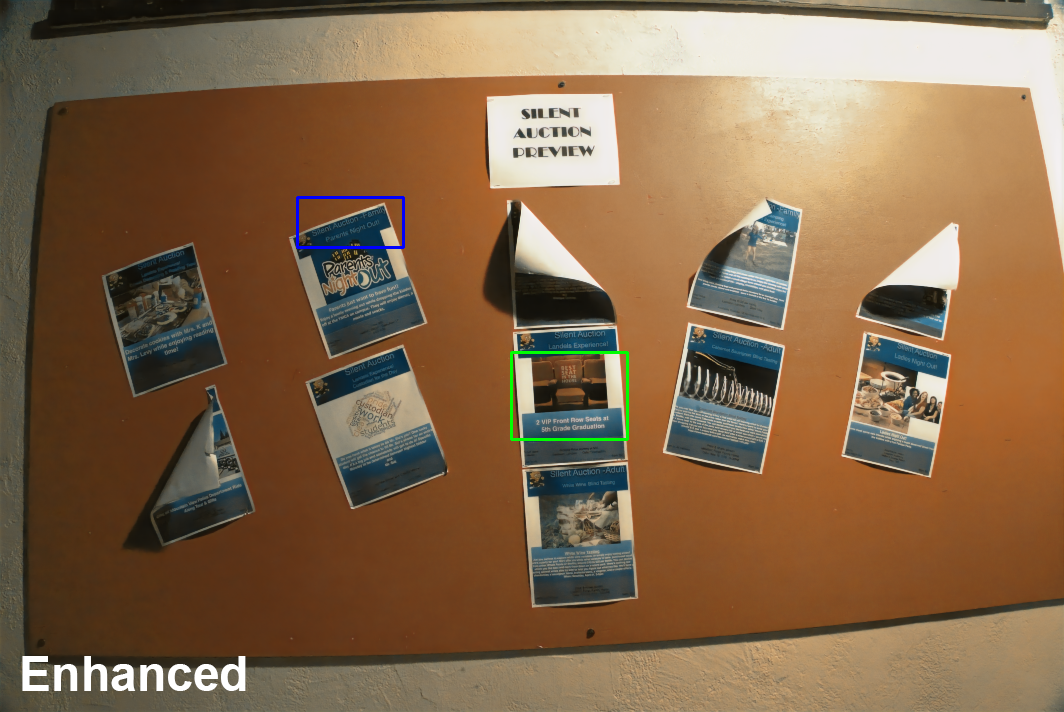}    
    \includegraphics[keepaspectratio=true, width=\textwidth]{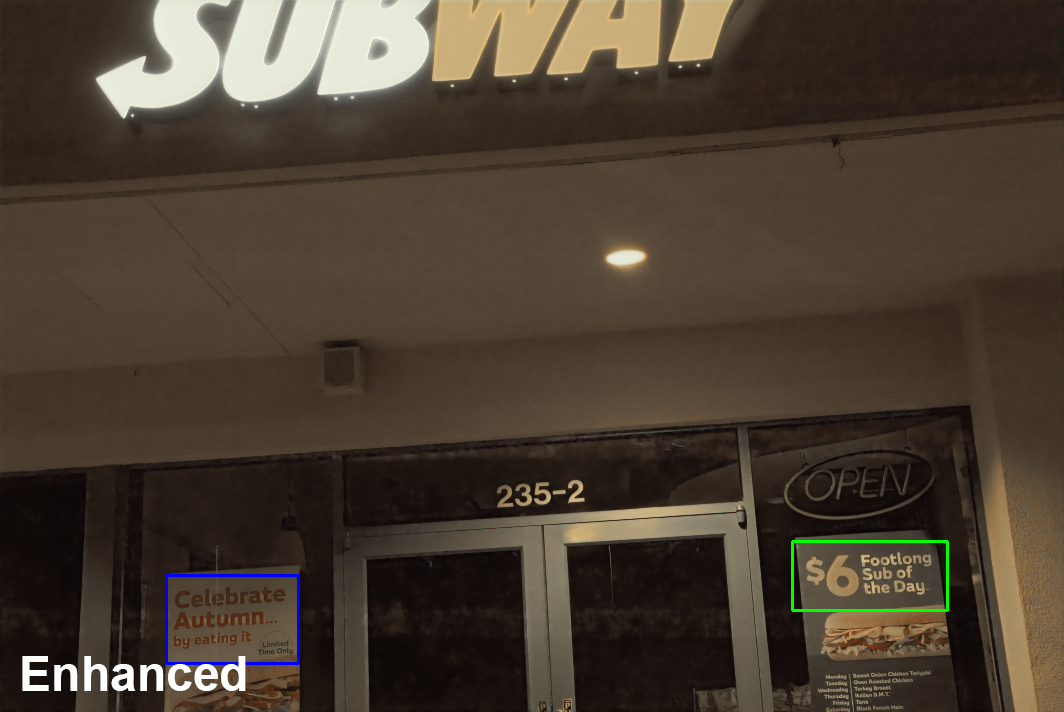}
    \includegraphics[height=2cm, width=\textwidth]
    {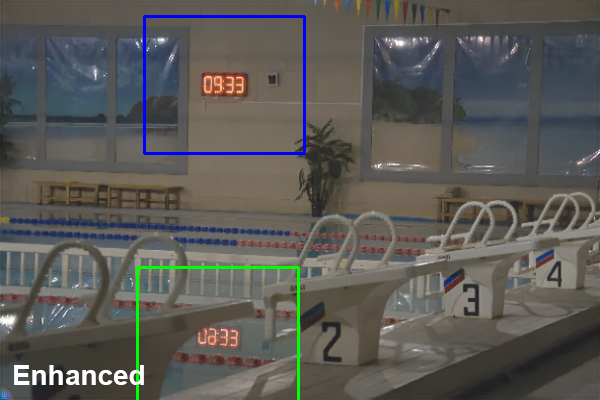}
    \caption{}
\end{subfigure}
\begin{subfigure}{.24\linewidth}
    \centering
    \includegraphics[keepaspectratio=true, width=\textwidth]{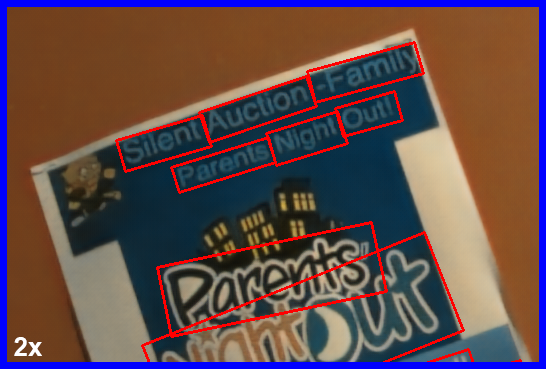}    
    \includegraphics[keepaspectratio=true, width=\textwidth]{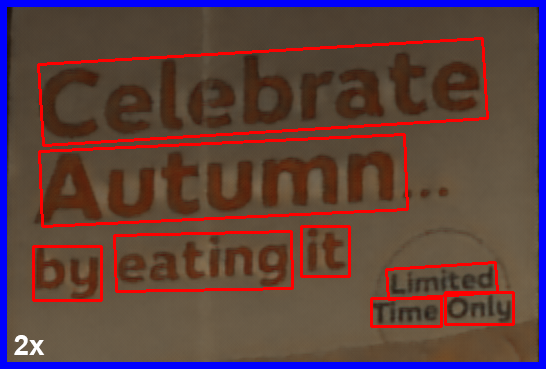}
    \includegraphics[height=2cm, width=\textwidth]
    {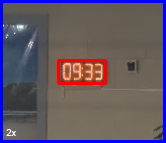}
    \caption{}
\end{subfigure}
\begin{subfigure}{.24\linewidth}
    \centering
    \includegraphics[keepaspectratio=true, width=\textwidth]{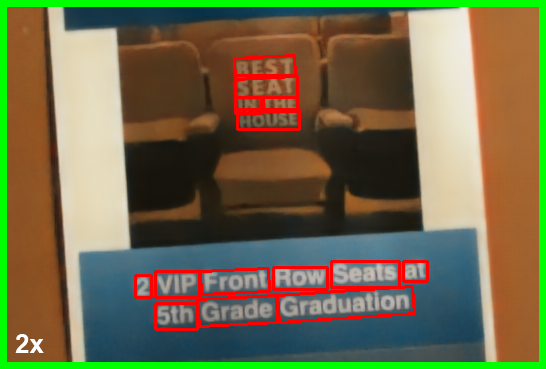}
    \includegraphics[keepaspectratio=true, width=\textwidth]{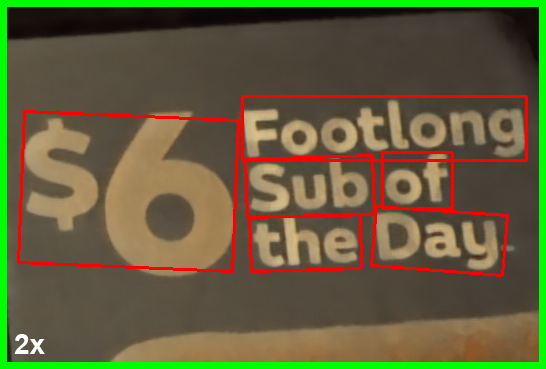}
    \includegraphics[height=2cm, width=\textwidth]
    {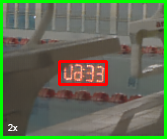}
    \caption{}
\end{subfigure}

\caption{From left to right: (a) Original images; (b) Enhanced results with our proposed method; (c-d) Zoomed-in (2x) regions of the blue and green bounding boxes. Top row: SID-Sony-Text; Middle row: SID-Fuji-Text; Bottom row: LOL-Text. Extremely low-light images in the SID dataset are significantly darker than those in the LOL dataset, and our model enhances the images to the extent that texts are clearly visible with sharp edges.}
\label{fig:overall_results}
\end{figure*}

Scene text understanding involves extracting text information from images through text detection and recognition, which is a fundamental task in computer vision. However, performance drops sharply when images are captured under low-light conditions. The main difficulty in detecting text in low-light images is that low-level features, such as edges and character strokes, are no longer prominent or hardly visible. On the other hand, enhancing images captured in extremely low-light conditions pose a greater challenge than ordinary low-light images due to the higher noise levels and greater information loss. For instance, we show the difference in darkness level in Figure~\ref{fig:overall_results}, where it is evident that the See In the Dark (SID) datasets~\cite{Chen2018LearningTS} are darker and, in theory, more difficult to enhance than the LOw-Light (LOL) dataset~\cite{Wei2018DeepRD}. Quantitatively, we calculated the PSNR and SSIM values for two subsets of SID, SID-Sony and SID-Fuji, and LOL by comparing each image against pure black images in Table~\ref{table:SID_LOL_black_level_comparison}. Based on each dataset's average perceptual lightness (L* in the CIELAB color space), images in SID are at least 15 times darker than those in LOL. Hence, low-light image enhancement is a necessary pre-processing step for scene text extraction under such conditions. 

Over the years, many general or low-light image enhancement models have been proposed to improve the interpretability and extraction of information in images by providing better input for subsequent image content analysis. Early methods~\cite{jobson1997properties, Guo2017LIMELI, Ying2017ABM} typically attempted to restore the statistical properties of low-light images to those of long-exposure images from a mathematical perspective. On the other hand, deep learning-based methods~\cite{Wei2018DeepRD, Chen2018LearningTS, Jiang2019EnlightenGANDL, Zero-DCE} aim to learn the mapping between low-light images and their corresponding long-exposure versions via regression. To the best of our knowledge, most existing low-light image enhancement works have not explicitly addressed the restored image quality in terms of downstream scene text tasks.



\begin{table*}[!t]
\centering
\begin{adjustbox}{width=0.6\linewidth,center}
\begin{tabular}{lccc}
\cmidrule[2pt]{1-4}
Dataset & PSNR $\uparrow$ & SSIM $\uparrow$ & Avg. L* $\downarrow$ \\ \cmidrule[2pt]{1-4}
SID-Sony~\cite{Chen2018LearningTS} & 44.350 & 0.907& 0.009 \\
SID-Fuji~\cite{Chen2018LearningTS} & 41.987 & 0.820 & 0.004 \\
LOL~\cite{Wei2018DeepRD} & 23.892 & 0.195 & 0.142 \\ \hline
Pure Black & $\infty$ & 1.000 & 0.000 \\ \cmidrule[2pt]{1-4}
\end{tabular}
\end{adjustbox}
\caption{The difference between the extremely low-light dataset, SID, and the ordinary low-light dataset, LOL, is shown in terms of PSNR and SSIM values, computed by comparing short-exposure images against pure black images. Avg. L* is the average perceptual lightness in the CIELAB color space, calculated based on short-exposure images. Scores are averaged across training and test sets. Higher PSNR and SSIM values, along with lower Avg. L*, indicate darker images that are more challenging for image enhancement and scene text extraction.}
\label{table:SID_LOL_black_level_comparison}
\end{table*}

Recent advancements in visual attention mechanisms have demonstrated their effectiveness in identifying and boosting salient features in images. Channel-only attention~\cite{hu2018squeeze,hu2018gather,chen2023interpretable}, spatial attention~\cite{ju2023keep,hou2023canet} or the subsequent channel-spatial attention~\cite{woo2018cbam, fu2019dual} modules were proposed to emphasize the most informative areas. However, these methods cannot preserve texture details, especially fine-grained edge information that is intuitively needed to enhance extremely low-light images with complex textures. To overcome this limitation, we introduce Edge-Aware Attention (Edge-Att). This novel attention module simultaneously performs channel and spatial attention-based feature learning on high-level image and edge features. Our model also considers text information in the image through a text-aware loss function. This way, our model can effectively enhance low-light images while preserving fine-grained edge information, texture details, and legibility of text.

The scarcity of extremely low-light text datasets presents a hurdle for further research. To address this, we annotated all text instances in both the training and testing sets of the SID and LOL datasets, creating three new low-light text datasets: SID-Sony-Text, SID-Fuji-Text, and LOL-Text. We then proposed a novel Supervised Deep Curve Estimation (Supervised-DCE) model to synthesize extremely low-light scene text images based on the commonly used ICDAR15 (IC15) scene text dataset. It allows researchers to easily translate naive scene text datasets into extremely low-light text datasets. In addition to the previously published conference version of this work~\cite{hsu2022extremely}, we have made four significant extensions. Firstly, we propose a novel dual encoder-decoder framework that can achieve superior performance on low-light scene text tasks (Section~\ref{sec:prposed_en}). Secondly, we introduce a new image synthesis method capable of generating more realistic extremely low-light text images (Section~\ref{sec:img_syn}). Thirdly, we have further annotated texts in the Fuji and LOL datasets, thereby forming the largest low-light scene text datasets to date (Section~\ref{sec:datasets}). Fourthly, comprehensive experiments and analyses are carried out to study the latest methods along with our proposed methods on all synthetic and real low-light text datasets. The main contributions of our work are as follows:

\begin{itemize}
\item We present a novel scene text-aware extremely low-light image enhancement framework with dual encoders and decoders to enhance extremely low-light images, especially scene text regions within them. Our proposed method is equipped with Edge-Aware Attention modules and trained with new Text-Aware Copy-Paste (Text-CP) augmentation. Our model can restore images in challenging lighting conditions without losing low-level features.

\item We developed a Supervised-DCE model to synthesize extremely low-light images. This allows us to use existing publicly available scene text datasets such as IC15 to train our model alongside genuine ones for scene text research under such extreme lighting conditions.

\item We labeled the texts in the SID-Sony, SID-Fuji, and LOL datasets and named them SID-Sony-Text, SID-Fuji-Text, and LOL-Text, respectively. This provides a new perspective for objectively assessing enhanced extremely low-light images through scene text tasks.
\end{itemize}



\section{Related Works}
{\noindent\bf Low-light Image Enhancement.} Retinex theory assumes that an image can be decomposed into illumination and reflectance. Most Retinex-based methods enhance results by removing the illumination part~\cite{jobson1997properties}, while others such as LIME~\cite{Guo2017LIMELI} keep a portion of the illumination to preserve naturalness. BIMEF~\cite{Ying2017ABM} further designs a dual-exposure fusion framework for accurate contrast and lightness enhancement. RetinexNet~\cite{Wei2018DeepRD} combines deep learning and Retinex theory, adjusting illumination for enhancement after image decomposition. The recent successes of generative adversarial networks (GANs)~\cite{Goodfellow2014GenerativeAN} have attracted attention from low-light image enhancement because GANs have proven successful in image translation. Pix2pix~\cite{isola2017image} and CycleGAN~\cite{CycleGAN2017} have shown good image-translation results in paired and unpaired image settings, respectively. To overcome the complexity of CycleGAN, EnlightenGAN~\cite{Jiang2019EnlightenGANDL} proposed an unsupervised one-path GAN structure. Besides general image translation, \cite{Chen2018LearningTS} proposed learning-based low-light image enhancement on raw sensor data to replace much of the traditional image processing pipeline, which tends to perform poorly on such data. EEMEFN~\cite{zhu2020eemefn} also attempted to enhance images using multi-exposure raw data that is not always available.



Zero-Reference Deep Curve Estimation (Zero-DCE)~\cite{Zero-DCE} designed a light-weight CNN to estimate pixel-wise high-order curves for dynamic range adjustment of a given image without needing paired images. \cite{ma2022toward} designed a novel Self-Calibrated Illumination (SCI) learning with an unsupervised training loss to constrain the output at each stage under the effects of a self-calibrated module. ChebyLighter~\cite{pan2022chebylighter} learns to estimate an optimal pixel-wise adjustment curve under the paired setting. Recently, the Transformer~\cite{vaswani2017attention} architecture has become the de-facto standard for Natural Language Processing (NLP) tasks. ViT~\cite{dosovitskiyimage} applied the attention mechanism in the vision task by splitting the image into tokens before sending it into Transformer. Illumination Adaptive Transformer (IAT)~\cite{cui2022you} uses attention queries to represent and adjust ISP-related parameters. Most existing models enhance images in the spatial domain. Fourier-based Exposure Correction Network (FECNet)~\cite{huang2022deep} presents a new perspective for exposure correction with spatial-frequency interaction and has shown that their model can be extended to low-light image enhancement.

{\noindent\bf Scene Text Extraction.}
Deep neural networks have been widely used for scene text detection. CRAFT~\cite{Baek2019CharacterRA} predicts two heatmaps: the character region score map and the affinity score map. The region score map localizes individual characters in the image, while the affinity score map groups each character into a single word instance. Another notable scene text detection method is Pixel Aggregation Network (PAN)~\cite{wang2019efficient} which is trained to predict text regions, kernels, and similarity vectors. Both text segmentation models have proven to work well on commonly used scene text datasets such as IC15~\cite{Karatzas2015ICDAR2C} and TotalText~\cite{totaltext}. Inspired by them, we introduced a text detection loss in our proposed model to focus on scene text regions during extremely low-light image enhancement. Furthermore, state-of-the-art text recognition methods such as ASTER~\cite{Aster2019} and TRBA~\cite{trba2019} are known to perform well on images captured in complex scenarios. ASTER~\cite{Aster2019} employs a flexible rectification module to straighten the word images before passing them to a sequence-to-sequence model with the bi-directional decoder. The experimental results of ASTER showed that the rectification module could achieve superior performance on multiple scene text recognition datasets, including the likes of IC15 and many more. Besides, TRBA~\cite{trba2019} provided interesting insights by breaking down the scene text recognition framework into four main stages: spatial transformation, character feature extraction, followed by sequence modeling, and the prediction of character sequences. Given these methods' robustness on difficult texts, they are well-suited to recognize texts from enhanced low-light images.


\section{Extremely Low-Light Text Image Enhancement}

\subsection{Problem Formulations}
Let $x \in R^{W\times H\times 3}$ be a short-exposure image of width $W$ and height $H$. An ideal image enhancement expects that a neural network $LE(x;\theta)$ parameterized by $\theta$ can restore this image to its corresponding long-exposure image, $y \in R^{W\times H\times 3}$, i.e., $LE(x;\theta)\simeq y$. However, previous works normally pursued the lowest per-pixel intensity difference, which should not be the goal for image enhancement because we usually expect that some high-level computer vision tasks can work reasonably well on those enhanced images. For example, in terms of text detection, the goal of the neural network can be the lowest detection bounding boxes discrepancy, i.e., $B(LE(x;\theta)) \simeq B(y)$.

Our novel image enhancement model consists of a U-Net accommodating extremely low-light images and edge maps using two independent encoders. During model training, instead of channel attention, the encoded edges guide the spatial attention sub-module in the proposed Edge-Att to attend to edge pixels related to text representations. Besides the image enhancement losses, our model incorporates text detection and edge reconstruction losses into the training process. This integration effectively guides the model's attention towards text-related features and regions, facilitating improved image textual content analysis. As a pre-processing step, we introduced a novel augmentation technique called Text-CP to increase the presence of non-overlapping and unique text instances in training images, thereby promoting comprehensive learning of text representations.

\label{sec:prposed_en}
\subsection{Network Design}

\begin{figure*}[!ht]
	\centering
	\includegraphics[width=\linewidth]{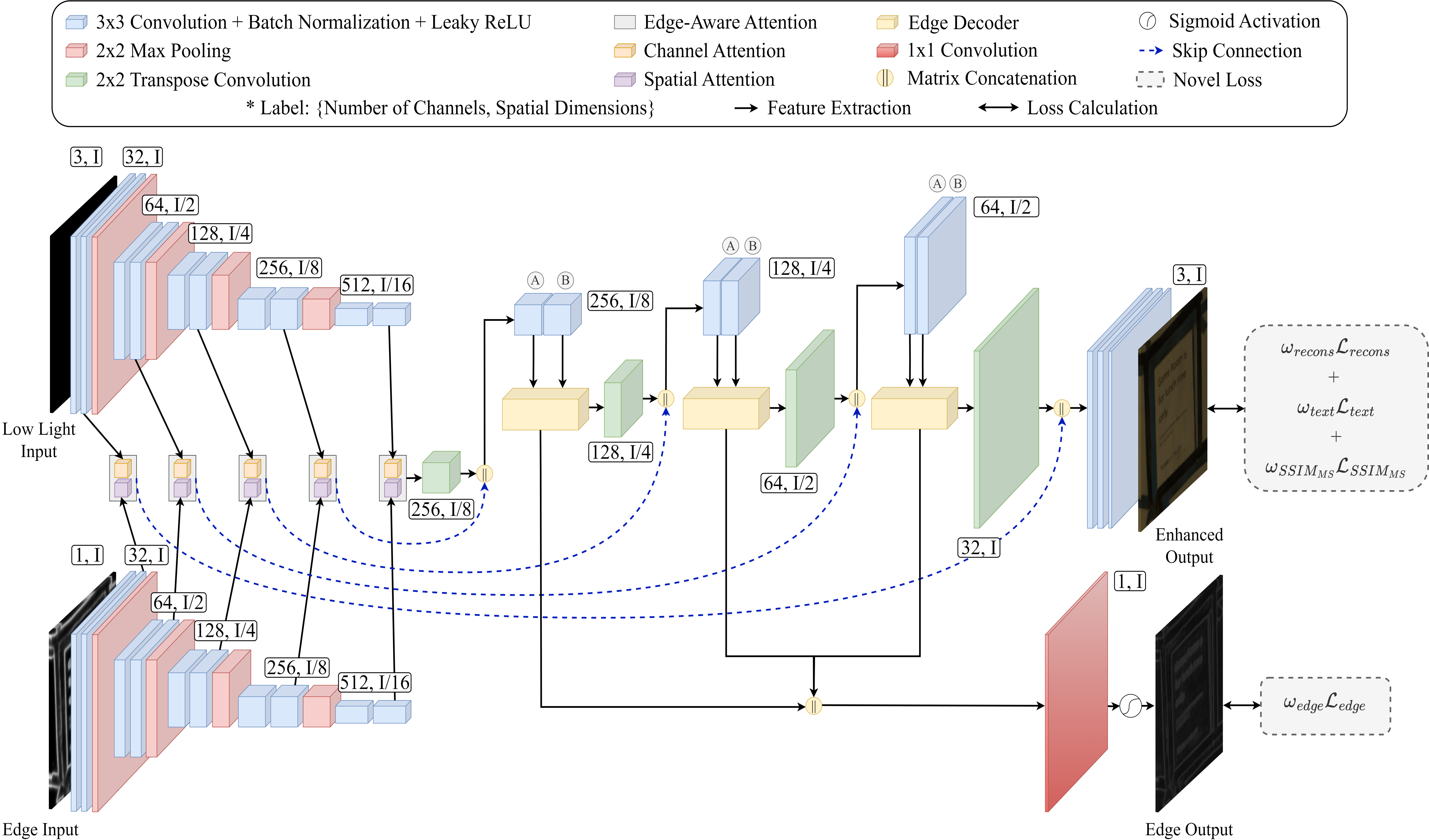}
	\caption{Illustration of the architecture of our proposed framework, designed to enhance extremely low-light images while incorporating scene text awareness.}
	\label{fig:architecture}
\end{figure*}

\begin{figure*}[!t]
	\centering
	\includegraphics[width=\linewidth]{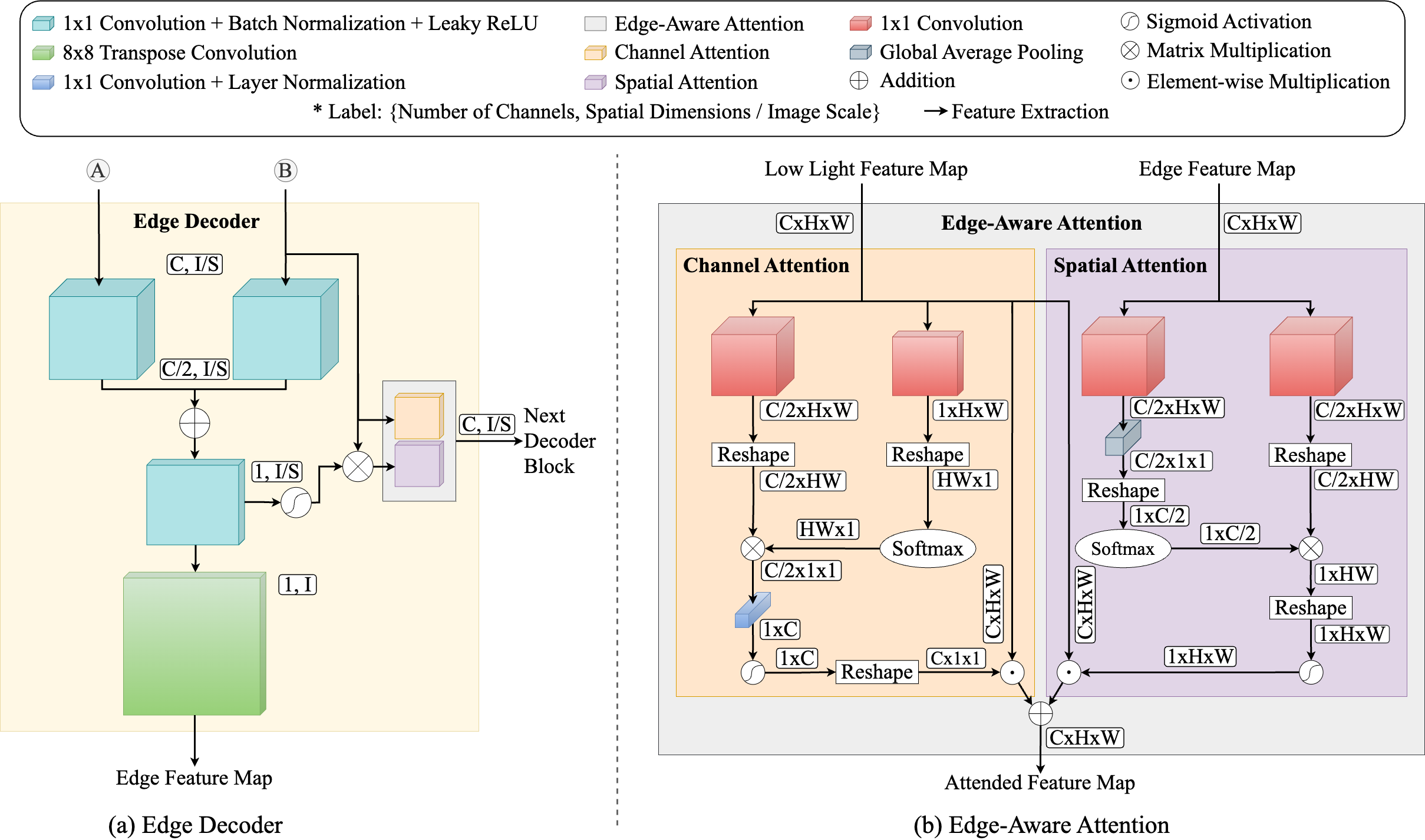}
	\caption{(a) Visual representation of our edge decoder, wherein A and B represent the output from the corresponding convolution blocks in Figure ~\ref{fig:architecture} and S denotes the scaling of the image. (b) Illustration of the proposed Edge-Aware Attention module.}
	\label{fig:edge_decoder_psa_combine}
\end{figure*}

Our model was inspired by U-Net\cite{Chen2018LearningTS} with some refinements. Firstly, the network expects heterogeneous inputs, i.e., extremely low-light images, $x$, and the corresponding RCF~\cite{Liu2019RicherCF} edge maps, $e$. Secondly, input-edge pairs are handled by two separate encoders with edge-aware attention modules between them. The attended features are then bridged with the decoder through skip connections. Finally, our multi-tasking network predicts the enhanced image, $x'$, and the corresponding reconstructed edge, $e'$. The overall architecture of our network can be seen in Figure ~\ref{fig:architecture} and modeled as:
\begin{equation}
x',e' = LE(x,e;\theta).
\label{eq:overall_network}
\end{equation}

\subsection{Objectives}
Our proposed model is trained to optimize four loss functions. The first two, Smooth L1 loss and multi-scale SSIM loss focus on enhancing the overall image quality. The third, text detection loss, targets the enhancement of scene text regions specifically. The fourth, edge reconstruction loss, focuses on crucial low-level edge features.

Firstly, we employ smooth L1 loss as the reconstruction loss to better enforce low-frequency correctness~\cite{isola2017image} between $x'$ and $y$ as:
\begin{equation}
\begin{aligned}
\mathcal{L}_{recons}=\left\{\begin{matrix}
& 0.5 \cdot (x'-y)^2/\delta, & \text{if } {|x'-y| < \delta} \\
& |x'-y| - 0.5 \cdot \delta, & \text{otherwise}{}
\end{matrix}\right.
\end{aligned}
\label{eq:reconstruction_loss}
\end{equation}

\noindent where we empirically found that $\delta = 1$ achieved good result. The authors of Pix2Pix \cite{isola2017image} showed that by utilizing L1 loss, the model can achieve better results as the generated images are less blurry and proved that L1 loss can better enforce the learning of low-frequency details, which is also essential for OCR tasks. On the other hand, the L1 norm is less sensitive to outliers than the L2 norm, thus resulting in a more robust model towards extreme pixel intensities.

Secondly, the multi-scale SSIM metric was proposed in~\cite{Wang2003MultiscaleSS} for reference-based image quality assessment, focusing on image structure consistency. An $M$-scale SSIM between the enhanced image $x'$ and ground truth image $y$ is:
\begin{equation}
	SSIM_{MS}(x', y) = [l_M(x', y)]^{\tau} \cdot \prod\nolimits^M_{j=1} [c_j(x', y)]^{\phi} [s_j(x', y)]^{\psi},
 \label{eq:msssim_loss}
\end{equation}
where $l_M$ is the luminance at \textit{M}-scale; $c_j$ and $s_j$ represent the contrast and the structure similarity measures at the $j$-th scale; $\tau$, $\phi$, and $\psi$ are parameters to adjust the importance of the three components. Inspired by ~\cite{Wang2003MultiscaleSS}, we adopted the $M$-scale SSIM loss function in our work to enforce the image structure of $x'$ to be close to that of $y$:
\begin{equation}
	\mathcal{L}_{SSIM_{MS}} = 1 - {SSIM_{MS}}(x', y).
\end{equation}

Thirdly, a well-enhanced extremely low-light image implies that we could obtain similar text detection results on both the enhanced and ground truth images. As such, we propose to employ CRAFT~\cite{Baek2019CharacterRA} to localize texts in images through its region score heatmap. To implicitly enforce our model to focus on scene text regions, we define the text detection loss, \textit{$\mathcal{L}_{text}$} as: 
\begin{equation}
	\mathcal{L}_{text} = \| R(x') - R(y)\|_1,
\label{eq:text_loss}
\end{equation}
\noindent where $R(x')$ and $R(y)$ denote the region score heatmaps of the enhanced and ground truth images, respectively. 

Fourthly, the edge reconstruction decoder in our model is designed to extract edges better, which are essential for text pixels. Figure \ref{fig:edge_decoder_psa_combine}(a) shows an overview of the edge decoder. The loss at pixel $i$ of detected edge, $e_i$, with respect to the ground truth edge, $g_i$ is defined as:
\begin{equation}
\begin{aligned}
	l(e_i)=\left\{\begin{matrix}
& \alpha \cdot log(1-P(e_i)), & \text{if } g_i = 0 \\
& \beta \cdot logP(e_i), & \text{if } g_i = 1
\end{matrix}\right.
\end{aligned}
\label{eq:edge_loss1}
\end{equation}

\noindent where
\begin{equation}
\begin{aligned}
& \alpha = \lambda \cdot \frac{\left | Y^+ \right |}{\left | Y^+ \right | + \left | Y^- \right |}, \\
& \beta =  \frac{\left | Y^- \right |}{\left | Y^+ \right | + \left | Y^- \right |},
 \end{aligned}
 \label{eq:edge_loss2}
\end{equation}

\noindent $Y^+$ and $Y^-$ denote the positive and negative sample sets, respectively. $\lambda$ is set to 1.1 to balance both types of samples. The ground truth edge is generated using a Canny edge detector~\cite{Canny1986ACA}, and P($e_i$) is the sigmoid function. Then, the overall edge reconstruction loss can be formulated as:
\begin{equation}
\mathcal{L}_{edge} =\sum_{i=1}^{|I|}\sum_{j=1}^{J} l(e^{j}_i)+l(e^{'}_i),
 \label{eq:edge_fus}
\end{equation}

\noindent where $l(e_i^{j})$ is the predicted edge at pixel $i$ and level $j$. $J=3$ is the number of side edge outputs in our model. $e_i^{'}$ is the final predicted edge map from the concatenation of side outputs. $|I|$ is the number of pixels in a cropped image during training.

Finally, the total joint loss function, $\mathcal{L}_{total\_en}$ of our proposed model is:
\begin{equation}
	\mathcal{L}_{total\_en} = \omega_{recons} \mathcal{L}_{recons} + \omega_{text} \mathcal{L}_{text} + \omega_{SSIM_{MS}} \mathcal{L}_{SSIM_{MS}} + \omega_{edge} \mathcal{L}_{edge},
	 \label{eq:loss_all}
\end{equation}
\noindent where $\omega_{recons}$, $\omega_{text}$, $\omega_{SSIM_{MS}}$, and $\omega_{edge}$ are the weights to address the importance of each loss term during training.

\subsection{Edge-Aware Attention}
Polarized Self-Attention (PSA)~\cite{liu2021polarized} is one of the first works to propose an attention mechanism catered to high-quality pixel-wise regression tasks. However, we found that the original PSA module that only considers a single source of feature map for both channel and spatial attention is ineffective for extremely low-light image enhancement. Under low light conditions, the details of content such as the edges of the texts are barely discernible which is less effective in guiding the network to attend to spatial details. Therefore, we designed our Edge-Aware Attention (Edge-Att) module to take in feature maps from two encoders and process them differently, i.e., the feature maps of extremely low-light images from the image encoder are attended channel-wise, whereas the spatial attention submodule attends to feature maps from the edge encoder. By doing so, we can ensure that Edge-Att can attend to rich images and edge features simultaneously. The proposed attention module is illustrated in Figure ~\ref{fig:edge_decoder_psa_combine}(b).

Firstly, the feature map from the image encoder, $F$ is fed into the channel attention, $A^{ch}(F) \in \mathbb{R}^{C \times 1 \times 1}$ with calculation as follows:
\begin{equation}
A^{ch}(F)= \sigma_3 \left [ F_{SG}(W_z(\sigma_1(W_v(F)) \times F_{SM}(\sigma_2(W_q(F))))) \right ],
\label{eq:channel_attention}
\end{equation}

\noindent where $W_q$, $W_v$, and $W_z$ are 1x1 convolution layers, $\sigma_1$, $\sigma_2$ and $\sigma_3$ are tensor reshape operators. $F_{SM}(.)$ and $F_{SG}(.)$ refer to softmax and sigmoid operators. The output of this branch is $A^{ch}(F)\bigodot^{ch} F \in \mathbb{R}^{C\times H \times W}$, where $\bigodot^{ch}$ is a channel-wise multiplication operator.

Secondly, given the edge-branch feature map $E$, the edge-aware spatial attention, $A^{sp}(E) \in \mathbb{R}^{1 \times H \times W}$, is defined as:
\begin{equation}
A^{sp}(E)= F_{SG} \left [\sigma_3 (F_{SM}(\sigma_1(F_{GP}(W_q(E))))\times 
\sigma_2(W_v(E))) \right ],
\label{eq:edge_attention}
\end{equation}

\noindent where $W_q$ and $W_v$ are 1x1 convolution layers, $\sigma_1$, $\sigma_2$, and $\sigma_3$ are three tensor reshape operators. $F_{SM}(.)$ is a softmax operator, $F_{GP}(.)$ is a a global pooling operator, and $F_{SG}(.)$ is a sigmoid operator. The output of this branch is $A^{sp}(E)\bigodot^{sp} F \in \mathbb{R}^{C\times H \times W}$, where $\bigodot^{sp}$ is a spatial-wise multiplication operator, and $F$ is the image enhancement branch's feature map. Finally, output of the proposed Edge-Att module is the composition of two submodules:
\begin{equation}
Edge{\text -}Att(F,E) = A^{ch}(F)\odot^{ch} F + A^{sp}(E)\odot^{sp} F.
\label{eq:psa_attention}
\end{equation}

\subsection{Text-Aware Copy-Paste Augmentation}
\begin{figure*}[!htb]
	\centering
	\includegraphics[width=\linewidth]{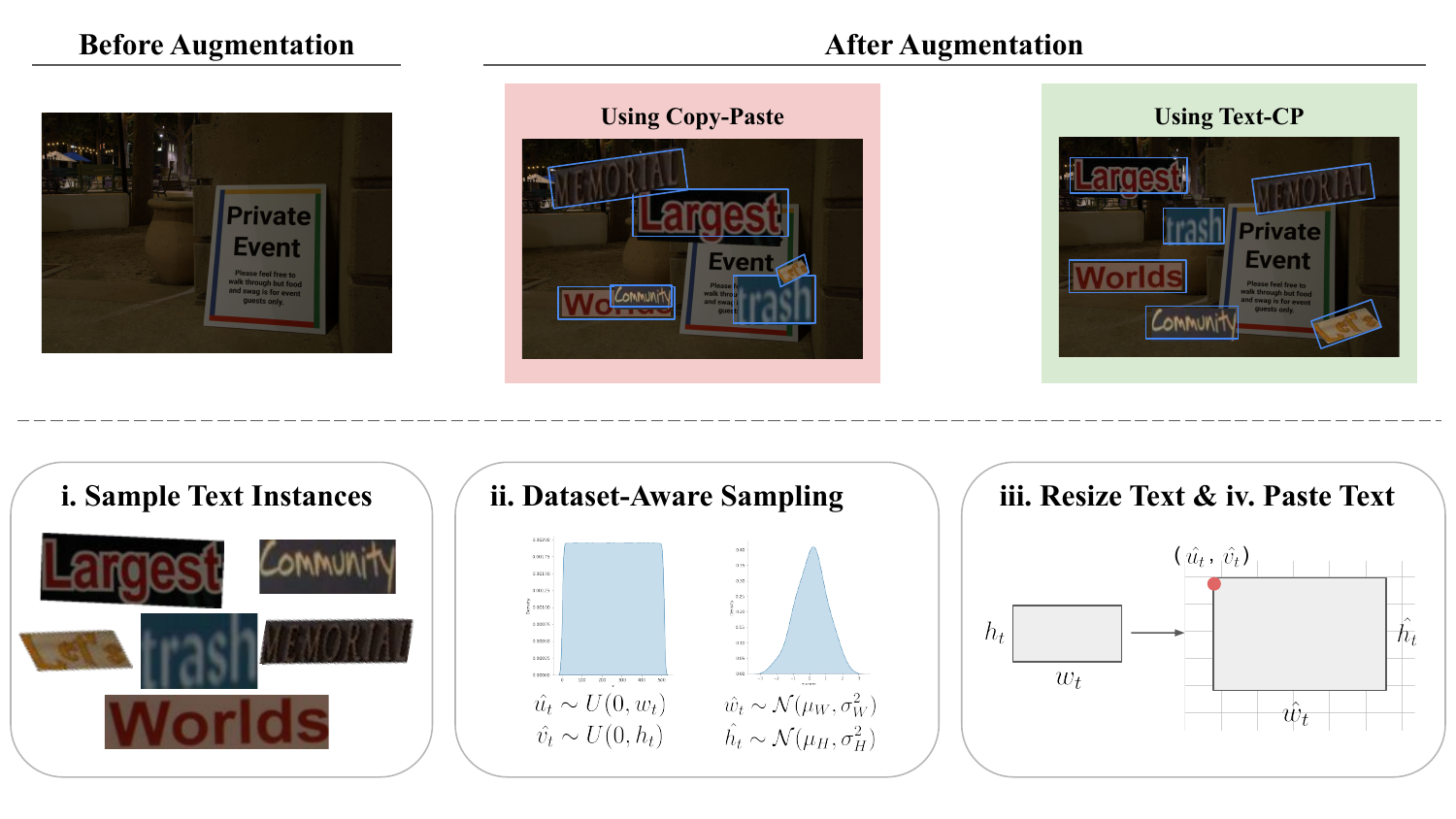}
    \caption{Illustration of the Text-Aware Copy-Paste (Text-CP) data augmentation. Compared with the original Copy-Paste, our method generates images with non-overlapping text instances that allow the detection of texts outside their usual context.}
	\label{fig:cap_gaussian}
\end{figure*}

This work aims to enhance extremely low-light images to improve text detection and recognition. However, the dataset's limited number of text instances could hinder the model's ability. Although Copy-Paste Augmentation \cite{Ghiasi_2021_CVPR_CAP} can increase the number of text instances, overlapping texts introduced by random placement might confuse CRAFT in text detection loss since CRAFT is not trained to detect such texts. In the commonly used scene text datasets such as ICDAR15~\cite{Karatzas2015ICDAR2C}, overlapping texts are marked as "do not care" regions which are excluded from models' training and evaluation. Thus, to adhere to ICDAR's standard and to address overlapping text issues, we propose a novel approach called Text-Aware Copy-Paste Augmentation (Text-CP). Text-CP considers each text box's location and size by leveraging uniform and Gaussian distributions derived from the dataset.
For a training image $t$ of width $w_t$ and height $h_t$ to be augmented, we initialize a set of labeled text boxes in the training set as $C$, which is:
\begin{equation}
\mathit{\emph{C}}=\left \{ (u_1,v_1,w_1,h_1), ...,(u_{\left | \mathit{\emph{C}}  \right |},v_{\left | \mathit{\emph{C}}  \right |},w_{\left | \mathit{\emph{C}}  \right |},h_{\left | \mathit{\emph{C}}  \right |})) \right \},
\label{eq:box_annotation}
\end{equation}

\noindent where each tuple represents the top left position of a text located at $u_k$ and $v_k$ with width, $w_k$, and height, $h_k$ with $k$ representing the index of the current text's box in the set. We then sample a target number of text instances, $n_{\text{target}}$, from the set of $C$ to form $C_t$, defined as the set of text boxes to be pasted on that training image, $t$. The next step is to crop and paste the sampled texts without overlapping. For each $c_k \in C_t$, we adopt two uniform distributions in modeling the position of the texts, $\hat{u}_k$ and $\hat{v}_k$:
\begin{equation}
\begin{aligned}
& \hat{u}_k \sim U(0,w_t), \\
& \hat{v}_k \sim U(0,h_t).
 \label{eq:uniform_distribution_height_width}
 \end{aligned}
\end{equation}

\noindent As for $w_k$ and $h_k$, they are sampled from Gaussian distribution as:
\begin{equation}
\begin{aligned}
& \hat{w}_k \sim \mathcal{N}(\mu_W, \sigma_{W}^{2}), \\
& \hat{h}_k \sim \mathcal{N}(\mu_H, \sigma_{H}^{2}),
 \label{eq:uniform_distribution_wh}
 \end{aligned}
\end{equation}

\noindent where $\mu$ and $\sigma^2$ are estimated means and variances of width $W$ and height $H$ from all the labeled texts in the training set. We illustrate the overall data augmentation process of Text-CP and its augmented results in Figure \ref{fig:cap_gaussian}. The pseudocode of Text-CP is detailed in the supplementary material. 

\section{Extremely Low-Light Image Synthesis}
\subsection{Problem Formulations}
To the best of our knowledge, the research community has not extensively explored extremely low-light image synthesis, mainly due to the limited availability of datasets designed explicitly for extremely low-light scene text. While extremely low-light dataset, SID, and low-light dataset, LOL, exist, they are not primarily collected with scene text in mind. This scarcity of dedicated datasets for extremely low-light scene text poses challenges for evaluating the performance of existing image enhancement methods in terms of image quality and scene text metrics. In order to address this issue, we define the extremely low-light image synthesis problem as follows:
\begin{equation}
\hat{x} = LS(y; \theta_s),
\label{eq:img_syn}
\end{equation}

\noindent where given a long-exposure image $y$, a low-light image synthesis neural network, $LS(y;\theta_s)$ parameterized by $\theta_s$, will synthesize a set of images $\hat{x}$, such that $B(LS(y;\theta_s)) \simeq B(x)$. We want the synthesized extremely low-light images to be as realistic as possible to genuine low-light images, $x$.

Therefore, we introduce a Supervised-DCE model focusing on synthesizing a set of realistic extremely low-light images, enabling existing image enhancement techniques to leverage publicly available scene text datasets. Consequently, existing low-light image enhancement methods can benefit from training with synthetic data to the extent that they can perform better on the downstream scene text detection task, as detailed in Section \ref{exp:mixing_results}.

\label{sec:img_syn}
\subsection{Network Design}

\begin{figure*}[!htbp]
	\centering
	\includegraphics[width=\linewidth]{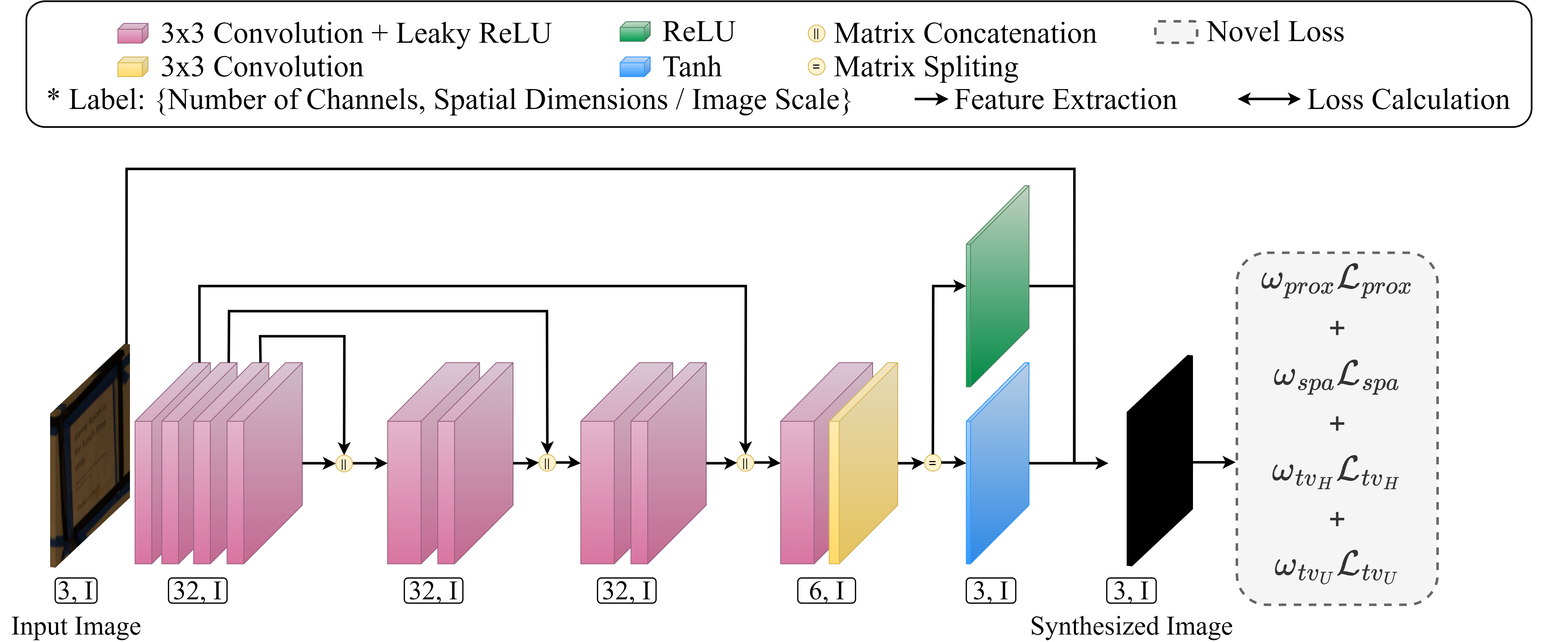}
	\caption{Illustration of the proposed Supervised-DCE model for extremely low-light image synthesis.}
	\label{fig:S-DCE-architecture}
\end{figure*}
Zero-DCE~\cite{Zero-DCE} was originally proposed to perform image enhancement through curve estimation. However, its network can only adjust brightness slightly since the per-pixel trainable curve parameter, $\alpha$, in the quadratic curve limits the pixel variation. The advantage of performing intensity adjustment in terms of the quadratic curve is that the pixel range can be better constrained. In this work, we propose a Supervised-DCE model that learns to provide reconstructable extremely low-light images with paired short- and long-exposure images. The framework of our image synthesis network, Supervised-DCE, can be seen in Figure \ref{fig:S-DCE-architecture}. Our goal is to push most values closer to zero n the context of synthesizing extremely low-light images. Accordingly, we propose a reformulation of the DCE model as follows:
\begin{equation}
\hat{x} = -(H(y) + U(y))y^2 + (1 + H(y))y,
\label{eq:DCE_enhancement}
\end{equation}

\noindent where $y$ is the input (i.e., long-exposure image); $\hat{x}$ is the synthesized low-light image; $H(y)$ and $U(y)$ are the output of Tanh and ReLU branches, respectively. By introducing the second $U(y)$ branch, we eliminate the need for iteratively applying the model to produce the desired output, and drastic intensity adjustment can be done with only a single iteration.

In the original Zero-DCE model, image enhancement is learned by setting the exposure value to 0.6 in the exposure control loss. However, manually setting an exposure value to synthesize extremely low-light images is too heuristic and inefficient. 
In our proposed synthesis framework, the overall learning is done by training SID long-exposure images to be similar to their short-exposure counterparts. Most importantly, these images are translated in the hope that their text information is somewhat deteriorated as the one in genuine extremely low-light images and cannot be easily reconstructed. Then, the trained model can be used to transform scene text datasets in the public domain to boost the performance of extremely low-light image enhancement in terms of text detection.

\subsection{Objectives}
During extremely low-light image synthesis, we expect the output to maintain spatial consistency while reducing the overall proximity loss:
\begin{equation}
	\mathcal{L}_{prox} = \| \hat{x} - x\|_1 + \mathcal{L}_{entropy}(\hat{x}, x) + \mathcal{L}_{smoothness}(\hat{x}, x),
\label{eq:prox_loss}
\end{equation}

\noindent where $\hat{x}$ is the synthesized extremely low-light image given the long-exposure image $y$, and $x$, is the genuine low-light image, i.e., ground truth for $\hat{x}$. Entropy loss, $\mathcal{L}_{entropy}$, and smoothness loss, $\mathcal{L}_{smoothness}$~\cite{Samangouei2018ExplainGANME}, are also used to encourage the differences to be both sparse and local. With the introduction of $\mathcal{L}_{prox}$, we removed the color constancy loss of the original Zero-DCE model since color constancy can be enforced through the supervised loss. 

The spatial consistency loss, $\mathcal{L}_{spa}$ encourages spatial coherence of the synthesized image by preserving the difference of neighboring regions between the input image and its synthesized low-light version:
\begin{equation}
\mathcal{L}_{spa} = \frac{1}{\mathcal{M}}\sum_{i=1}^{\mathcal{M}}\sum_{j \in \omega(i)}(|\hat{X}_i-\hat{X}_j|- \alpha_s \log_{10} (9 |Y_i-Y_j|+1))^2,
\label{eq:spa_loss}
\end{equation}

\noindent where $\mathcal{M}$ is the number of local regions, and $\omega(i)$ is four neighboring regions (top, down, left, right) centered at region $i$. $\hat{X}$ and $Y$ are the averaged intensity values of local regions of the synthesized images and the long-exposure images, respectively. We introduced logarithm operation and $\alpha_s$ parameter to reduce the large spatial difference of $Y$ where $\alpha_s$ is set to 0.05. We set the local region size to $4\times4$, following the original setting of Zero-DCE.

Besides spatial consistency, we also expect the monotonicity relation between neighboring pixels to be preserved. To achieve this, we reused the illumination smoothness loss:
\begin{equation}
\mathcal{L}_{tv_{Z}} = \sum_{\forall c \in \xi} (|\nabla_x Z^c| + |\nabla_y Z^c|)^2, \xi = \left \{ R,G,B \right \},
\label{eq:illu_smoothness_loss}
\end{equation}

\noindent where $\nabla_x$ and $\nabla_y$ are gradient operations on the x-axis and y-axis, respectively. Illumination smoothness loss, $\mathcal{L}_{tv_{Z}}$, is applied on both $H(y)$ and $U(y)$, i.e., the curve parameter maps of the two branches, respectively, by substituting $Z$ with $H$ and $U$, resulting in $\mathcal{L}_{tv_{H}}$ and $\mathcal{L}_{tv_{U}}$.

In summary, the overall learning objective, $\mathcal{L}_{total\_syn}$ to train our extremely low-light image synthesis network is defined as:
\begin{equation}
\mathcal{L}_{total\_syn} = \omega_{prox}\mathcal{L}_{prox} + \omega_{spa}\mathcal{L}_{spa} + \omega_{tv_{H}}\mathcal{L}_{tv_{H}} + \omega_{tv_{U}}\mathcal{L}_{tv_{U}}.
\label{eq:DCE_loss_all}
\end{equation}


\section{New Low-Light Text Datasets}
\label{sec:datasets}

\begin{table*}[!htbp]
\begin{adjustbox}{width=\linewidth,center}
\begin{tabular}{lcccccccccc}
\cmidrule[2pt]{1-11}
\multirow{2}{*}{Dataset} & \multicolumn{7}{c}{Training Set} & \multicolumn{3}{c}{Testing Set} \\ \cmidrule(lr){2-8} \cmidrule(lr){9-11} 
 & GT Img. & Leg. & Illeg. & $\mu_W$ & $\mu_H$ & $\sigma_W$ & \multicolumn{1}{c}{$\sigma_H$} & GT Img. & Leg. & Illeg. \\ \cmidrule[2pt]{1-11}
SID-Sony-Text & 161 & 5937 & 2128 & 79.270 & 34.122 & 123.635 & 50.920 & 50 & 611 & 359 \\
SID-Fuji-Text & 135 & 6213 & 4534 & 128.579 & 57.787 & 183.199 & 68.466 & 41 & 1018 & 1083 \\
LOL-Text & 485 & 613 & 1423 & 23.017 & 14.011 & 21.105 & 17.542 & 15 & 28 & 45 \\
IC15 & 1000 & 4468 & 7418 & 78.410 & 29.991 & 55.947 & 24.183 & 500 & 2077 & 3153 \\ \cmidrule[2pt]{1-11}
\end{tabular}
\end{adjustbox}
\caption{Statistics reported based on long-exposure images for all datasets. GT Img. stands for ground truth image count, where Leg. and Illeg. stand for legible and illegible text count, respectively.}
\label{table:dataset_stats}
\end{table*}

\begin{figure*}[!ht]
\centering
\begin{subfigure}{.32\linewidth}
    \centering
    \includegraphics[keepaspectratio=true, width=\textwidth]{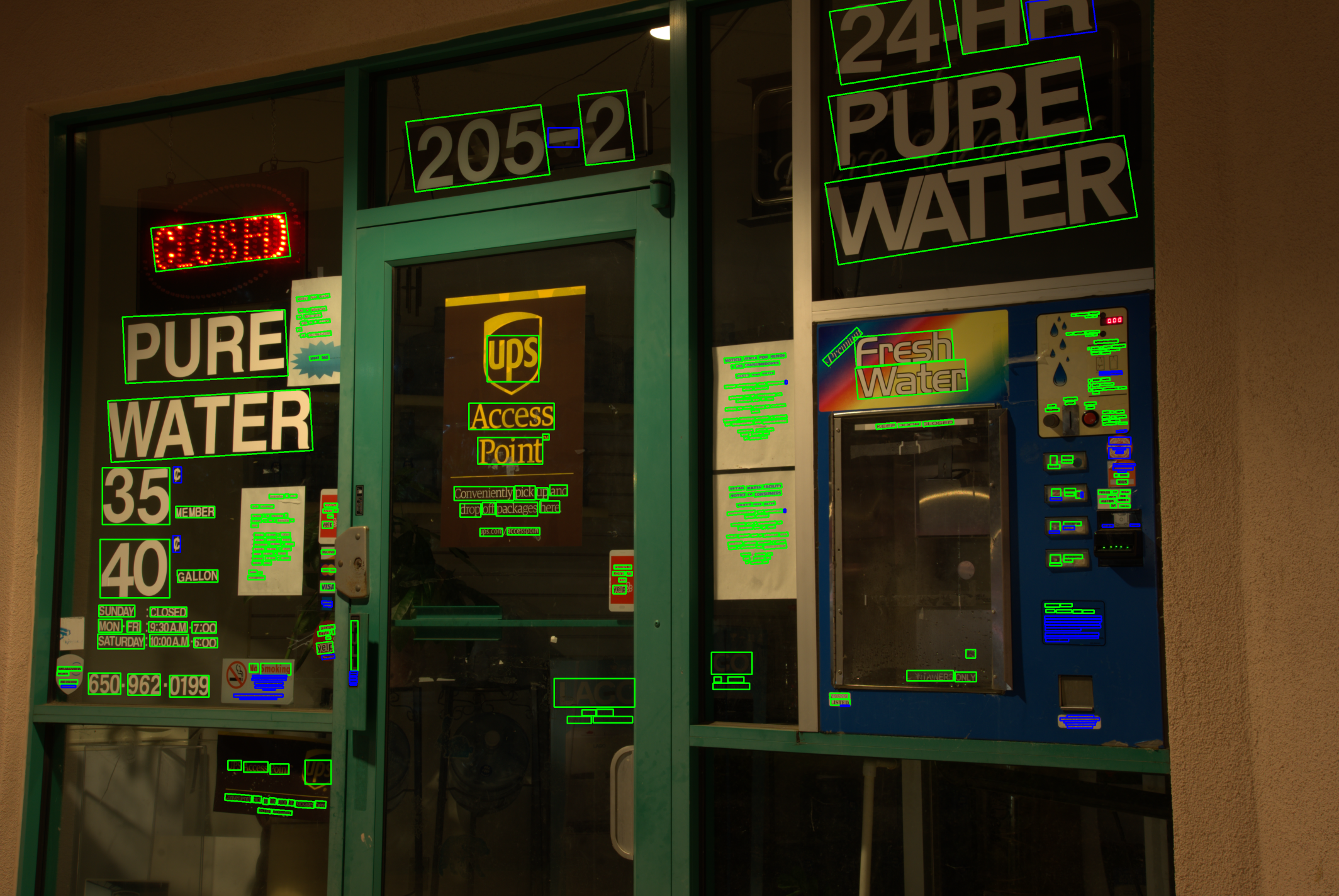}
    \caption{SID-Sony-Text}
\end{subfigure}
\begin{subfigure}{.32\linewidth}
    \centering
    \includegraphics[keepaspectratio=true, width=\textwidth]{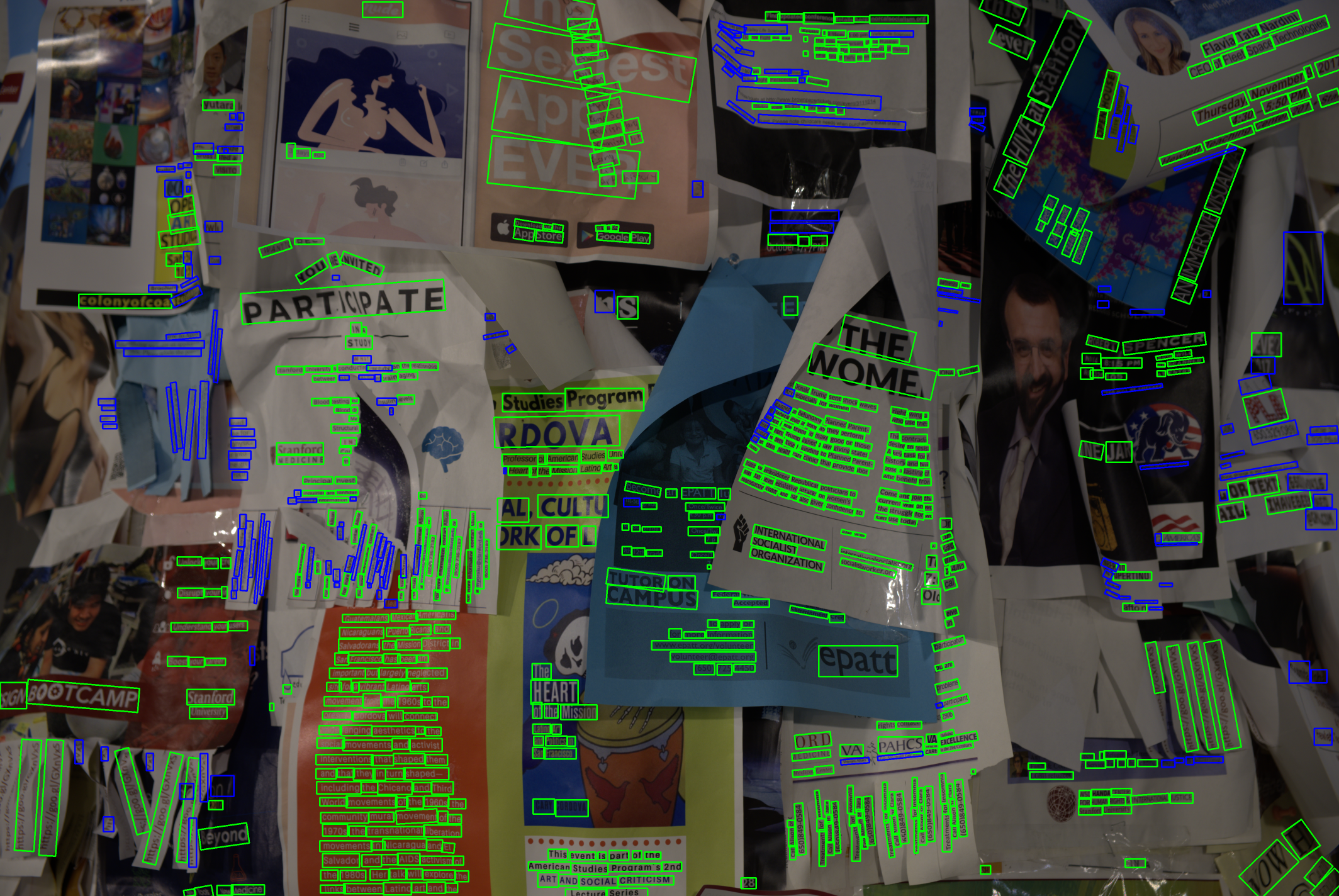}
    \caption{SID-Fuji-Text}
\end{subfigure}
\begin{subfigure}{.32\linewidth}
    \centering
    \includegraphics[keepaspectratio=true, width=\textwidth]{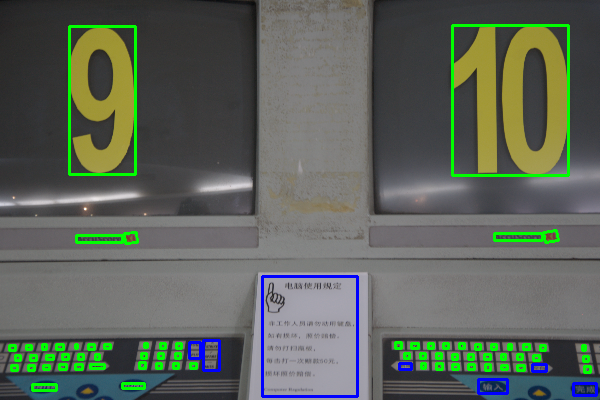}
    \caption{LOL-Text}
\end{subfigure}
\caption{Green boxes represent legible texts, and blue boxes represent illegible texts.}
\label{fig:dataset_samples}
\end{figure*}

In this work, we annotated all text instances in the extremely low-light dataset, SID~\cite{Chen2018LearningTS}, and the ordinary low-light dataset, LOL~\cite{Wei2018DeepRD}. SID has two subsets: SID-Sony, captured by Sony $\alpha$7S II, and SID-Fuji, captured by Fujifilm X-T2. For this work, we included 878/810 short-exposure images and 211/176 long-exposure images at a resolution of 4240×2832/6000×4000 from SID-Sony and SID-Fuji, respectively. The short-exposure time is 1/30, 1/25, and 1/10, while the corresponding reference (long-exposure) images were captured with 100 to 300 times longer exposure, i.e., 10 to 30 seconds. In our experiments, we converted short- and long-exposure SID images to RGB format. The LOL dataset provides low/normal-light image pairs taken from real scenes by controlling exposure time and ISO. There are 485 and 15 images at a resolution of 600×400 in the training and test sets, respectively. We closely annotated text instances in the SID and LOL datasets following the common IC15 standard. We show some samples in Figure \ref{fig:dataset_samples}. The newly annotated datasets are named SID-Sony-Text, SID-Fuji-Text, and LOL-Text to differentiate them from their low-light counterparts.

IC15 dataset was introduced in the ICDAR 2015 Robust Reading Competition for incidental scene text detection and recognition. It contains 1500 scene text images at a resolution of $1280\times720$. In this study, IC15 is primarily used to synthesize extremely low-light scene text images. Detailed statistics of the text annotations for SID-Sony-Text, SID-Fuji-Text, LOL-Text, and IC15 are shown in Table \ref{table:dataset_stats}, where we included the statistics for long-exposure images only for the sake of brevity. In this table, we also report relevant statistics of the mean and standard deviation of labeled texts' width and height to be used by the proposed Text-Aware Copy-Paste augmentation. The text annotations for SID-Sony-Text, SID-Fuji-Text, and LOL-Text datasets will be released at \href{https://github.com/chunchet-ng/Text-in-the-Dark}{https://github.com/chunchet-ng/Text-in-the-Dark}.

Moreover, we synthesized extremely low-light images based on IC15 by using U-Net and our proposed Supervised-DCE model, respectively. To study the difference between these two variations of image synthesis methods, we generated a total of four sets of images by using the aforementioned two models trained on SID-Sony and SID-Fuji, individually. Naming convention of such synthetic datasets follows the format of \enquote{\{Syn-IC15\}-\{Sony/Fuji\}-\{v1/v2\}}. \enquote{\{Sony/Fuji\}} is an indication of which dataset the image synthesis model is trained on, while \enquote{\{v1/v2\}} differentiates the image synthesis models where v1 is U-Net and v2 is our proposed Supervised-DCE model. For instance, the synthetic images generated by a U-Net trained on SID-Sony and SID-Fuji, are named Syn-IC15-Sony-v1 and Syn-IC15-Fuji-v1. And, synthetic images generated by our proposed Supervised-DCE model are denoted as Syn-IC15-Sony-v2 and Syn-IC15-Fuji-v2.

\section{Experimental Results}
\subsection{Experiment Setup}
\label{sec:exp_setup}
{\noindent\bf Datasets and Metrics.}
All low-light image enhancement methods are trained and tested on the datasets detailed in Section \ref{sec:datasets}. They are then evaluated in terms of intensity metrics (PSNR, SSIM), perceptual similarity (LPIPS), and text detection (H-Mean). For the SID-Sony-Text, SID-Fuji-Text, and LOL-Text datasets, which are annotated with text bounding boxes only, we used well-known and commonly used scene text detectors (CRAFT~\cite{Baek2019CharacterRA} and PAN~\cite{wang2019efficient}) to analyze the enhanced images. For IC15, which provides both text detection and text recognition labels, we conducted a two-stage text spotting experiment using the aforementioned text detectors (CRAFT, PAN) and two robust text recognizers (TRBA~\cite{trba2019} and ASTER~\cite{Aster2019}) on the synthesized IC15 images after enhancement. The metric for text spotting is case-insensitive word accuracy.

{\noindent\bf Implementation Details.}
We trained our image enhancement model for 4000 epochs using the Adam optimizer~\cite{Kingma2015AdamAM} with a batch size of 2. The initial learning rate is set to $1e^{-4}$ and decreased to $1e^{-5}$ after 2000 epochs. At each training iteration, we randomly cropped a $512\times512$ patch with at least one labeled text box inside and applied random flipping and image transpose as data augmentation strategies. The weightings of each loss term, i.e., $\omega_{recons}$, $\omega_{text}$, $\omega_{SSIM_{MS}}$, and $\omega_{edge}$, were empirically set to 0.2125, 0.425, 0.15, and 0.2125 respectively, following the work of ELIE\_STR~\cite{hsu2022extremely}. For other image enhancement methods, we re-trained them on all datasets using the best set of hyperparameters specified in their respective code repositories or papers.

As for the Supervised-DCE model, we used a batch size of 8 and trained for 200 epochs using the Adam optimizer with default parameters and a fixed learning rate of $1e^{-4}$. It was trained on $256\times256$ image patches with loss weightings of $\omega_{prox}$, $\omega_{spa}$, $\omega_{tv_{A}}$ and $\omega_{tv_{B}}$, set to 1, 20, 10, and 10 respectively.

\subsection{Results on SID-Sony-Text and SID-Fuji-Text Datasets}
\label{sec:SID_results}
Our model's performance is demonstrated in Table~\ref{table:sony_fuji_lol_results}, achieving the highest H-Mean scores on all datasets with CRAFT and PAN. Following \cite{hsu2022extremely}, we illustrate the CRAFT text detection results on SID-Sony-Text in Figure \ref{fig:sony_fig}. Qualitative results of existing methods on SID-Fuji-Text are presented in the supplementary material. The effectiveness of our model in enhancing extremely low-light images to a level where text can be accurately detected is readily apparent. In Figure~\ref{fig:sony_fig}, only the images enhanced by our proposed model yield accurate text detection results. On the other hand, existing methods generally produce noisier images, resulting in inferior text detection results. While GAN-enhanced images tend to be less noisy, the text regions are blurry, making text detection challenging. Moreover, our model achieves the highest PSNR and SSIM scores on both SID-Sony-Text and SID-Fuji-Text datasets, showing that our enhanced images are the closest to the image quality of ground truth images. In short, better text detection is achieved on our enhanced images through the improvement of overall image quality and preservation of fine details within text regions.


\begin{figure*}[!htb]
    \centering
    \begin{subfigure}[b]{0.24\textwidth}
    \begin{tabular}[b]{@{}l@{}}
        \includegraphics[width=\linewidth]{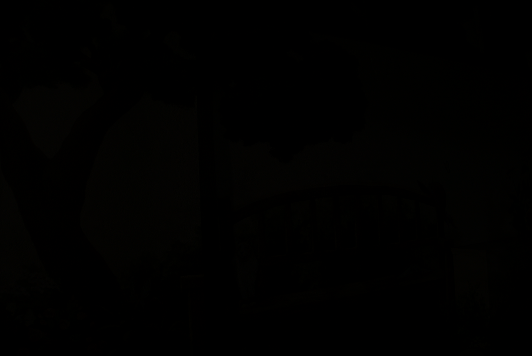}\\
        \includegraphics[width=\linewidth]{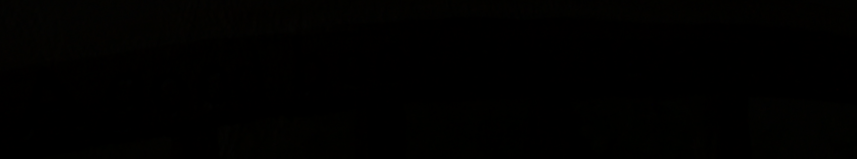}\\
     \end{tabular}
    \caption{Low-Light}
    \label{fig:sony_craft_input}
    \end{subfigure}\hspace{-0.5mm}
    \begin{subfigure}[b]{0.24\textwidth}
    \begin{tabular}[b]{@{}l@{}}
        \includegraphics[width=\linewidth]{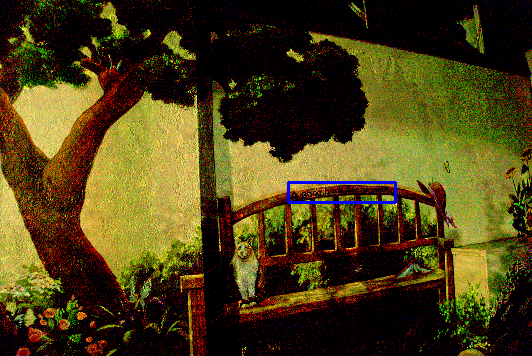}\\
        \includegraphics[width=\linewidth]{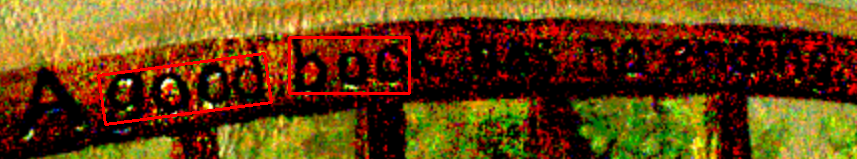}\\
     \end{tabular}
    \caption{LIME~\cite{Guo2017LIMELI}}
    \label{fig:sony_craft_lime}
    \end{subfigure}\hspace{-0.5mm}
    \begin{subfigure}[b]{0.24\textwidth}
    \begin{tabular}[b]{@{}l@{}}
        \includegraphics[width=\linewidth]{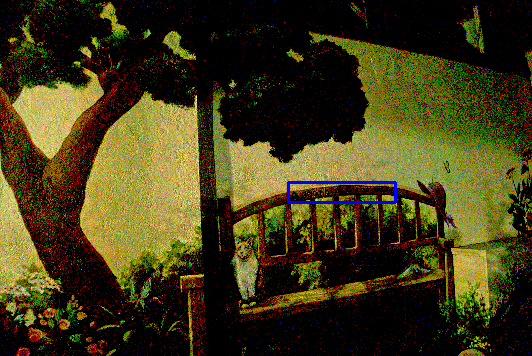}\\
        \includegraphics[width=\linewidth]{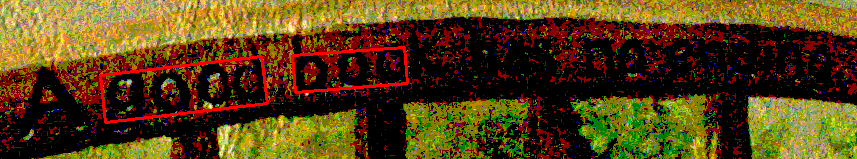}\\
     \end{tabular}
    \caption{BIMEF~\cite{Ying2017ABM}}
    \label{fig:sony_craft_bimef}
    \end{subfigure}\hspace{-0.5mm}
    \begin{subfigure}[b]{0.24\textwidth}
    \begin{tabular}[b]{@{}l@{}}
        \includegraphics[width=\linewidth]{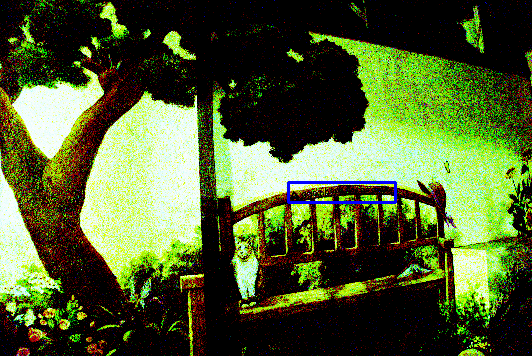}\\
        \includegraphics[width=\linewidth]{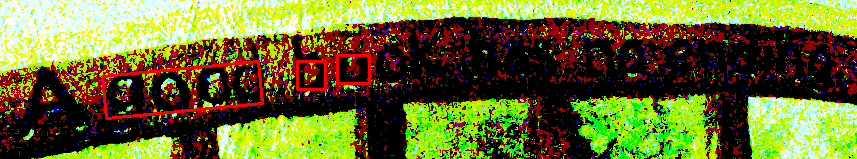}\\
     \end{tabular}
    \caption{Zero-DCE~\cite{Zero-DCE}}
    \label{fig:sony_craft_zero_dce}
    \end{subfigure}\hspace{-0.5mm}
    \\[1ex]
    \begin{subfigure}[b]{0.24\textwidth}
    \begin{tabular}[b]{@{}l@{}}
        \includegraphics[width=\linewidth]{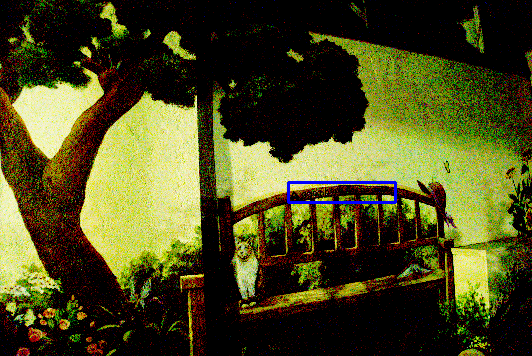}\\
        \includegraphics[width=\linewidth]{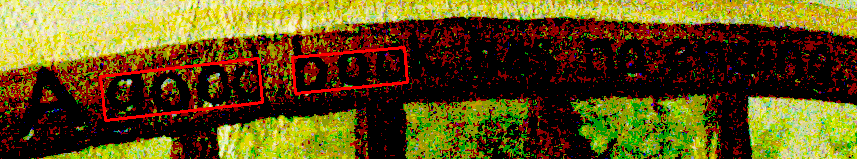}\\
     \end{tabular}
    \caption{Zero-DCE++~\cite{li2021learning}}
    \label{fig:sony_craft_zero_dce_pp}
    \end{subfigure}\hspace{-0.5mm}
    \begin{subfigure}[b]{0.24\textwidth}
    \begin{tabular}[b]{@{}l@{}}
        \includegraphics[width=\linewidth]{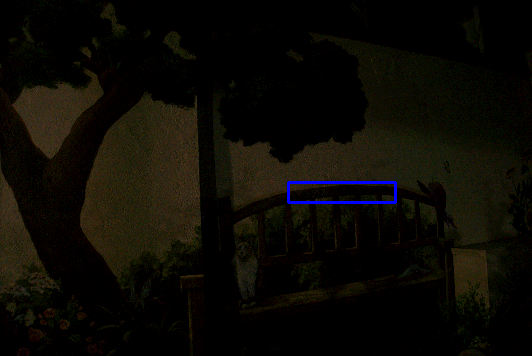}\\
        \includegraphics[width=\linewidth]{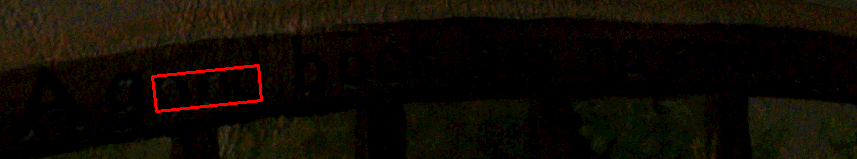}\\
     \end{tabular}
    \caption{SCI~\cite{ma2022toward}}
    \label{fig:sony_craft_sci}
    \end{subfigure}\hspace{-0.5mm}
    \begin{subfigure}[b]{0.24\textwidth}
    \begin{tabular}[b]{@{}l@{}}
        \includegraphics[width=\linewidth]{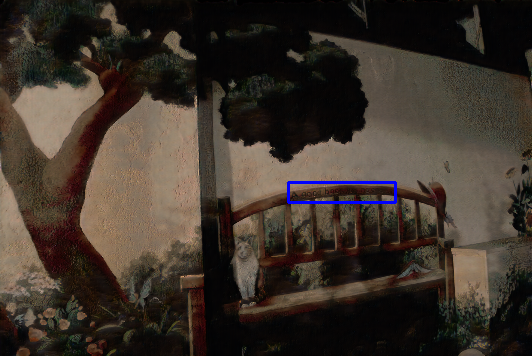}\\
        \includegraphics[width=\linewidth]{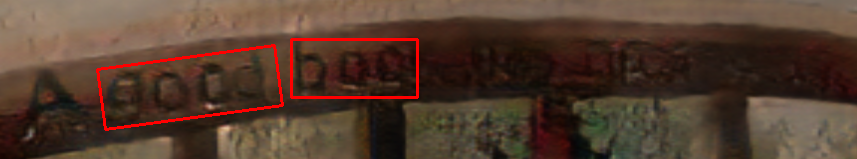}\\
     \end{tabular}
    \caption{CycleGAN~\cite{CycleGAN2017}}
    \label{fig:sony_craft_cyclegan}
    \end{subfigure}\hspace{-0.5mm}
    \begin{subfigure}[b]{0.24\textwidth}
    \begin{tabular}[b]{@{}l@{}}
        \includegraphics[width=\linewidth]{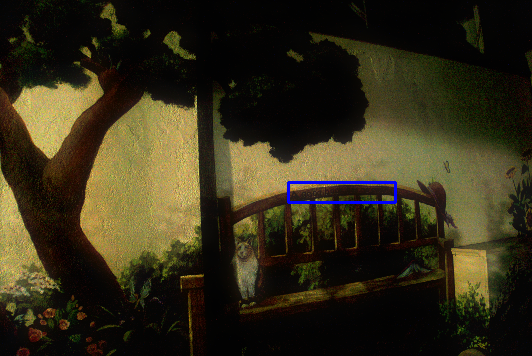}\\
        \includegraphics[width=\linewidth]{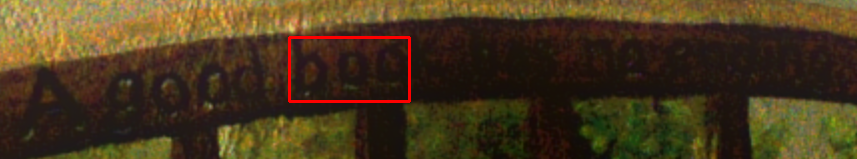}\\
     \end{tabular}
    \caption{EnlightenGAN~\cite{Jiang2019EnlightenGANDL}}
    \label{fig:sony_craft_enlightengan}
    \end{subfigure}\hspace{-0.5mm}
    \\[1ex]
    \begin{subfigure}[b]{0.24\textwidth}
    \begin{tabular}[b]{@{}l@{}}
        \includegraphics[width=\linewidth]{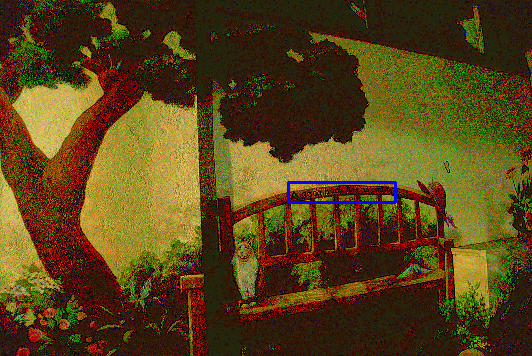}\\
        \includegraphics[width=\linewidth]{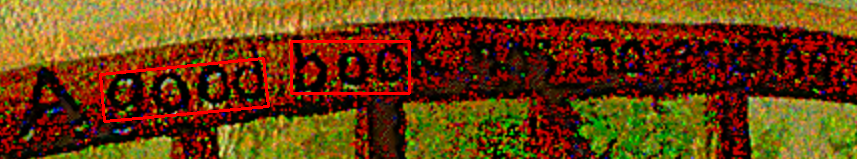}\\
     \end{tabular}
    \caption{RetinexNet~\cite{Wei2018DeepRD}}
    \label{fig:sony_craft_retinexnet}
    \end{subfigure}\hspace{-0.5mm}
    \begin{subfigure}[b]{0.24\textwidth}
    \begin{tabular}[b]{@{}l@{}}
        \includegraphics[width=\linewidth]{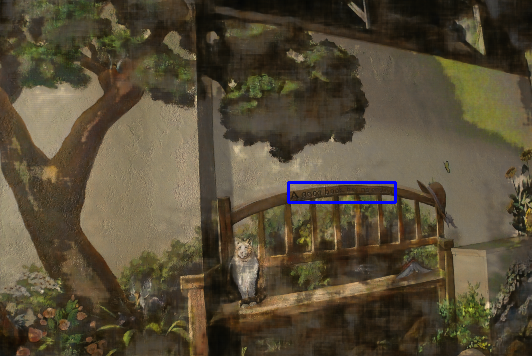}\\
        \includegraphics[width=\linewidth]{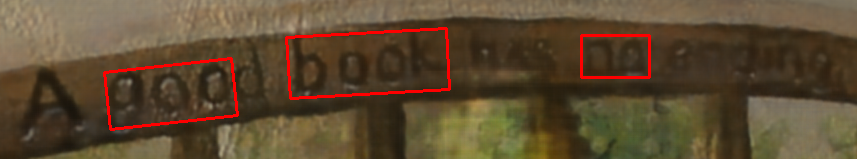}\\
     \end{tabular}
    \caption{Pix2Pix~\cite{isola2017image}}
    \label{fig:sony_craft_pix2pix}
    \end{subfigure}\hspace{-0.5mm}
    \begin{subfigure}[b]{0.24\textwidth}
    \begin{tabular}[b]{@{}l@{}}
        \includegraphics[width=\linewidth]{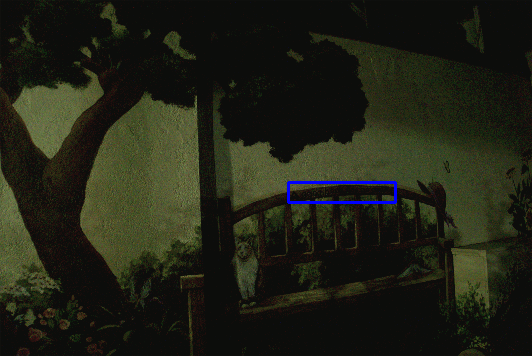}\\
        \includegraphics[width=\linewidth]{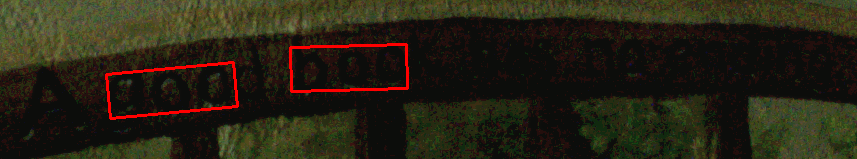}\\
     \end{tabular}
    \caption{ChebyLighter~\cite{pan2022chebylighter}}
    \label{fig:sony_craft_chebylighter}
    \end{subfigure}\hspace{-0.5mm}
    \begin{subfigure}[b]{0.24\textwidth}
    \begin{tabular}[b]{@{}l@{}}
        \includegraphics[width=\linewidth]{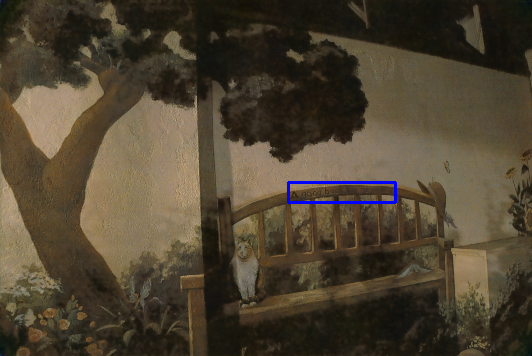}\\
        \includegraphics[width=\linewidth]{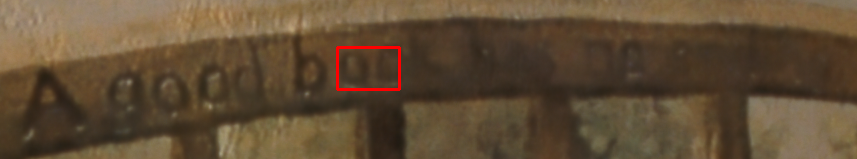}\\
     \end{tabular}
    \caption{FECNet~\cite{huang2022deep}}
    \label{fig:sony_craft_fecnet}
    \end{subfigure}\hspace{-0.5mm}
    \\[1ex]
    \begin{subfigure}[b]{0.24\textwidth}
    \begin{tabular}[b]{@{}l@{}}
        \includegraphics[width=\linewidth]{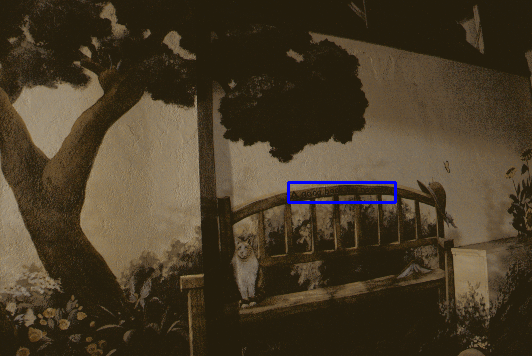}\\
        \includegraphics[width=\linewidth]{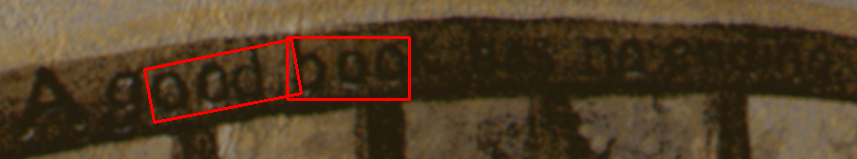}\\
     \end{tabular}
    \caption{IAT~\cite{cui2022you}}
    \label{fig:sony_craft_iat}
    \end{subfigure}\hspace{-0.5mm}
    \begin{subfigure}[b]{0.24\textwidth}
    \begin{tabular}[b]{@{}l@{}}
        \includegraphics[width=\linewidth]{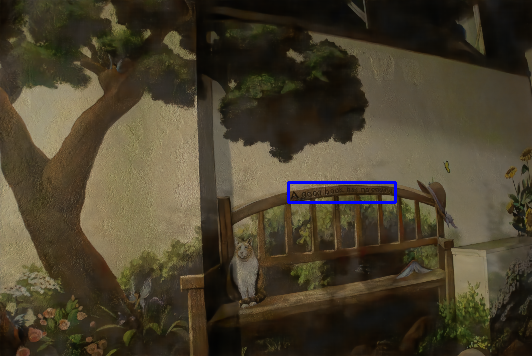}\\
        \includegraphics[width=\linewidth]{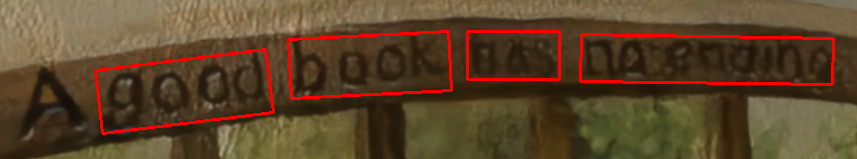}\\
     \end{tabular}
    \caption{ELIE\_STR~\cite{hsu2022extremely}}
    \label{fig:sony_craft_elie}
    \end{subfigure}\hspace{-0.5mm}
    \begin{subfigure}[b]{0.24\textwidth}
    \begin{tabular}[b]{@{}l@{}}
        \includegraphics[width=\linewidth]{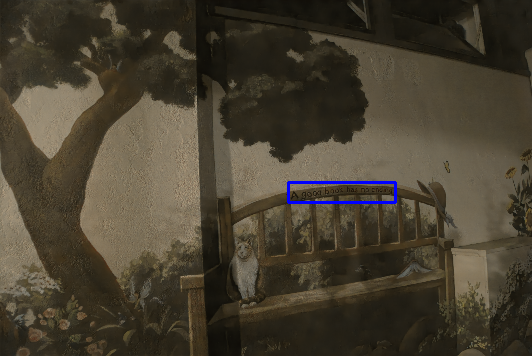}\\
        \includegraphics[width=\linewidth]{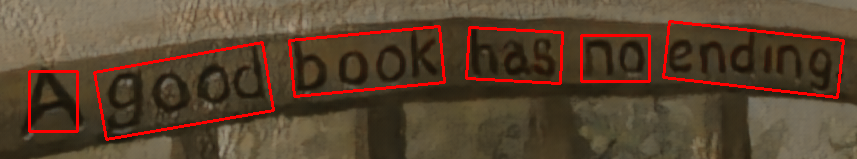}\\
     \end{tabular}
    \caption{Ours}
    \label{fig:sony_craft_ours}
    \end{subfigure}\hspace{-0.5mm}
    \begin{subfigure}[b]{0.24\textwidth}
    \begin{tabular}[b]{@{}l@{}}
        \includegraphics[width=\linewidth]{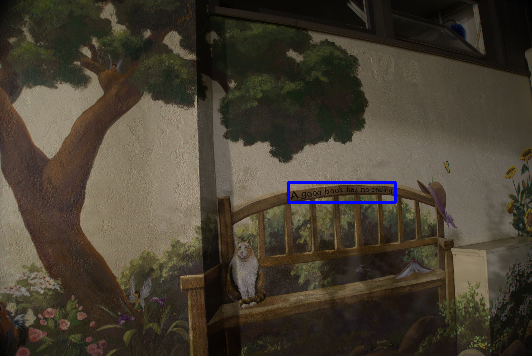}\\
        \includegraphics[width=\linewidth]{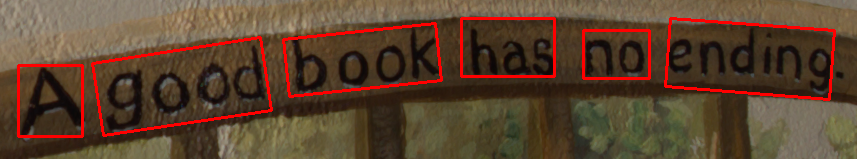}\\
     \end{tabular}
    \caption{Ground Truth}
    \label{fig:sony_craft_gt}
    \end{subfigure}
    \caption{Comparison with state-of-the-art methods on the SID-Sony-Text dataset is shown in the following manner: for each column, the first row displays enhanced images marked with blue boxes as regions of interest. The second row displays zoomed-in regions of enhanced images overlaid with red text detection boxes from CRAFT~\cite{Baek2019CharacterRA}. Column \ref{fig:sony_craft_input} displays the low-light image. Columns \ref{fig:sony_craft_lime} to \ref{fig:sony_craft_ours} show image enhancement results from all related methods. The last cell displays ground truth images.}
    \label{fig:sony_fig}
\end{figure*}

\begin{table*}[!htbp]
\setlength\tabcolsep{14pt}
\centering
\begin{adjustbox}{width=\linewidth,center}
\begin{tabular}{c?{0.8mm}llccccc}
\cmidrule[2pt]{1-8}
\multicolumn{1}{l}{\multirow{2}{*}{}} & \multirow{2}[2]{*}{Type} & \multirow{2}[2]{*}{Method} & \multicolumn{3}{c}{Image Quality} & \multicolumn{2}{c}{H-Mean} \\ \cmidrule(lr){4-6} \cmidrule(lr){7-8}
\multicolumn{1}{l}{} & & & PSNR $\uparrow$ & SSIM $\uparrow$ & LPIPS $\downarrow$ & CRAFT $\uparrow$ & PAN $\uparrow$ \\ \cmidrule[2pt]{1-8}

\multirow{16}{*}{\rotatebox[origin=c]{90}{\textbf{SID-Sony-Text}}} & & Input & - & - & - & 0.057 & 0.026 \\ \cline{2-8}
& \multirow{2}{*}{TRAD} & LIME~\cite{Guo2017LIMELI} & 13.870 & 0.135 & 0.873 & 0.127 & 0.057 \\
& & BIMEF~\cite{Ying2017ABM} & 12.870 & 0.110 & 0.808 & 0.136 & 0.079 \\ \cline{2-8}
& \multirow{3}{*}{ZSL} & Zero-DCE~\cite{Zero-DCE} & 10.495 & 0.080 & 0.999 & 0.196 & 0.157 \\
& & Zero-DCE++~\cite{li2021learning} & 12.368 & 0.076 & 0.982 & 0.218 & 0.162 \\ 
& & SCI~\cite{ma2022toward} & 11.814 & 0.100 & 1.000 & 0.201 & 0.151 \\ \cline{2-8}
& \multirow{2}{*}{UL} & CycleGAN~\cite{CycleGAN2017} & 15.340 & 0.453 & 0.832 & 0.090 & 0.053 \\
& & EnlightenGAN~\cite{Jiang2019EnlightenGANDL} & 14.590 & 0.426 & 0.793 & 0.146 & 0.075 \\ \cline{2-8}
& \multirow{7}{*}{SL} & RetinexNet~\cite{Wei2018DeepRD} & 15.490 & 0.368 & 0.785 & 0.115 & 0.040 \\
& & Pix2Pix~\cite{isola2017image} & 21.070 & 0.662 & 0.837 & 0.266 & 0.190 \\
& & ChebyLighter~\cite{pan2022chebylighter} & 15.418 & 0.381 & 0.787 & 0.260 & 0.184 \\
& & FECNet~\cite{huang2022deep} & 22.575 & 0.648 & 0.788 & 0.245 & 0.188 \\
& & IAT~\cite{cui2022you} & 19.234 & 0.562 & 0.778 & 0.244 & 0.176 \\
& & ELIE\_STR~\cite{hsu2022extremely} & 25.507 & 0.716 & 0.789 & 0.324 & 0.266 \\
& & Ours & \textbf{25.596} & \textbf{0.751} & \textbf{0.751} & \textbf{0.368} & \textbf{0.298} \\ \cline{2-8}
& & GT & - & - & - & 0.842 & 0.661 \\ \cmidrule[2pt]{1-8}

\multicolumn{8}{c}{} \\ \cmidrule[2pt]{1-8}
\multirow{13}{*}{\rotatebox[origin=c]{90}{\textbf{SID-Fuji-Text}}} & & Input & - & - & - & 0.048 & 0.005 \\ \cline{2-8}
& \multirow{3}{*}{ZSL} & Zero-DCE~\cite{Zero-DCE} & 8.992 & 0.035 & 1.228 & 0.249 & 0.061 \\
& & Zero-DCE++~\cite{li2021learning} & 11.539 & 0.047 & 1.066 & 0.262 & 0.077 \\
& & SCI~\cite{ma2022toward} & 10.301 & 0.056 & 1.130 & 0.300 & 0.073 \\ \cline{2-8}
& \multirow{2}{*}{UL} & CycleGAN~\cite{CycleGAN2017} & 17.832 & 0.565 & 0.735 & 0.277 & 0.191 \\
& & EnlightenGAN~\cite{Jiang2019EnlightenGANDL} & 18.834 & 0.572 & 0.822 & 0.310 & 0.277 \\ \cline{2-8}
& \multirow{7}{*}{SL} & Pix2Pix~\cite{isola2017image} & 19.601 & 0.599 & 0.803 & 0.353 & 0.296 \\
& & ChebyLighter~\cite{pan2022chebylighter} & 20.313 & 0.616 & 0.791 & 0.412 & 0.318 \\
& & FECNet~\cite{huang2022deep} & 18.863 & 0.365 & 0.829 & 0.382 & 0.185 \\
& & IAT~\cite{cui2022you} & 19.647 & 0.537 & 0.844 & 0.445 & 0.277 \\
& & ELIE\_STR~\cite{hsu2022extremely} & 19.816 & 0.614 & 0.801 & 0.426 & 0.333 \\ 
& & Ours & \textbf{21.880} & \textbf{0.649} & \textbf{0.788} & \textbf{0.487} & \textbf{0.356} \\ \cline{2-8}
& & GT & - & - & - & 0.775 & 0.697 \\ \cmidrule[2pt]{1-8}

\multicolumn{8}{c}{} \\ \cmidrule[2pt]{1-8}
\multirow{13}{*}{\rotatebox[origin=c]{90}{\textbf{LOL-Text}}} & & Input & - & - & - & 0.333 & 0.133 \\ \cline{2-8}
& \multirow{3}{*}{ZSL} & Zero-DCE~\cite{Zero-DCE} & 14.928 & 0.587 & 0.328 & 0.421 & 0.229 \\
& & Zero-DCE++~\cite{li2021learning} & 15.829 & 0.537 & 0.408 & 0.389 & 0.242 \\
& & SCI~\cite{ma2022toward} & 14.835 & 0.549 & 0.335 & 0.421 & 0.171 \\ \cline{2-8}
& \multirow{2}{*}{UL} & CycleGAN~\cite{CycleGAN2017} & 19.826 & 0.734 & 0.288 & 0.250 & 0.133 \\
& & EnlightenGAN~\cite{Jiang2019EnlightenGANDL} & 15.800 & 0.654 & 0.300 & 0.343 & 0.125 \\ \cline{2-8}
& \multirow{7}{*}{SL} & Pix2Pix~\cite{isola2017image} & 20.581 & 0.771 & 0.247 & 0.353 & 0.129 \\
& & ChebyLighter~\cite{pan2022chebylighter} & 19.820 & 0.769 & 0.199 & 0.353 & 0.176 \\
& & FECNet~\cite{huang2022deep} & 20.432 & 0.787 & 0.231 & 0.378 & 0.229 \\
& & IAT~\cite{cui2022you} & 20.437 & 0.772 & 0.234 & 0.421 & 0.188 \\
& & ELIE\_STR~\cite{hsu2022extremely} & 19.782 & 0.824 & 0.167 & 0.462 & 0.235 \\
& & Ours & \textbf{21.330} & \textbf{0.828} & \textbf{0.163} & \textbf{0.474} & \textbf{0.294} \\ \cline{2-8}
& & GT & - & - & - & 0.439 & 0.205 \\ \cmidrule[2pt]{1-8}
 
\end{tabular}
\end{adjustbox}

\caption{Quantitative results of PSNR, SSIM, LPIPS, and text detection H-Mean for low-light image enhancement methods on SID-Sony-Text, SID-Fuji-Text, and LOL-Text datasets. Please note that TRAD, ZSL, UL, and SL stand for traditional methods, zero-shot learning, unsupervised learning, and supervised learning respectively. Scores in bold are the best of all.}
\label{table:sony_fuji_lol_results}
\end{table*}

\subsection{Results on LOL-Text Dataset}
\label{sec:LOL_results} 
To demonstrate the effectiveness of our model in enhancing low-light images with varying levels of darkness, we conducted experiments on the widely used LOL dataset, which is relatively brighter than the SID dataset, as depicted in Table \ref{table:SID_LOL_black_level_comparison}. Interestingly, we found that our enhanced images achieved the best detection results on LOL-Text among existing methods, as shown in Table \ref{table:sony_fuji_lol_results}. Surprisingly, despite the lower resolution (600x400) of the images in LOL, our method’s enhanced images with sharper and crisper low-level details surpassed the ground truth images’ H-Mean scores. Qualitative results on the LOL-Text dataset are illustrated in the supplementary material. Although certain methods yielded output images with acceptable image quality (i.e., bright images without color shift), their text detection results were inferior to ours. Furthermore, our superior results on the LOL-Text dataset emphasize our method’s ability to generalize well on both ordinary and extremely low-light images, effectively enhancing a broader range of low-light images while making the text clearly visible.

\subsection{Effectiveness of the Proposed Supervised-DCE Model}
The goal of image synthesis in our work is to translate images captured in well-illuminated scenarios to extremely low light. In this work, we choose the commonly used IC15 scene text dataset as our main synthesis target. The synthesized dataset then serves as additional data to train better scene text-aware image enhancement models, which are studied in Section \ref{exp:mixing_results}.

Intuitively, realistic synthesized images should possess similar characteristics to genuine extremely low-light images. To verify the effectiveness of our synthesis model, we compared our proposed Supervised-DCE model (v2) with the U-Net proposed in SID~\cite{Chen2018LearningTS} (v1). Specifically, we trained the synthesizing models on the training set and synthesized the images based on the corresponding test set. Then, we calculated the PSNR and SSIM of the synthesized images by comparing them with the genuine ones along with the average perceptual lightness in CIELAB color space. The comparison was made on two SID datasets, SID-Sony and SID-Fuji.

In Table~\ref{table:det_realism}, we show that v2's PSNR and SSIM are higher than v1's, indicating higher similarity between our synthesized and genuine images. Our new method (v2) also exhibits closer Avg. L* values and H-Mean scores to the genuine images than v1, indicating darker and more accurate deterioration of fine text details. In addition, qualitative results for the proposed Supervised-DCE model and results of synthetic IC15 datasets including Syn-IC15-Sony-v1, Syn-IC15-Sony-v2, Syn-IC15-Fuji-v1, and Syn-IC15-Fuji-v2 are presented in the supplementary material for comprehensive analyses. 

\begin{table*}[!ht]
\centering
\begin{adjustbox}{width=0.7\linewidth,center}
\begin{tabular}{lccccc}
\cmidrule[2pt]{1-6}
Dataset & PSNR & SSIM & Avg. L* & CRAFT & PAN  \\ \cmidrule[2pt]{1-6}
Syn-SID-Sony-v1 & 41.095 & 0.809 & 0.176 & 0.294 & 0.083 \\
Syn-SID-Sony-v2 & 45.442 & 0.942 & 0.003 & 0.135 & 0.014 \\ \hline
Genuine SID-Sony & - & - & 0.008 & 0.057 & 0.026 \\
\cmidrule[2pt]{1-6}
Syn-SID-Fuji-v1 & 39.187 & 0.784 & 0.172 & 0.402 & 0.042 \\
Syn-SID-Fuji-v2 & 41.881 & 0.863 & 0.002 & 0.093 & 0.002 \\\hline
Genuine SID-Fuji & - & - & 0.004 & 0.048 & 0.005 \\
 \cmidrule[2pt]{1-6}
\end{tabular}
\end{adjustbox}
\caption{The difference between genuine extremely low-light dataset, SID, and synthetic extremely low-light images generated using U-Net (v1) and Supervised-DCE (v2). Please note that synthetic images' PSNR and SSIM values are based on comparison against genuine low-light images in the test set instead of pure black images calculated in Table \ref{table:SID_LOL_black_level_comparison}. Additionally, we can notice that v2-images are more realistic and darker, similar to genuine extremely low-light images due to their higher values of PSNR and SSIM, along with closer Avg. L*.}
\label{table:det_realism}
\end{table*}

\subsection{Results on Training with Mixed Datasets}
\label{exp:mixing_results}
We trained top-performing models from Section~\ref{sec:SID_results} using a mixture of genuine (SID) and synthetic low-light (IC15) datasets to test whether extremely low-light image enhancement can benefit from synthesized images. The trained models were evaluated on their respective genuine low-light datasets. Results in Table \ref{table:mix_data_sony_n_Fuji} showed a significant increase in H-Mean, and we found that both versions (v1 and v2) can fill the gap caused by the scarcity of genuine low-light images. This justifies the creation of a synthetic IC15 dataset for such a purpose. Furthermore, v2-images, i.e., extremely low-light images synthesized by our proposed Supervised-DCE, further pushed the limit of H-mean scores on genuine extremely low-light images, and our enhancement model benefited the most because it could learn more from text instances and reconstruct necessary details to represent texts. Despite our method's success, a noticeable gap exists between our results and the ground truth, emphasizing the need for further research and development to achieve even more accurate and reliable scene text extraction in low-light conditions.

\begin{table*}[!ht]
\centering
\begin{adjustbox}{width=\linewidth,center}
\begin{tabular}{llcccccccc}
\cmidrule[2pt]{1-10}
\multirow{2}{*}{Type} & \multirow{2}{*}{Method} & 
\multicolumn{2}{c}{\begin{tabular}[c]{@{}c@{}}SID-Sony-Text +\\ Syn-IC15-Sony-v1\end{tabular}} & \multicolumn{2}{c}{\begin{tabular}[c]{@{}c@{}}SID-Sony-Text +\\ Syn-IC15-Sony-v2\end{tabular}} & \multicolumn{2}{c}{\begin{tabular}[c]{@{}c@{}}SID-Fuji-Text +\\ Syn-IC15-Fuji-v1\end{tabular}} & \multicolumn{2}{c}{\begin{tabular}[c]{@{}c@{}}SID-Fuji-Text + \\ Syn-IC15-Fuji-v2\end{tabular}} \\ \cmidrule(lr){3-4} \cmidrule(lr){5-6} \cmidrule(lr){7-8} \cmidrule(lr){9-10}

& & CRAFT $\uparrow$ & PAN $\uparrow$ & CRAFT $\uparrow$ & PAN $\uparrow$ & CRAFT $\uparrow$ & PAN $\uparrow$ & CRAFT $\uparrow$ & PAN $\uparrow$ \\ \cmidrule[2pt]{1-10}

& Input & 0.057 & 0.026 & 0.057& 0.026 & 0.048& 0.005 & 0.048 & 0.005\\ \hline
\multirow{2}{*}{ZSL} & Zero-DCE++~\cite{li2021learning} & 0.230 & 0.159 & 0.242 & 0.153 & 0.274 & 0.080 & 0.281 & 0.076 \\
& SCI~\cite{ma2022toward} & 0.240 & 0.154 & 0.243 & 0.160 & 0.307 & 0.076 & 0.313 & 0.084 \\ \hline
\multirow{2}{*}{UL} & CycleGAN~\cite{CycleGAN2017} & 0.180 & 0.071 & 0.219 & 0.143 & 0.297 & 0.284 & 0.310 & 0.277 \\ 
& EnlightenGAN~\cite{Jiang2019EnlightenGANDL} & 0.205 & 0.146 & 0.237 & 0.163 & 0.329 & 0.246 & 0.342 & 0.282 \\ \hline
\multirow{2}{*}{SL} & ELIE\_STR~\cite{hsu2022extremely} & 0.348 & 0.278 & 0.361 & 0.296 & 0.444 & 0.359 & 0.466 & 0.375 \\ 
& Ours & \textbf{0.383} & \textbf{0.311} & \textbf{0.395} & \textbf{0.319} & \textbf{0.515} & \textbf{0.392} & \textbf{0.549} & \textbf{0.416} \\ \hline
& GT & 0.842 & 0.661 & 0.842 & 0.661 & 0.775 & 0.697 & 0.775 & 0.697 \\ \cmidrule[2pt]{1-10}

\end{tabular}
\end{adjustbox}
\caption{Text detection H-Mean on genuine extremely low-light datasets when trained on a combination of genuine and synthetic datasets. Scores in bold are the best of all.}
\label{table:mix_data_sony_n_Fuji}
\end{table*}

\subsection{Ablation Study of Proposed Modules}
\begin{table*}[!htbp]
\centering
\begin{adjustbox}{width=\linewidth,center}
\begin{tabular}{ccccccccc}
    \cmidrule[2pt]{1-9}
     \multicolumn{4}{c}{Proposed Modules} & \multicolumn{3}{c}{Image Quality} & \multicolumn{2}{c}{H-Mean} \\ \cmidrule(lr){1-4} \cmidrule(lr){5-7} \cmidrule(lr){8-9} 
    Text-CP & Dual Encoder & Edge-Att & Edge Decoder & PSNR $\uparrow$ & SSIM $\uparrow$ & LPIPS $\downarrow$ & CRAFT $\uparrow$ & PAN $\uparrow$ \\ \cmidrule[2pt]{1-9}
    - & - & - & - & 21.847 & 0.698 & 0.783 & 0.283 & 0.205 \\ 
    \checkmark & - & - & - & 21.263 & 0.658 & 0.771 & 0.304 & 0.252 \\
    \checkmark & \checkmark & - & - & 20.597 & 0.655 & 0.780 & 0.335 & 0.261 \\
    \checkmark & \checkmark & \checkmark & - & 21.440 & 0.669 & 0.776 & 0.342 & 0.256 \\
    \checkmark & \checkmark & - & \checkmark & 21.588 & 0.674 & 0.779 & 0.353 & 0.285 \\
    \checkmark & - & \checkmark & \checkmark & 23.074 & 0.712 & 0.783 & 0.350 & 0.281 \\
    - & \checkmark & \checkmark & \checkmark & 24.192 & 0.738 & 0.784 & 0.356 & 0.292 \\
    \checkmark & \checkmark & \checkmark & \checkmark & \textbf{25.596} & \textbf{0.751} & \textbf{0.751} & \textbf{0.368} & \textbf{0.298} \\ \cmidrule[2pt]{1-9}
\end{tabular}
\end{adjustbox}
\caption{Ablation study of proposed modules in terms of PSNR, SSIM, LPIPS, and text detection H-Mean on the SID-Sony-Text dataset. Scores in bold are the best of all.}
\label{table:det_ablation_study}
\end{table*}

To understand the effect of each component of our model, we conducted several ablation experiments by either adding or removing them one at a time. Results are presented in Table \ref{table:det_ablation_study}. The baseline was a plain U-Net without any proposed modules. We initiated the ablation study by adding Text-CP data augmentation, which improved CRAFT H-Mean from 0.283 to 0.304, indicating that involving more text instances during training is relevant to text-aware image enhancement for models to learn text representation. Moreover, scores increased steadily by gradually stacking the baseline with more modules. For instance, with the help of the dual encoder structure and Edge-Att module in our proposed framework, CRAFT H-Mean increased from 0.304 to 0.342. This shows that they can extract image features better and attend to edges that shape texts in enhanced images. The edge reconstruction loss calculated based on predictions from the edge decoder helped strengthen the learning of edge features and empowered encoders in our model. Interestingly, we found that removing one of the two most representative modules (i.e., dual encoder or Edge-Att module) led to significant differences in H-Mean because these two modules' designs allow them to extract and attend to significant image features independently. We concluded the experiment by showing that combining all proposed modules led to the highest scores, as each module played an integral role in our final network. Further analysis of Edge-Att and Text-CP are included in the supplementary material to study their effectiveness as compared to the original versions.


\section{Conclusion}
This paper presents a novel scene text-aware extremely low-light image enhancement framework consisting of a Text-Aware Copy-Paste augmentation method as a pre-processing step, followed by a new dual-encoder-decoder architecture armed with Edge-Aware attention modules. With further assistance from text detection and edge reconstruction losses, our model can enhance images to the extent that high-level perceptual reasoning tasks can be better fulfilled. Extremely low-light image synthesis has rarely been discussed over the years. Thus, we proposed a novel Supervised-DCE model to provide better synthesized extremely low-light images so that extremely low-light image enhancement can benefit from publicly available scene text datasets such as IC15. Furthermore, our proposed extremely low-light image enhancement model has been rigorously evaluated against various competing methods, including traditional techniques and deep learning-based approaches, on challenging datasets such as SID-Sony-Text, SID-Fuji-Text, LOL-Text, and synthetic IC15. Through extensive experimentation, our findings consistently demonstrate our model's superiority in extremely low-light image enhancement and text extraction tasks.

\clearpage
\bibliographystyle{elsarticle-num} 
\bibliography{egbib}
\end{document}


\begin{frontmatter}



\title{Supplementary Material: Text in the Dark: Extremely Low-Light Text Image Enhancement}
\author[chalmers]{Che-Tsung Lin\fnref{equal}}
\author[um]{Chun Chet Ng\fnref{equal}}
\author[um]{Zhi Qin Tan}
\author[um]{Wan Jun Nah}
\author[adelaide]{Xinyu Wang}
\author[um]{Jie Long Kew}
\author[nthu]{Pohao Hsu}
\author[nthu]{Shang Hong Lai}
\author[um]{Chee Seng Chan\corref{cor1}}
\ead{cs.chan@um.edu.my}
\author[chalmers]{Christopher Zach}

\fntext[equal]{These authors contributed equally to this work.}
\cortext[cor1]{Corresponding Author.}

\affiliation[chalmers]{organization={Chalmers University of Technology},
            state={Gothenburg},
            country={Sweden}}
            
\affiliation[um]{organization={Universiti Malaya},
            state={Kuala Lumpur},
            country={Malaysia}}
            
\affiliation[nthu]{organization={National Tsing Hua University},
            state={Hsinchu},
            country={Taiwan}}

\affiliation[adelaide]{organization={The University of Adelaide},
            state={Adelaide},
            country={Australia}}
            
\begin{abstract}
This supplementary material presents the pseudocode for Text-CP, as well as qualitative results for the SID-Fuji-Text and LOL-Text datasets and the proposed Supervised-DCE Model. We also provide both quantitative and qualitative results for the Synthesized IC15 dataset. Additionally, we analyze the effectiveness of our proposed Edge-Att and Text-CP methods. Overall, our proposed method outperforms existing approaches on all datasets.

\end{abstract}
\end{frontmatter}

\section{Pseudocode of Text-CP}
Algorithm~\ref{cap_algo} shows the detailed pseudocode of the proposed Text-CP.

\begin{spacing}{0.8}
\begin{algorithm}[t]	
    \caption{Text-Aware Copy-Paste Algorithm}	
    \label{cap_algo}	
    \begin{algorithmic}[1]	
    \Require{$C=\left \{ (u_1,v_1,w_1,h_1), ...,(u_{\left | \mathit{\emph{C}}  \right |},v_{\left | \mathit{\emph{C}}  \right |},w_{\left | \mathit{\emph{C}}  \right |},h_{\left | \mathit{\emph{C}}  \right |})) \right \}$: All text instances in dataset, $t$: Training image to be augmented, $C_t \in C$: Text instances in $t$, $\gamma$: Minimum width-to-height ratio of text instances, $n_{\text{target}}$: Target number of text instances in an image}
            \While{$|C_t| < n_{\text{target}}$}
                \Comment{i. Sample text instance not in image $t$}
                \State $bbox(u,v,w,h) \sim (C - C_{t})$ 
                \State $\hat{u} \sim U(0, w_t)$ \Comment{ii. Dataset-aware sampling}
                \State $\hat{v} \sim U(0, h_t)$
                \State $\hat{w} \sim \mathcal{N}(\mu_W, \sigma_W^2)$
                \State $\hat{h} \sim \mathcal{N}(\mu_H, \sigma_H^2)$
                \If{$\frac{\hat{w}}{\hat{h}} \ge \gamma \And (\hat{u}+\hat{w}) \le w_t \And (\hat{v} + \hat{h}) \le h_t$ }
                    \State Resize $bbox(u,v,w,h)$ to $bbox(\hat{u},\hat{v},\hat{w},\hat{h})$ \Comment{iii. Resize text instance}
                \EndIf
                \If{$bbox(\hat{u},\hat{v}, \hat{w}, \hat{h})$ does not overlap with all boxes in $C_t$}
                    \State Paste $bbox$ on $t$ at $(\hat{u},\hat{v})$ \Comment{iv. Paste text instance}
                    \State Add $bbox$ into $C_t$	
                \EndIf	
            \EndWhile
    \end{algorithmic}	
\end{algorithm}
\end{spacing}

\section{Qualitative Results of Existing Methods on the SID-Fuji-Text and LOL-Text}
\begin{figure*}[!htbp]
    \centering
    \begin{subfigure}[b]{0.19\textwidth}
    \begin{tabular}[b]{@{}l@{}}
        \includegraphics[width=\linewidth]{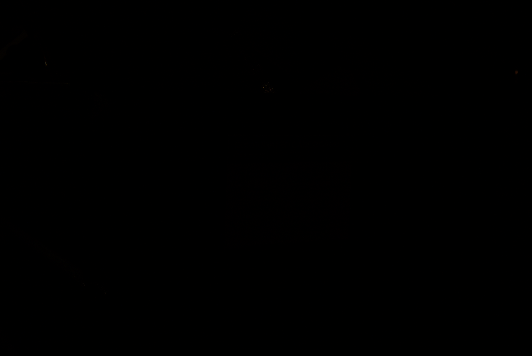}\\
        \includegraphics[width=\linewidth, height=2.3cm]{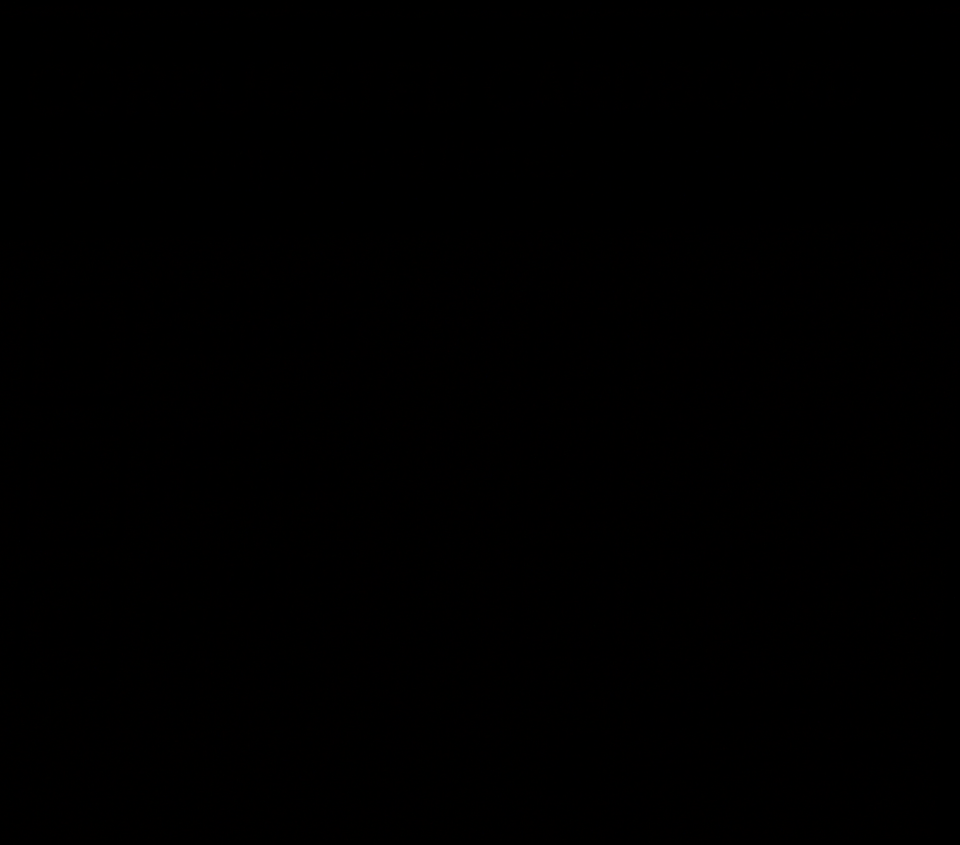}\\
     \end{tabular}
    \caption{Low-Light}
    \label{fig:fuji_craft_input}
    \end{subfigure}\hspace{-0.5mm}
    \begin{subfigure}[b]{0.19\textwidth}
    \begin{tabular}[b]{@{}l@{}}
        \includegraphics[width=\linewidth]{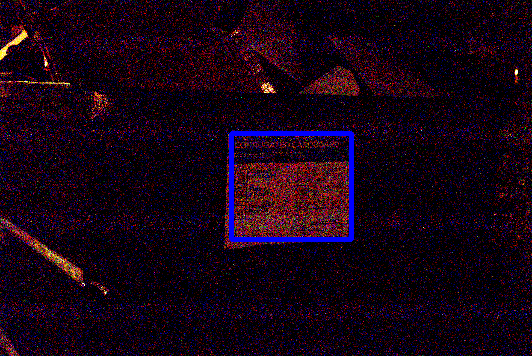}\\
        \includegraphics[width=\linewidth, height=2.3cm]{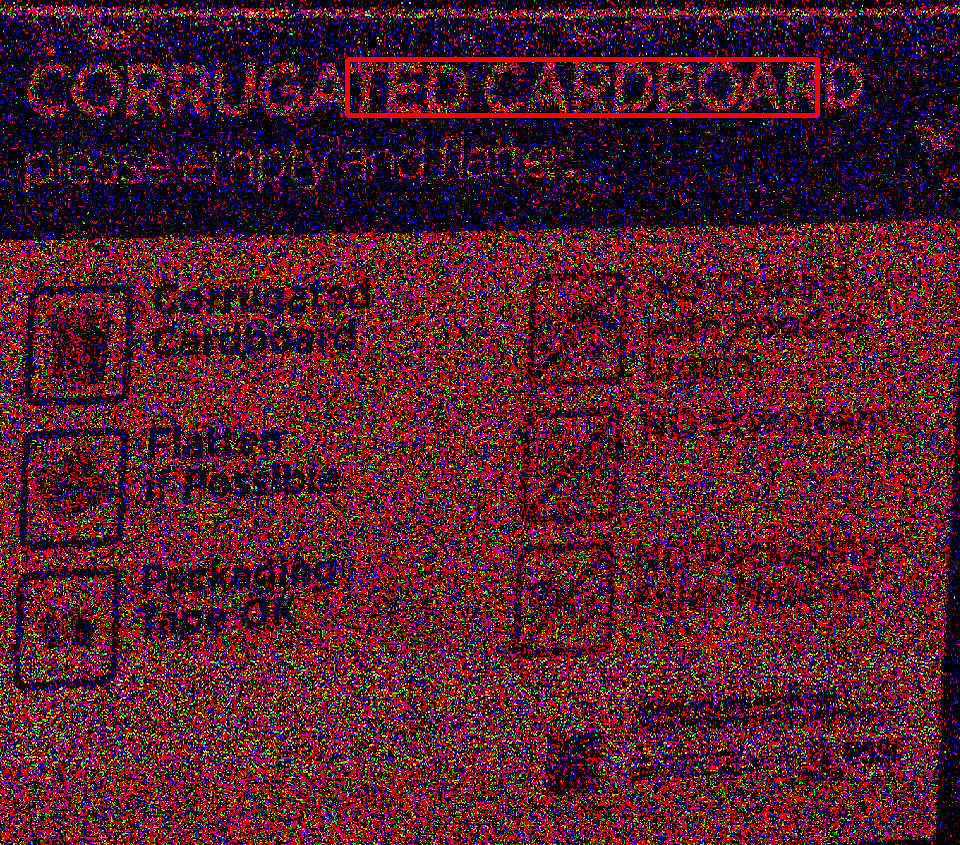}\\
     \end{tabular}
    \caption{Zero-DCE~\cite{Zero-DCE}}
    \label{fig:fuji_craft_zero_dce}
    \end{subfigure}\hspace{-0.5mm}
    \begin{subfigure}[b]{0.205\textwidth}
    \begin{tabular}[b]{@{}l@{}}
        \includegraphics[width=\linewidth, height=1.74cm]{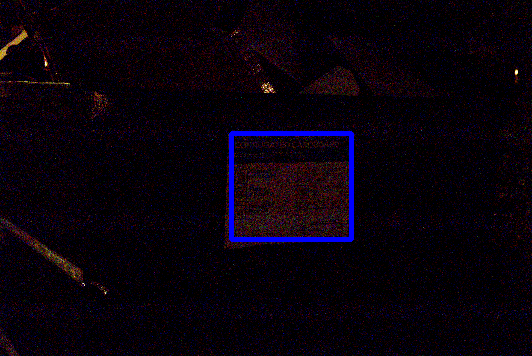}\\
        \includegraphics[width=\linewidth, height=2.3cm]{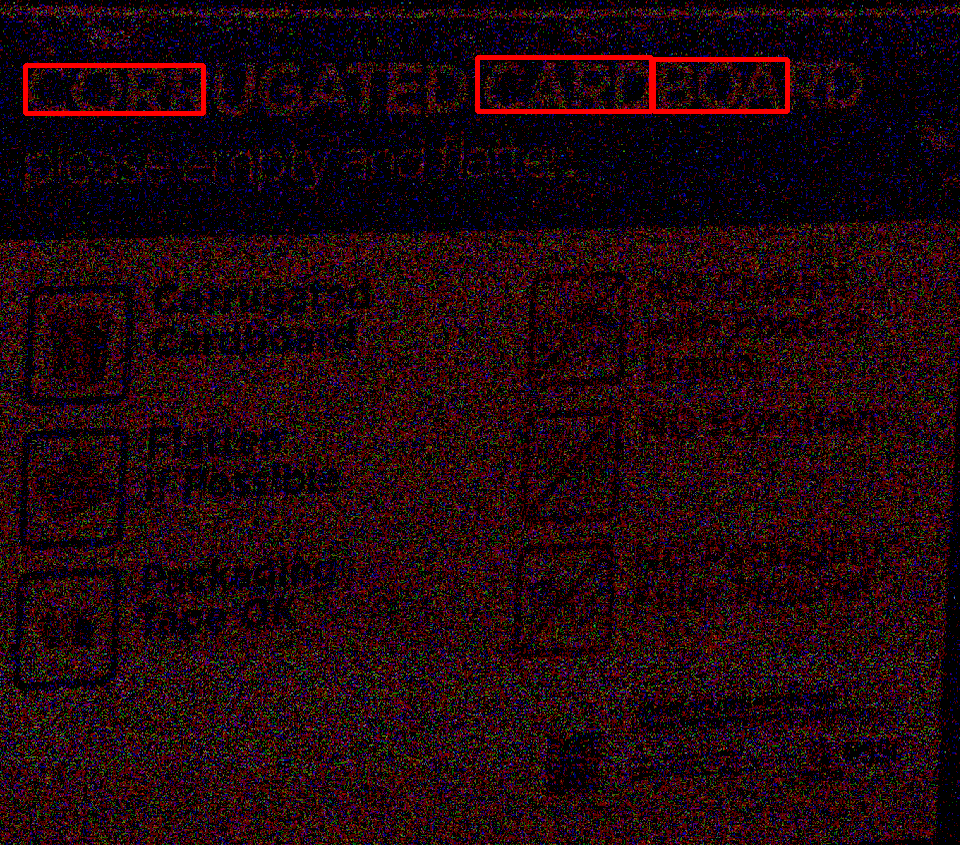}\\
     \end{tabular}
    \caption{Zero-DCE++~\cite{li2021learning}}
    \label{fig:fuji_zero_dce_pp}
    \end{subfigure}\hspace{-0.5mm}
    \begin{subfigure}[b]{0.19\textwidth}
    \begin{tabular}[b]{@{}l@{}}
        \includegraphics[width=\linewidth]{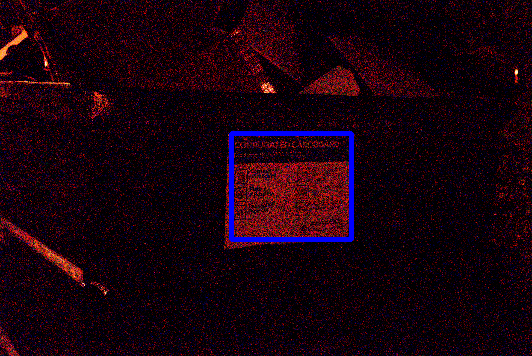}\\
        \includegraphics[width=\linewidth, height=2.3cm]{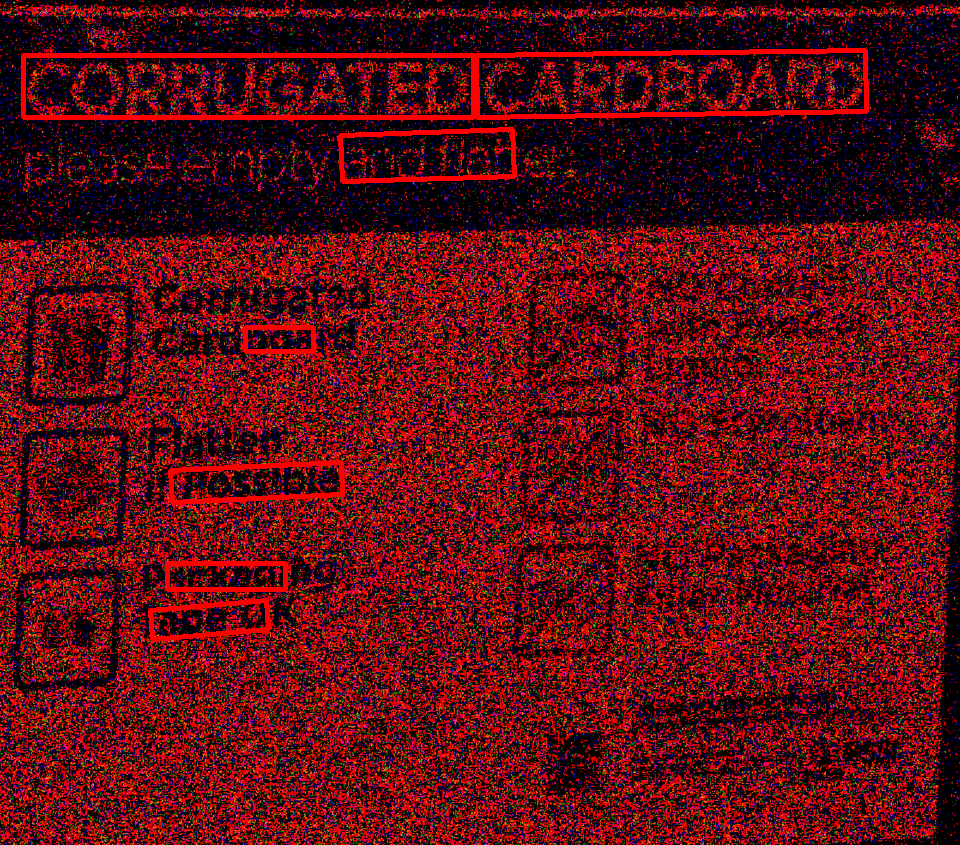}\\
     \end{tabular}
    \caption{SCI~\cite{ma2022toward}}
    \label{fig:fuji_sci}
    \end{subfigure}\hspace{-0.5mm}
    \begin{subfigure}[b]{0.19\textwidth}
    \begin{tabular}[b]{@{}l@{}}
        \includegraphics[width=\linewidth]{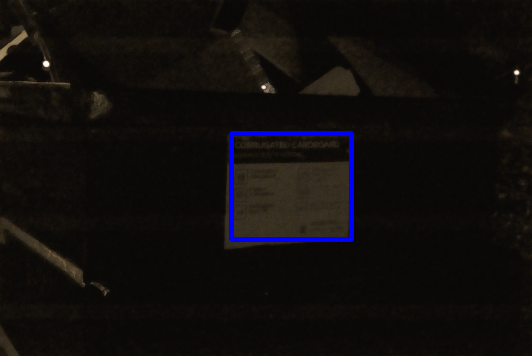}\\
        \includegraphics[width=\linewidth, height=2.3cm]{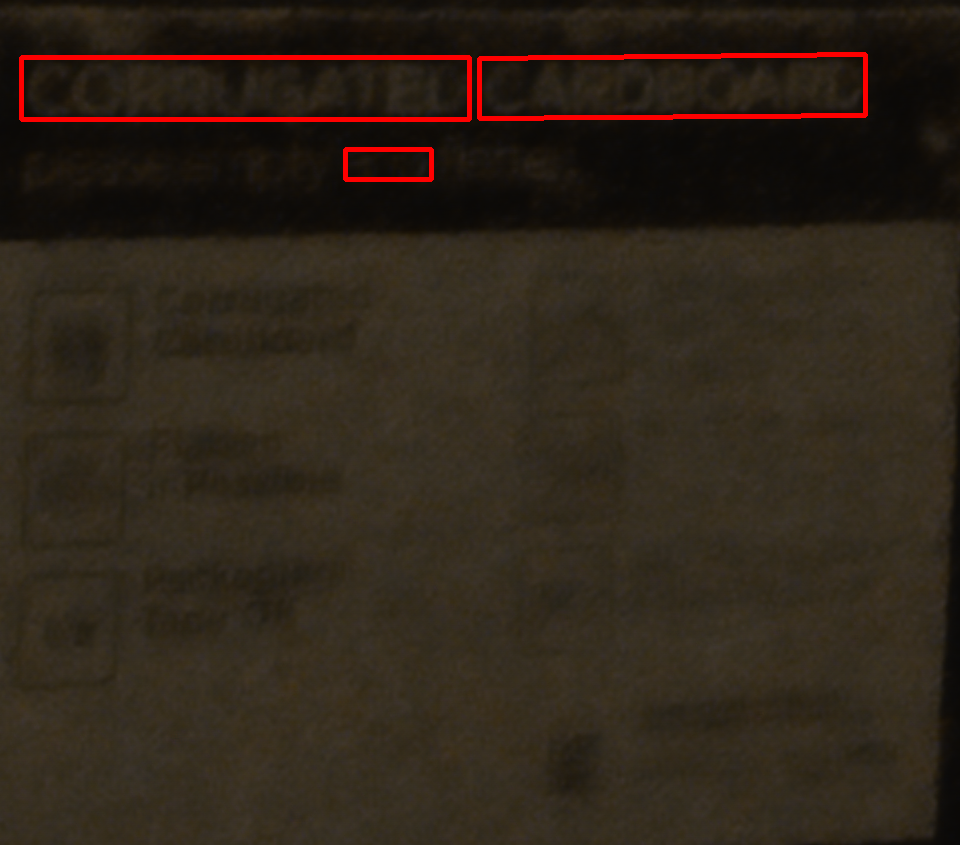}\\
     \end{tabular}
    \caption{CycleGAN~\cite{CycleGAN2017}}
    \label{fig:fuji_craft_cyclegan}
    \end{subfigure}\hspace{-0.5mm}
    \\[1ex]
    \begin{subfigure}[b]{0.24\textwidth}
    \begin{tabular}[b]{@{}l@{}}
        \includegraphics[width=\linewidth]{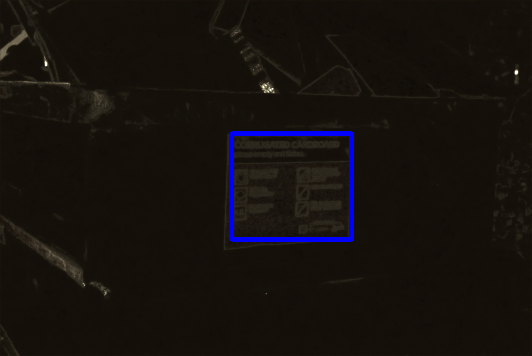}\\
        \includegraphics[width=\linewidth, height=2.3cm]{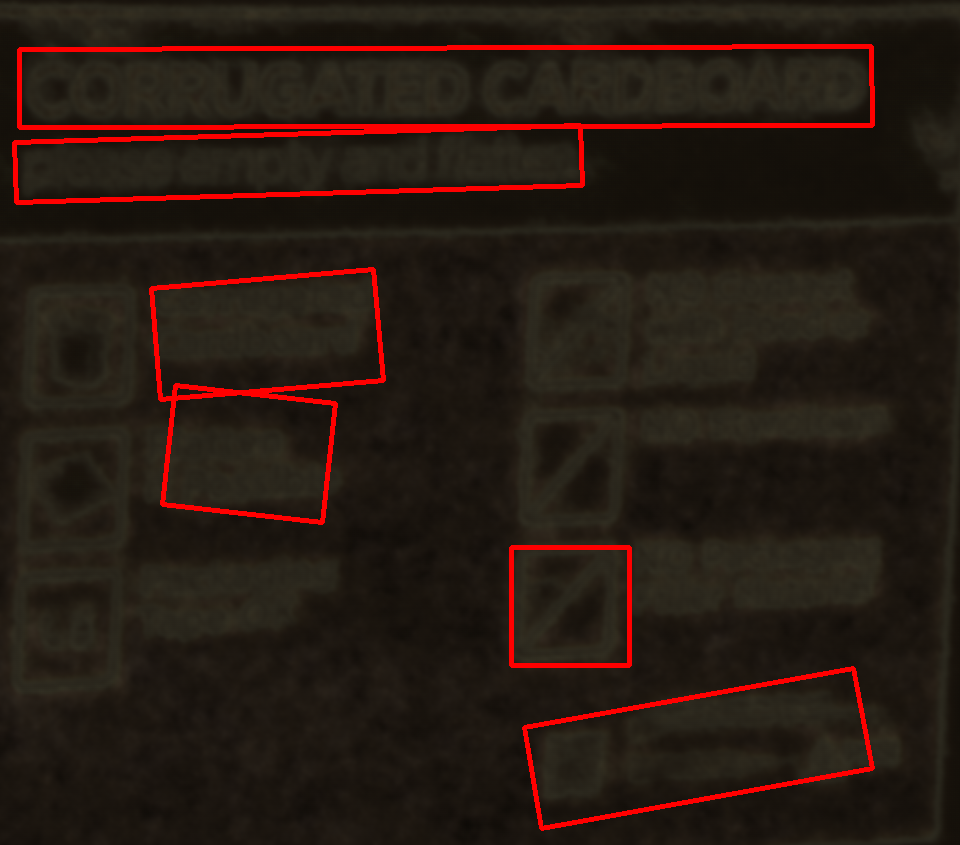}\\
     \end{tabular}
    \caption{EnlightenGAN~\cite{Jiang2019EnlightenGANDL}}
    \label{fig:fuji_craft_enlightengan}
    \end{subfigure}\hspace{-0.5mm}
    \begin{subfigure}[b]{0.24\textwidth}
    \begin{tabular}[b]{@{}l@{}}
        \includegraphics[width=\linewidth]{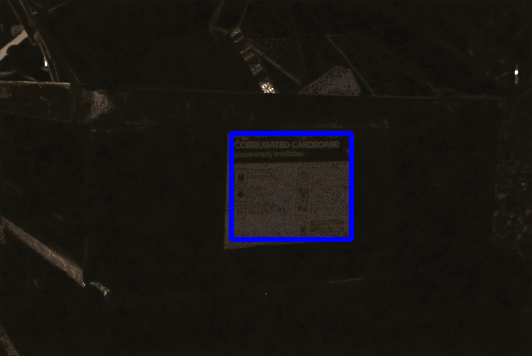}\\
        \includegraphics[width=\linewidth, height=2.3cm]{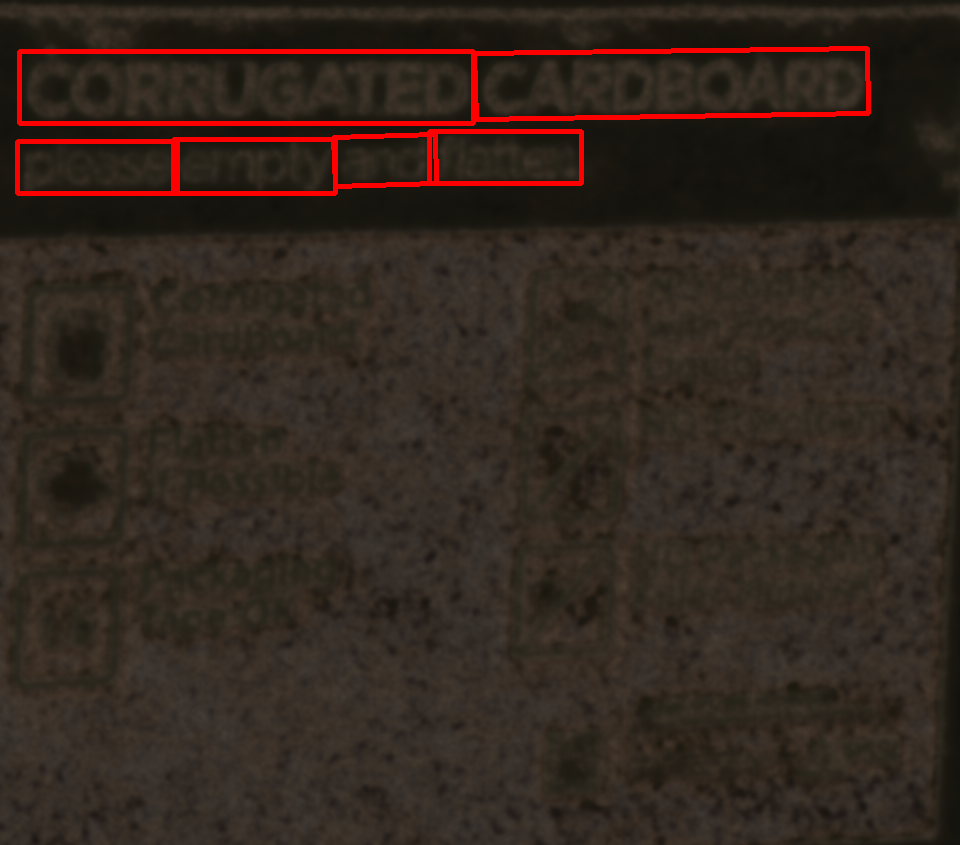}\\
     \end{tabular}
    \caption{Pix2Pix~\cite{isola2017image}}
    \label{fig:fuji_craft_pix2pix}
    \end{subfigure}\hspace{-0.5mm}
    \begin{subfigure}[b]{0.24\textwidth}
    \begin{tabular}[b]{@{}l@{}}
        \includegraphics[width=\linewidth]{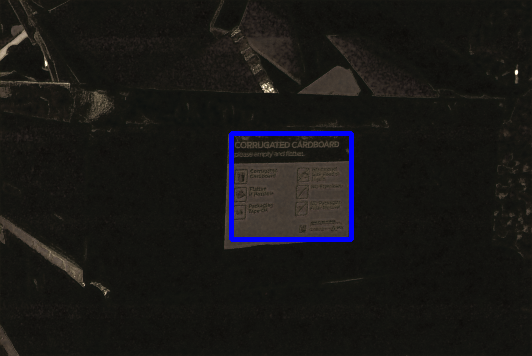}\\
        \includegraphics[width=\linewidth, height=2.3cm]{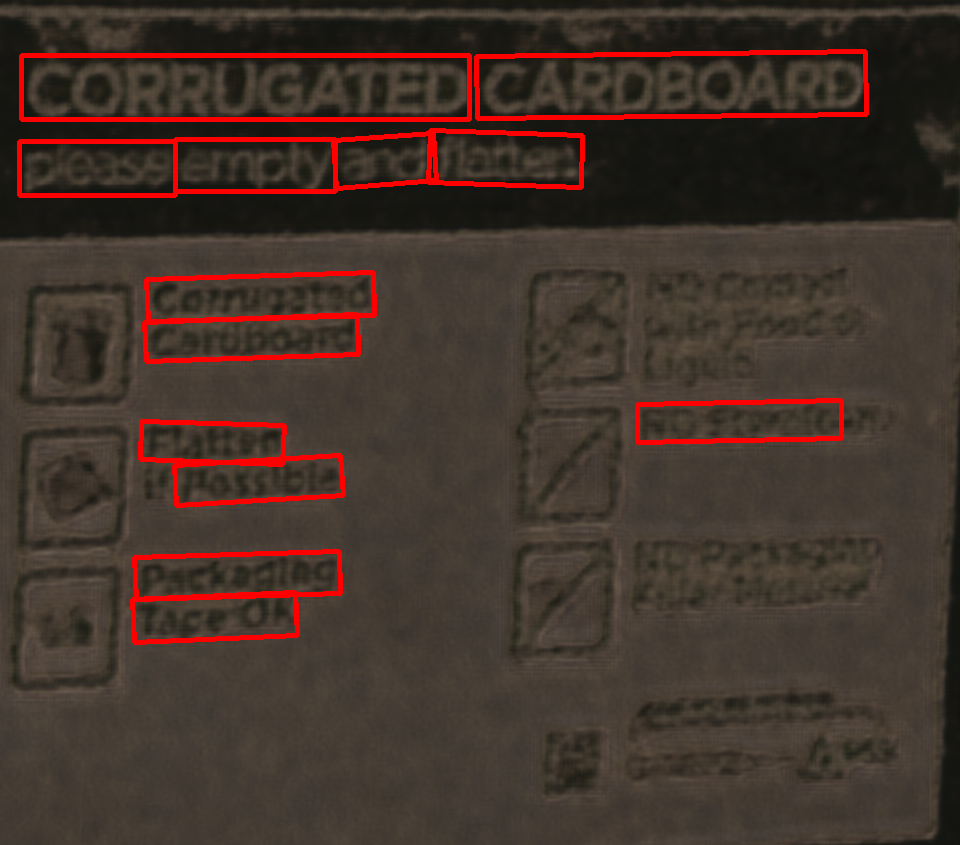}\\
     \end{tabular}
    \caption{ChebyLighter~\cite{pan2022chebylighter}}
    \label{fig:fuji_chebylighter}
    \end{subfigure}\hspace{-0.5mm}
    \begin{subfigure}[b]{0.24\textwidth}
    \begin{tabular}[b]{@{}l@{}}
        \includegraphics[width=\linewidth]{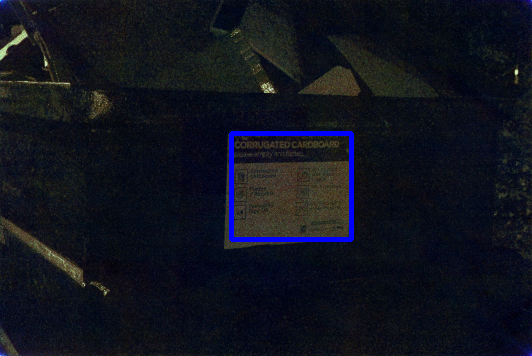}\\
        \includegraphics[width=\linewidth, height=2.3cm]{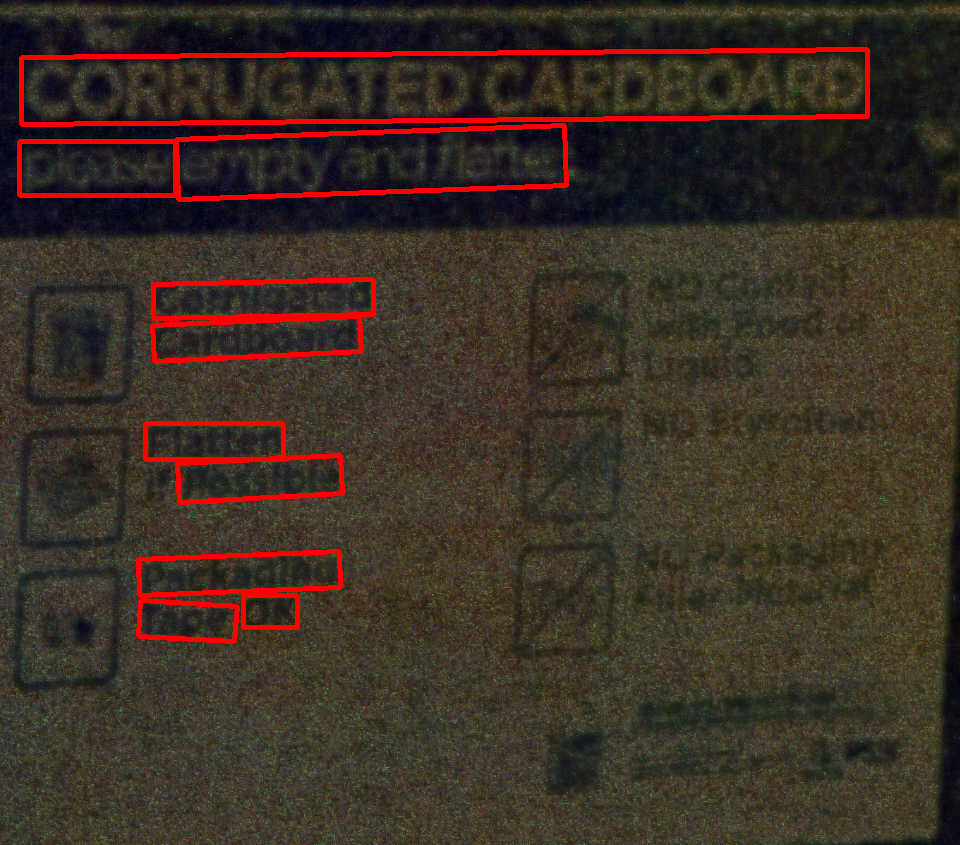}\\
     \end{tabular}
    \caption{FECNet~\cite{huang2022deep}}
    \label{fig:fuji_fecnet}
    \end{subfigure}\hspace{-0.5mm}
    \\[1ex]
    \begin{subfigure}[b]{0.24\textwidth}
    \begin{tabular}[b]{@{}l@{}}
        \includegraphics[width=\linewidth]{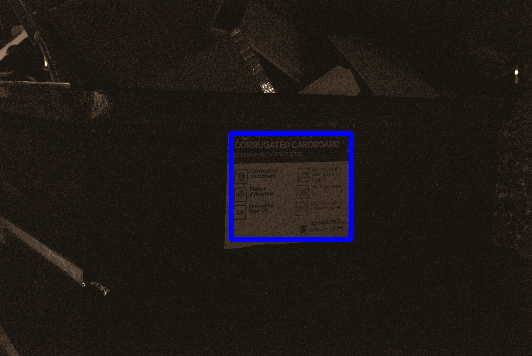}\\
        \includegraphics[width=\linewidth, height=2.3cm]{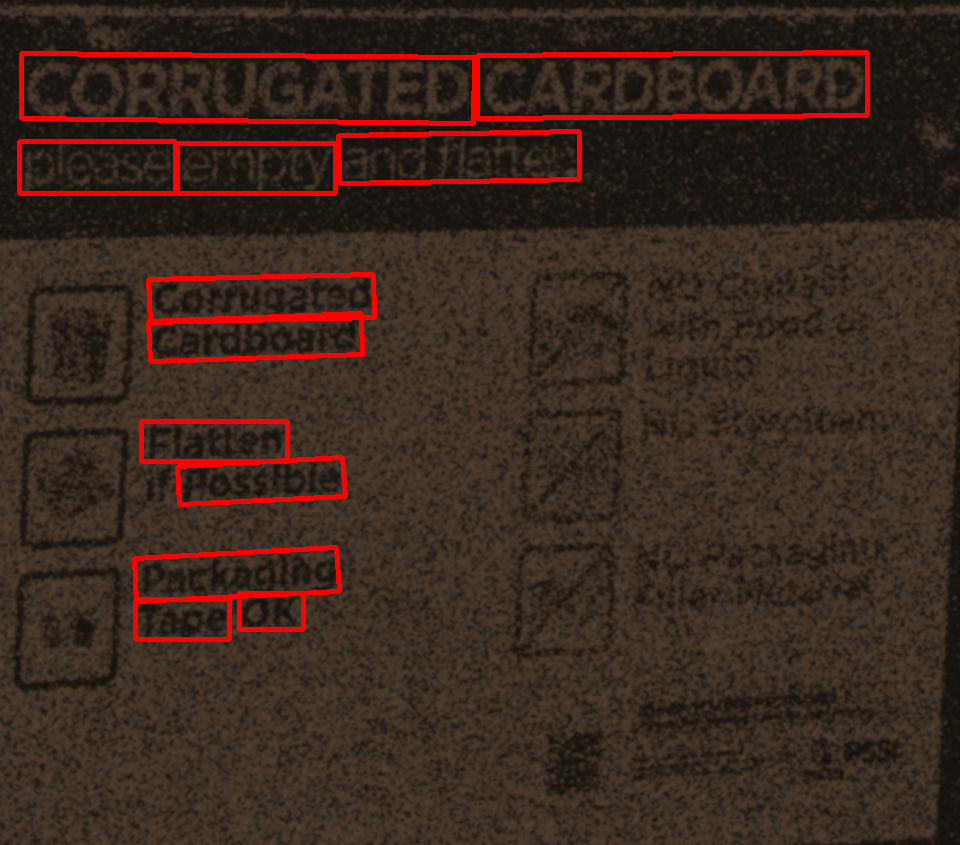}\\
     \end{tabular}
    \caption{IAT~\cite{cui2022you}}
    \label{fig:fuji_iat}
    \end{subfigure}\hspace{-0.5mm}
    \begin{subfigure}[b]{0.24\textwidth}
    \begin{tabular}[b]{@{}l@{}}
        \includegraphics[width=\linewidth]{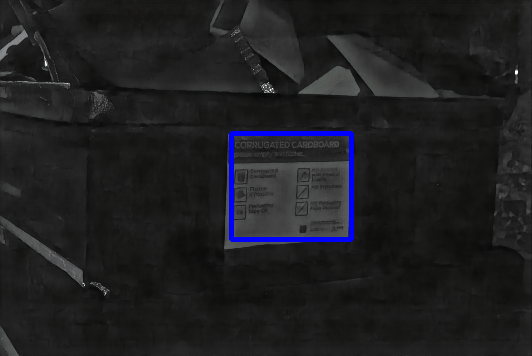}\\
        \includegraphics[width=\linewidth, height=2.3cm]{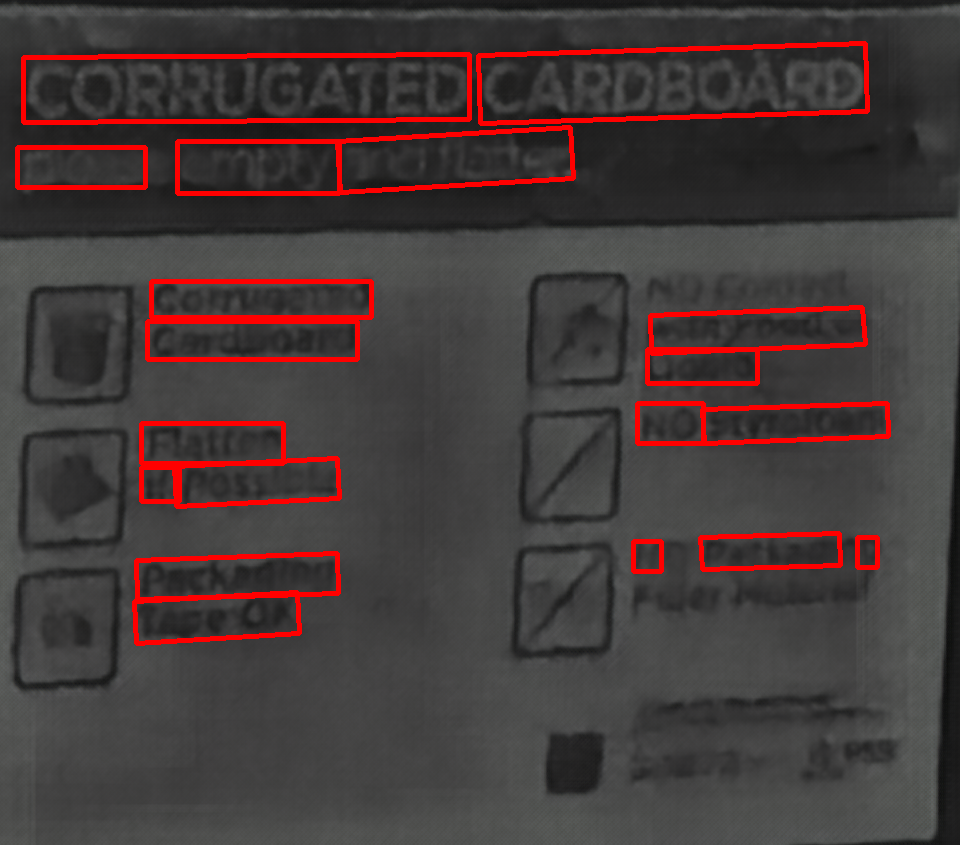}\\
     \end{tabular}
    \caption{ELIE\_STR~\cite{hsu2022extremely}}
    \label{fig:fuji_craft_elie}
    \end{subfigure}\hspace{-0.5mm}
    \begin{subfigure}[b]{0.24\textwidth}
    \begin{tabular}[b]{@{}l@{}}
        \includegraphics[width=\linewidth]{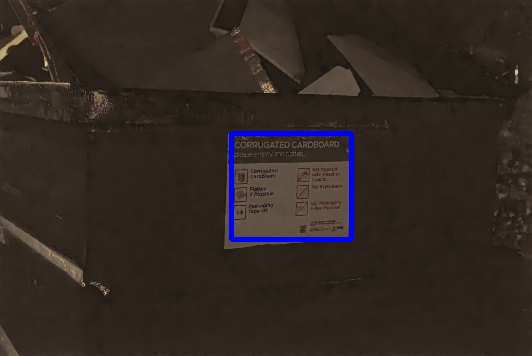}\\
        \includegraphics[width=\linewidth, height=2.3cm]{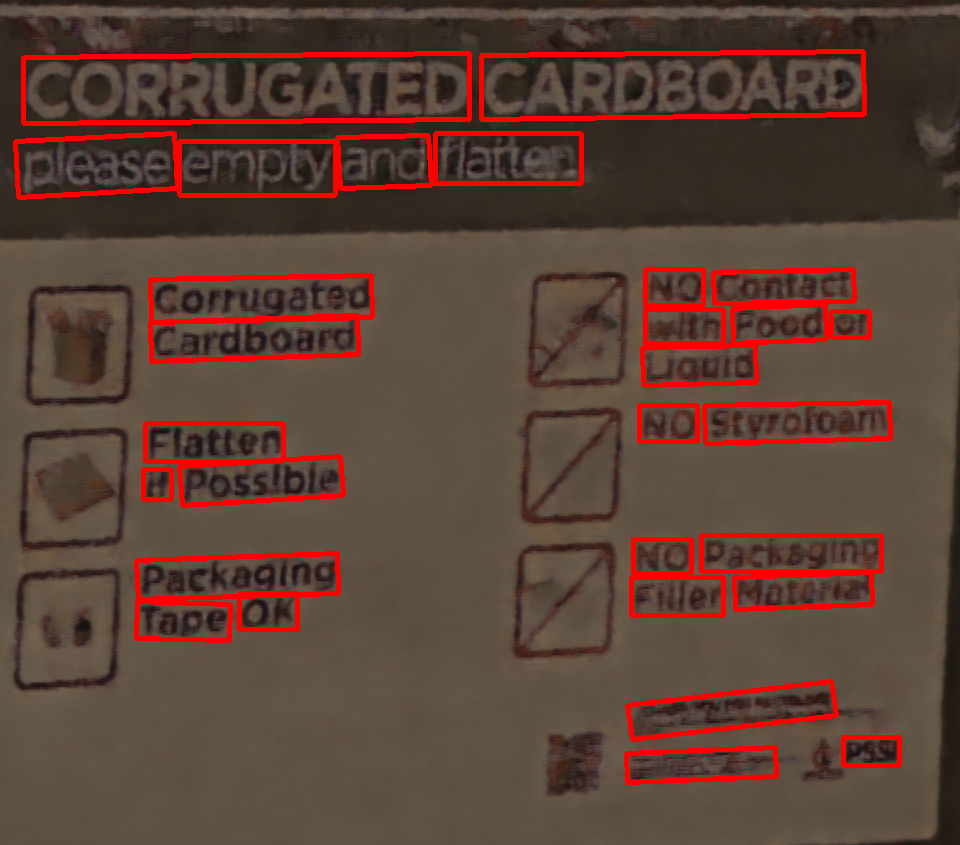}\\
     \end{tabular}
    \caption{Ours}
    \label{fig:fuji_craft_ours}
    \end{subfigure}\hspace{-0.5mm}
    \begin{subfigure}[b]{0.24\textwidth}
    \begin{tabular}[b]{@{}l@{}}
        \includegraphics[width=\linewidth]{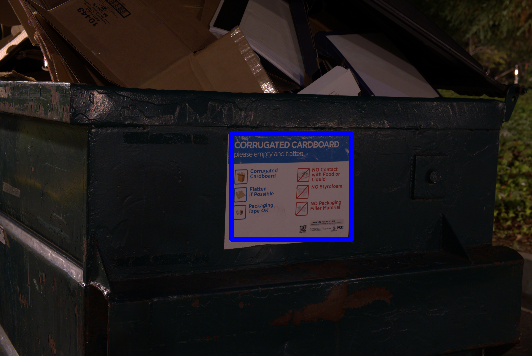}\\
        \includegraphics[width=\linewidth, height=2.3cm]{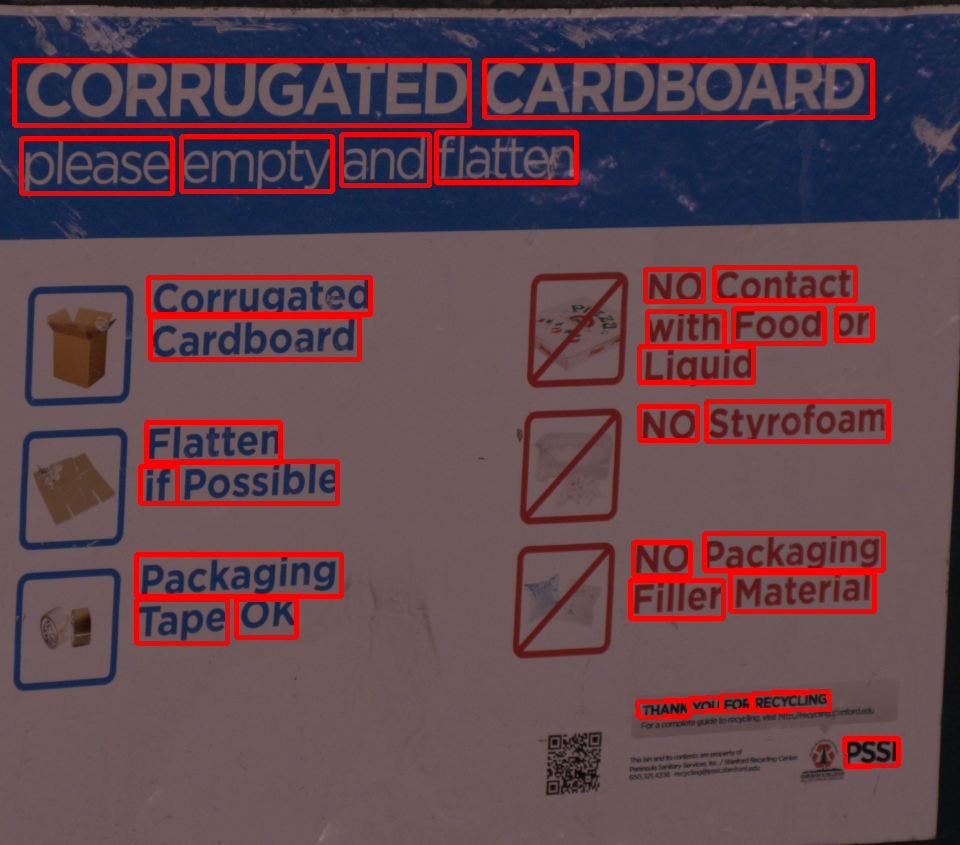}\\
     \end{tabular}
    \caption{Ground Truth}
    \label{fig:fuji_craft_gt}
    \end{subfigure}
    \caption{Comparison with state-of-the-art methods on the SID-Fuji-Text dataset is shown in the following manner: for each column, the first row displays enhanced images marked with blue boxes as regions of interest. The second row displays zoomed-in regions of enhanced images overlaid with red text detection boxes from CRAFT~\cite{Baek2019CharacterRA}. Column \ref{fig:fuji_craft_input} displays the low-light image. Columns \ref{fig:fuji_craft_zero_dce} to \ref{fig:fuji_craft_ours} show image enhancement results from all related methods. The last cell displays ground truth images.}
    \label{fig:fuji_fig}
\end{figure*}

\begin{figure*}[!htbp]
    \centering
    \centering
    \begin{subfigure}[b]{0.19\textwidth}
    \begin{tabular}[b]{@{}l@{}}
        \includegraphics[width=\linewidth]{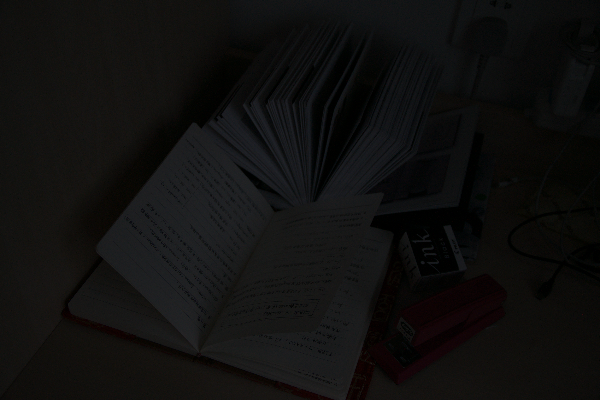}\\
        \includegraphics[width=\linewidth, height=2.3cm]{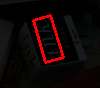}\\
     \end{tabular}
    \caption{Low-Light}
    \label{fig:lol_craft_input}
    \end{subfigure}\hspace{-0.5mm}
    \begin{subfigure}[b]{0.19\textwidth}
    \begin{tabular}[b]{@{}l@{}}
        \includegraphics[width=\linewidth]{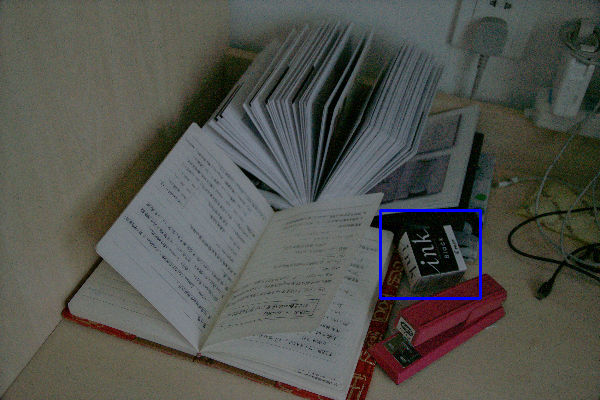}\\
        \includegraphics[width=\linewidth, height=2.3cm]{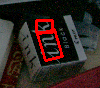}\\
     \end{tabular}
    \caption{Zero-DCE~\cite{Zero-DCE}}
    \label{fig:lol_craft_zero_dce}
    \end{subfigure}\hspace{-0.5mm}
    \begin{subfigure}[b]{0.205\textwidth}
    \begin{tabular}[b]{@{}l@{}}
        \includegraphics[width=\linewidth, height=1.74cm]{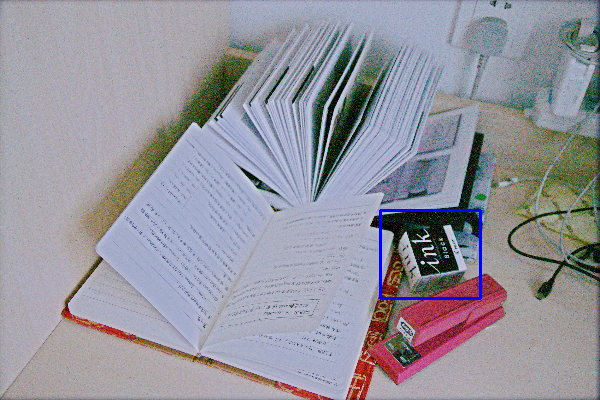}\\
        \includegraphics[width=\linewidth, height=2.3cm]{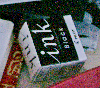}\\
     \end{tabular}
    \caption{Zero-DCE++~\cite{li2021learning}}
    \label{fig:lol_zero_dce_pp}
    \end{subfigure}\hspace{-0.5mm}
    \begin{subfigure}[b]{0.19\textwidth}
    \begin{tabular}[b]{@{}l@{}}
        \includegraphics[width=\linewidth]{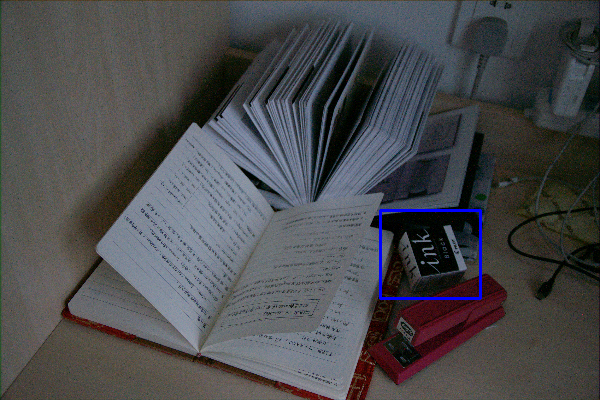}\\
        \includegraphics[width=\linewidth, height=2.3cm]{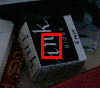}\\
     \end{tabular}
    \caption{SCI~\cite{ma2022toward}}
    \label{fig:lol_sci}
    \end{subfigure}\hspace{-0.5mm}
    \begin{subfigure}[b]{0.19\textwidth}
    \begin{tabular}[b]{@{}l@{}}
        \includegraphics[width=\linewidth]{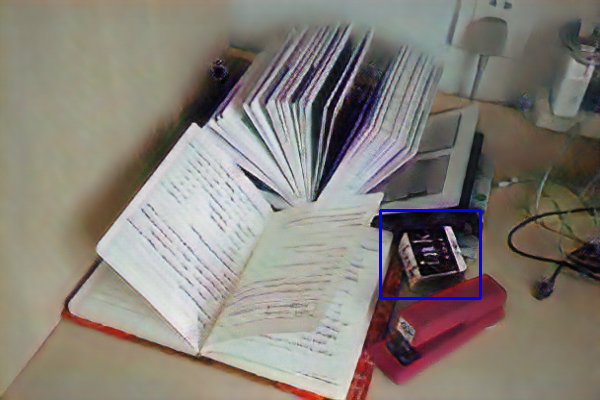}\\
        \includegraphics[width=\linewidth, height=2.3cm]{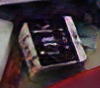}\\
     \end{tabular}
    \caption{CycleGAN~\cite{CycleGAN2017}}
    \label{fig:lol_craft_cyclegan}
    \end{subfigure}\hspace{-0.5mm}
    \\[1ex]
    \begin{subfigure}[b]{0.24\textwidth}
    \begin{tabular}[b]{@{}l@{}}
        \includegraphics[width=\linewidth]{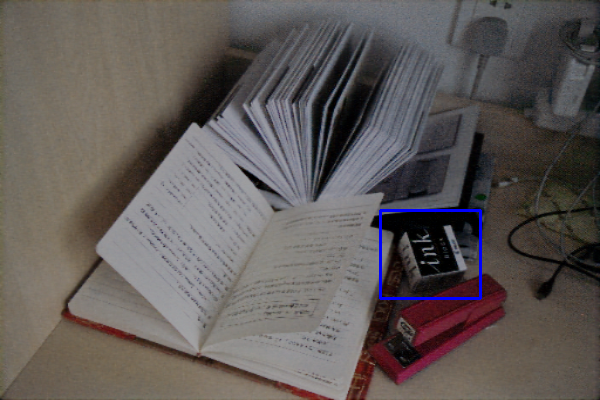}\\
        \includegraphics[width=\linewidth, height=2.3cm]{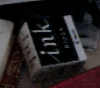}\\
     \end{tabular}
    \caption{EnlightenGAN~\cite{Jiang2019EnlightenGANDL}}
    \label{fig:lol_craft_enlightengan}
    \end{subfigure}\hspace{-0.5mm}
    \begin{subfigure}[b]{0.24\textwidth}
    \begin{tabular}[b]{@{}l@{}}
        \includegraphics[width=\linewidth]{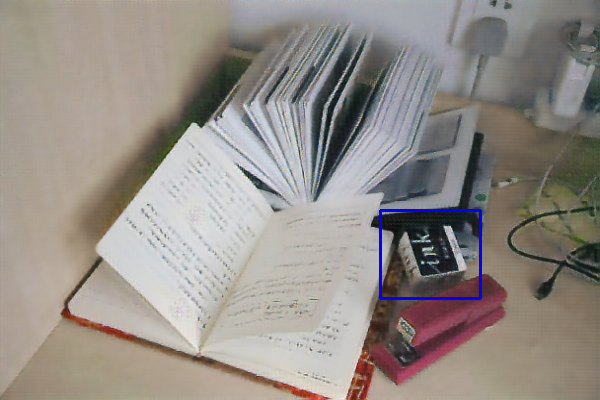}\\
        \includegraphics[width=\linewidth, height=2.3cm]{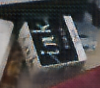}\\
     \end{tabular}
    \caption{Pix2Pix~\cite{isola2017image}}
    \label{fig:lol_craft_pix2pix}
    \end{subfigure}\hspace{-0.5mm}
    \begin{subfigure}[b]{0.24\textwidth}
    \begin{tabular}[b]{@{}l@{}}
        \includegraphics[width=\linewidth]{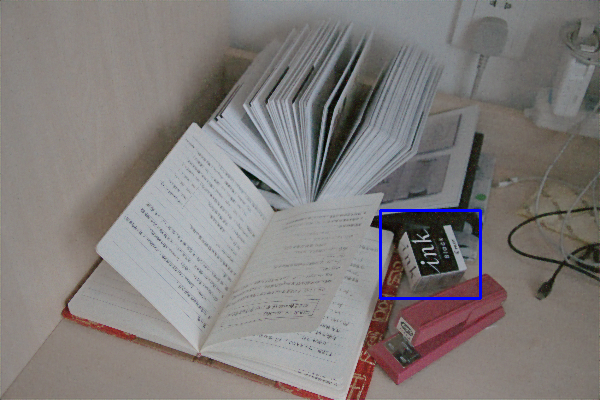}\\
        \includegraphics[width=\linewidth, height=2.3cm]{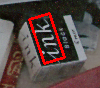}\\
     \end{tabular}
    \caption{ChebyLighter~\cite{pan2022chebylighter}}
    \label{fig:lol_chebylighter}
    \end{subfigure}\hspace{-0.5mm}
    \begin{subfigure}[b]{0.24\textwidth}
    \begin{tabular}[b]{@{}l@{}}
        \includegraphics[width=\linewidth]{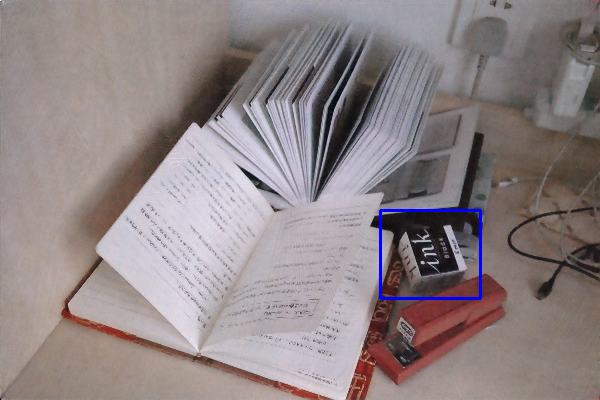}\\
        \includegraphics[width=\linewidth, height=2.3cm]{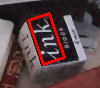}\\
     \end{tabular}
    \caption{FECNet~\cite{huang2022deep}}
    \label{fig:lol_fecnet}
    \end{subfigure}\hspace{-0.5mm}
    \\[1ex]
    \begin{subfigure}[b]{0.24\textwidth}
    \begin{tabular}[b]{@{}l@{}}
        \includegraphics[width=\linewidth]{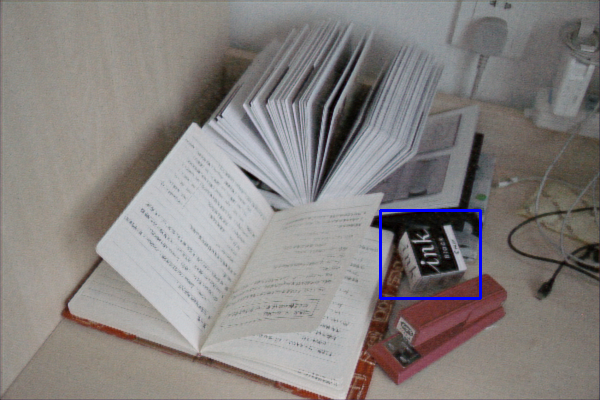}\\
        \includegraphics[width=\linewidth, height=2.3cm]{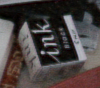}\\
     \end{tabular}
    \caption{IAT~\cite{cui2022you}}
    \label{fig:lol_iat}
    \end{subfigure}\hspace{-0.5mm}
    \begin{subfigure}[b]{0.24\textwidth}
    \begin{tabular}[b]{@{}l@{}}
        \includegraphics[width=\linewidth]{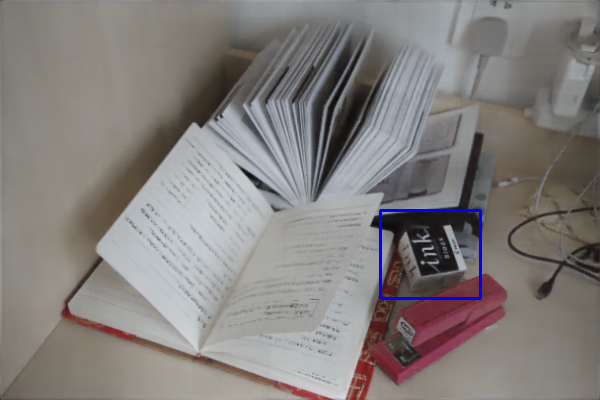}\\
        \includegraphics[width=\linewidth, height=2.3cm]{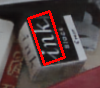}\\
     \end{tabular}
    \caption{ELIE\_STR~\cite{hsu2022extremely}}
    \label{fig:lol_craft_elie}
    \end{subfigure}\hspace{-0.5mm}
    \begin{subfigure}[b]{0.24\textwidth}
    \begin{tabular}[b]{@{}l@{}}
        \includegraphics[width=\linewidth]{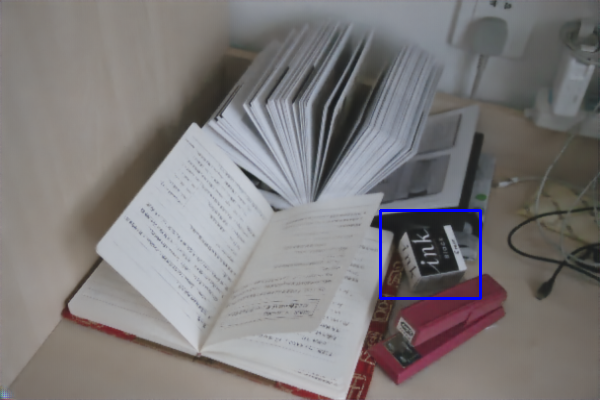}\\
        \includegraphics[width=\linewidth, height=2.3cm]{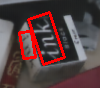}\\
     \end{tabular}
    \caption{Ours}
    \label{fig:lol_craft_ours}
    \end{subfigure}\hspace{-0.5mm}
    \begin{subfigure}[b]{0.24\textwidth}
    \begin{tabular}[b]{@{}l@{}}
        \includegraphics[width=\linewidth]{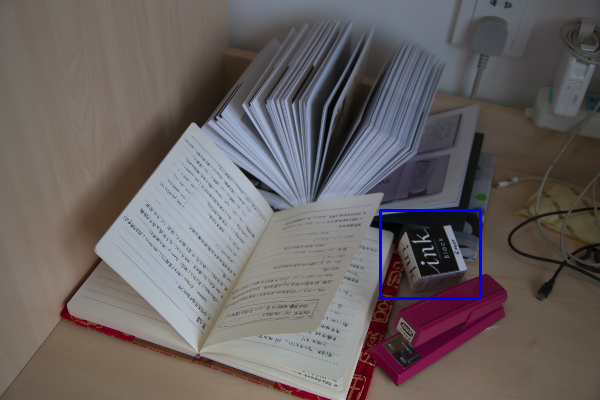}\\
        \includegraphics[width=\linewidth, height=2.3cm]{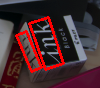}\\
     \end{tabular}
    \caption{Ground Truth}
    \label{fig:lol_craft_gt}
    \end{subfigure}\hspace{-0.5mm}
    \caption{Comparison with state-of-the-art methods on the LOL-Text dataset is shown in the following manner: for each column, the first row displays enhanced images marked with blue boxes as regions of interest. The second row displays zoomed-in regions of enhanced images overlaid with red text detection boxes from CRAFT~\cite{Baek2019CharacterRA}. Column \ref{fig:lol_craft_input} displays the low-light image. Columns \ref{fig:lol_craft_zero_dce} to \ref{fig:lol_craft_ours} show image enhancement results from all related methods. The last cell displays ground truth images.}
    \label{fig:lol_fig}
\end{figure*}

We present qualitative results of existing methods on the SID-Fuji-Text dataset in Figure \ref{fig:fuji_fig} and results on the LOL-Text dataset in Figure \ref{fig:lol_fig}. Through these two figures, we observe that the enhanced images of our proposed method achieve the closest text detection results compared to the ground truth images.

\section{Qualitative Results of the Proposed Supervised-DCE Model}

\begin{figure*}[!ht]
    \centering
    \begin{subfigure}{0.32\linewidth}
        \centering
        \begin{tabular}[b]{@{}l@{}}
         \includegraphics[keepaspectratio=true, scale=0.2]{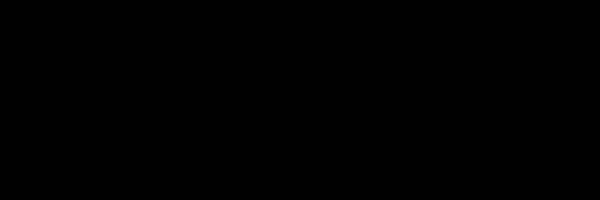} \\
         \includegraphics[keepaspectratio=true, scale=0.2]{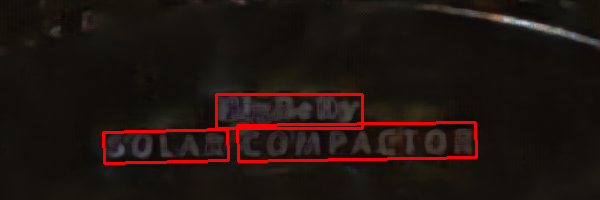} \\
         \centerline{\setstretch{3} \footnotesize \textcolor{red}{rightly}, \textcolor{green}{solar}, \textcolor{green}{compactor}} \\
         \includegraphics[keepaspectratio=true, scale=0.08, angle=180]{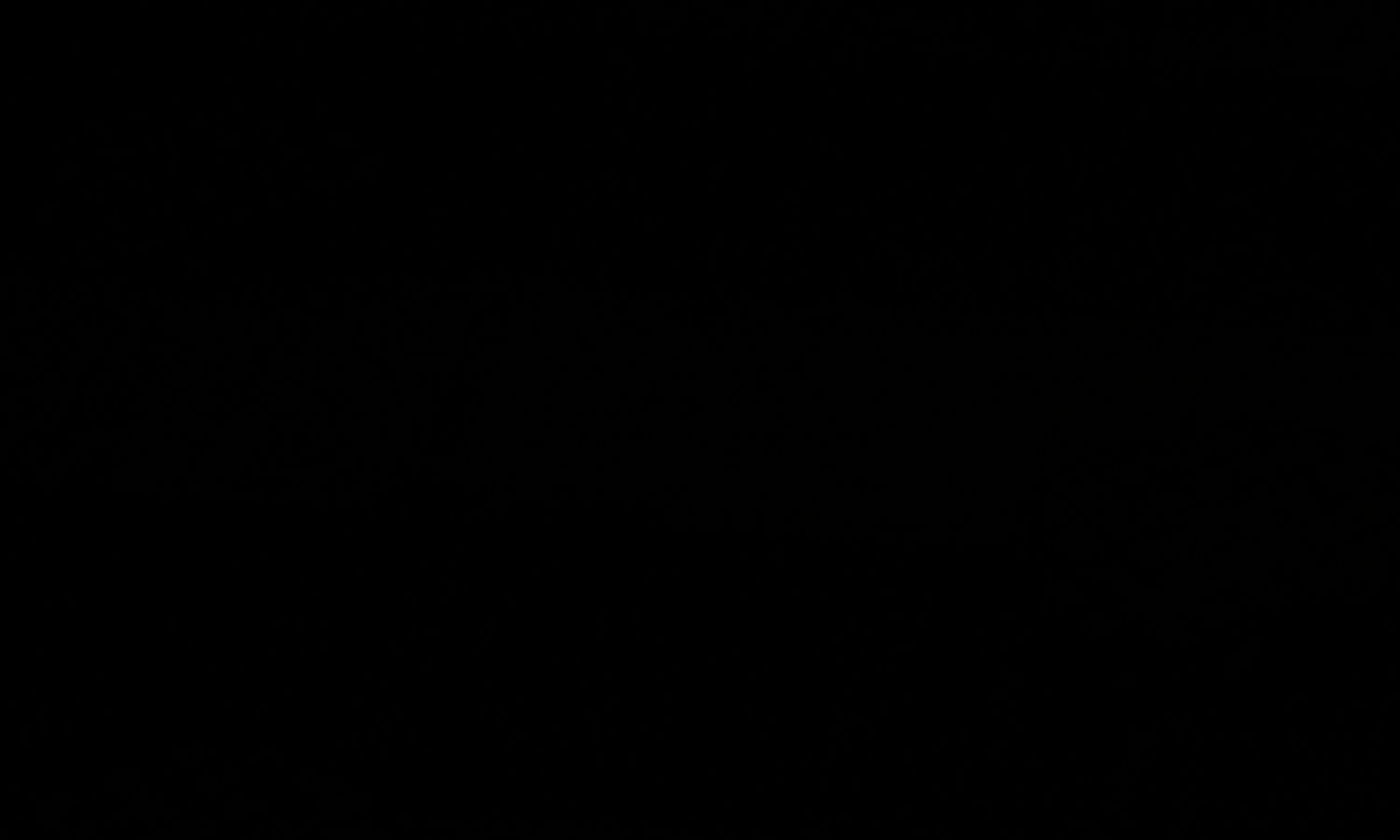} \\
         \includegraphics[keepaspectratio=true, scale=0.08, angle=180]{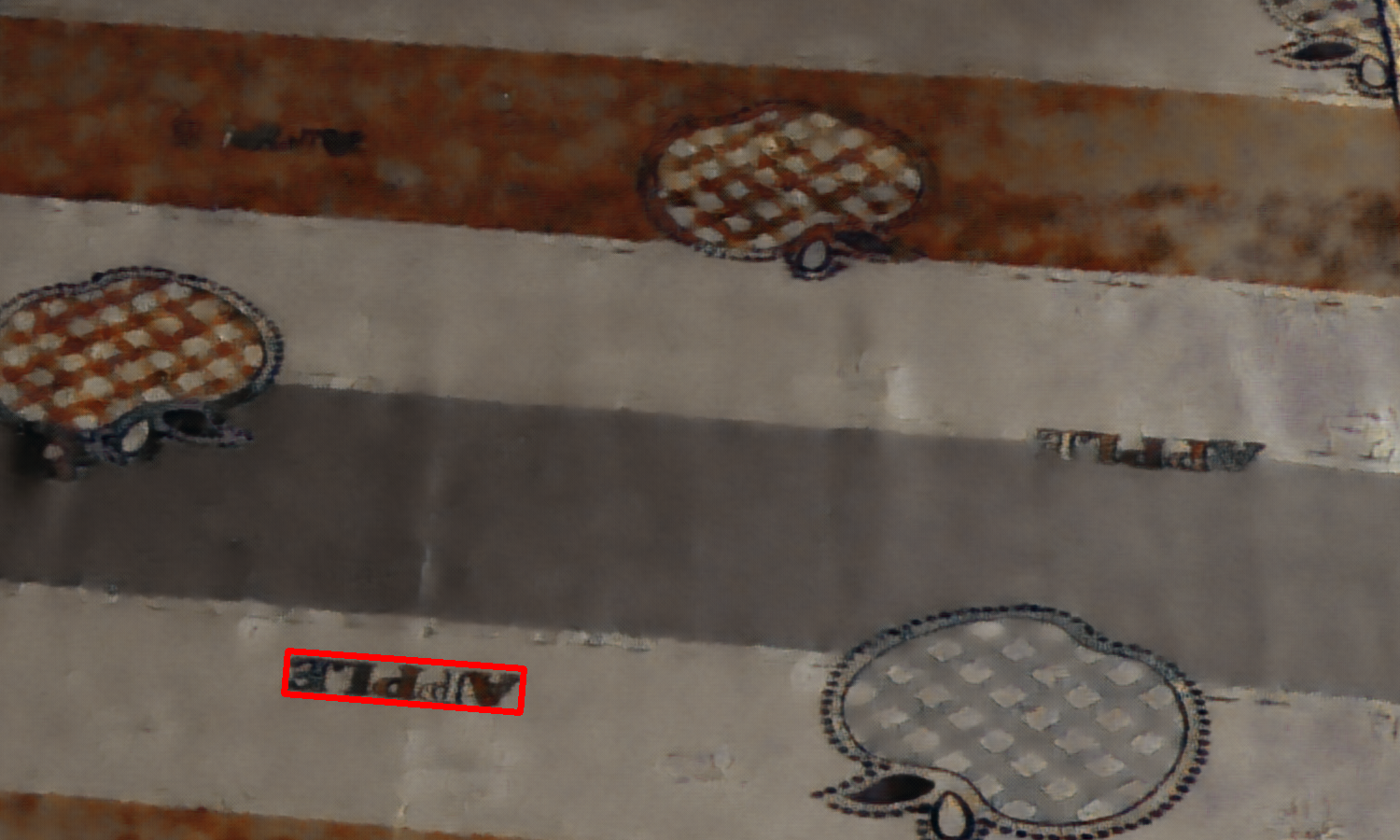} \\
         \centerline{\setstretch{3} \footnotesize \textcolor{green}{apple}, \textcolor{red}{nil}, \textcolor{red}{nil}} \\
         \end{tabular}
         \caption{Genuine SID-Sony}
     \end{subfigure}\hspace{-0.5mm}
     \begin{subfigure}{0.32\linewidth}
        \centering
        \begin{tabular}[b]{@{}l@{}}
         \includegraphics[keepaspectratio=true, scale=0.2]{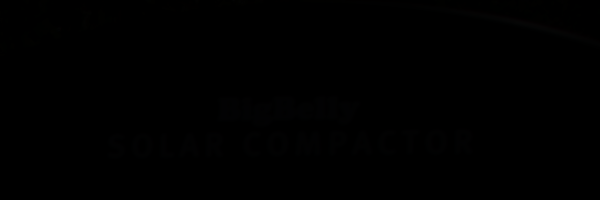} \\
         \includegraphics[keepaspectratio=true, scale=0.2]{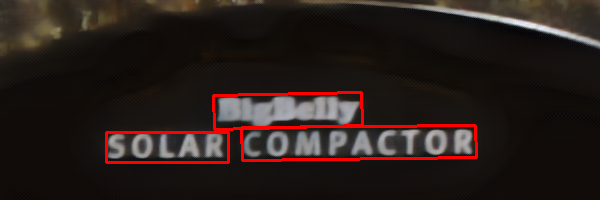} \\
         \centerline{\setstretch{3} \footnotesize \textcolor{green}{bigbelly}, \textcolor{green}{solar}, \textcolor{green}{compactor}} \\
         \includegraphics[keepaspectratio=true, scale=0.08, angle=180]{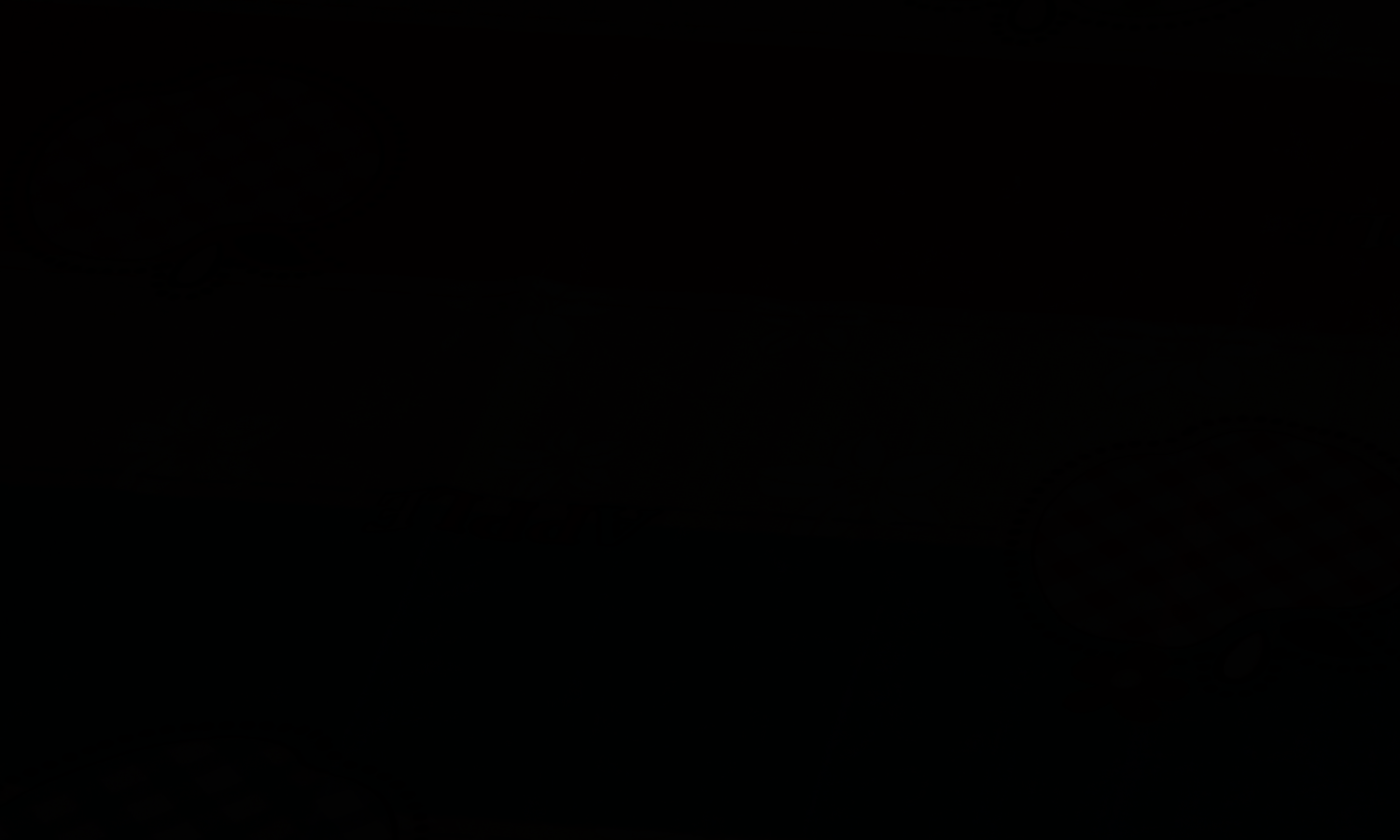} \\
         \includegraphics[keepaspectratio=true, scale=0.08, angle=180]{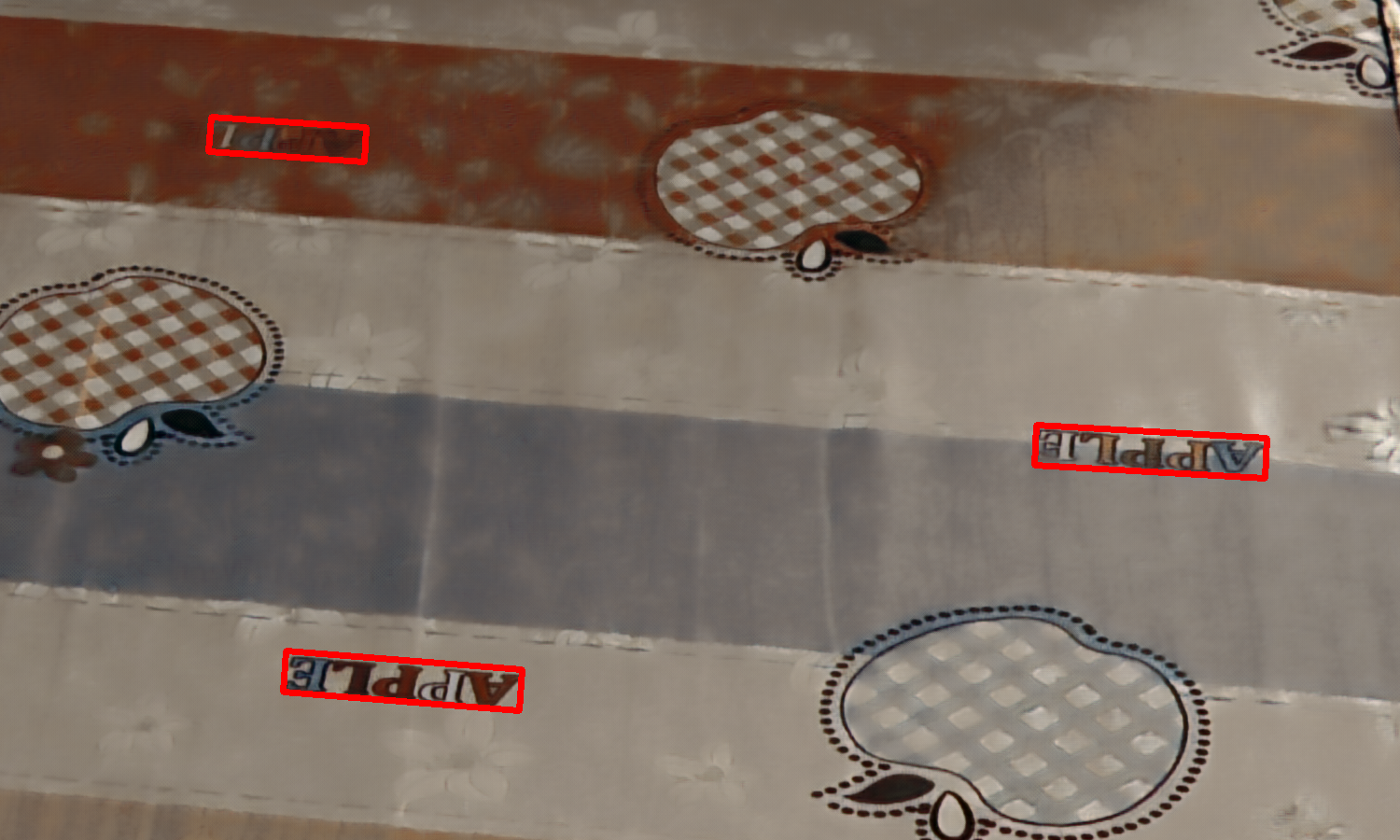} \\
         \centerline{\setstretch{3} \footnotesize \textcolor{green}{apple}, \textcolor{green}{apple}, \textcolor{red}{ampi}} \\
        \end{tabular}
        \caption{Syn-SID-Sony-v1}
     \end{subfigure}\hspace{-0.5mm}
     \begin{subfigure}{0.32\linewidth}
        \centering
        \begin{tabular}[b]{@{}l@{}}
         \includegraphics[keepaspectratio=true, scale=0.2]{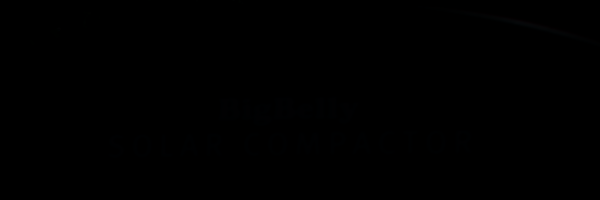} \\
         \includegraphics[keepaspectratio=true, scale=0.2]{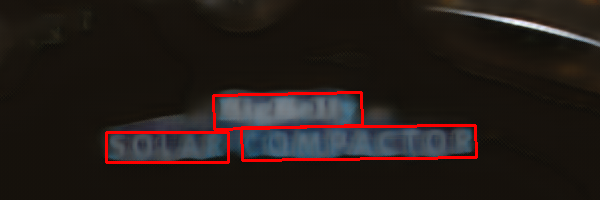} \\
         \centerline{\setstretch{3} \footnotesize \textcolor{red}{you}, \textcolor{green}{solar}, \textcolor{green}{compactor}} \\
         \includegraphics[keepaspectratio=true, scale=0.08, angle=180]{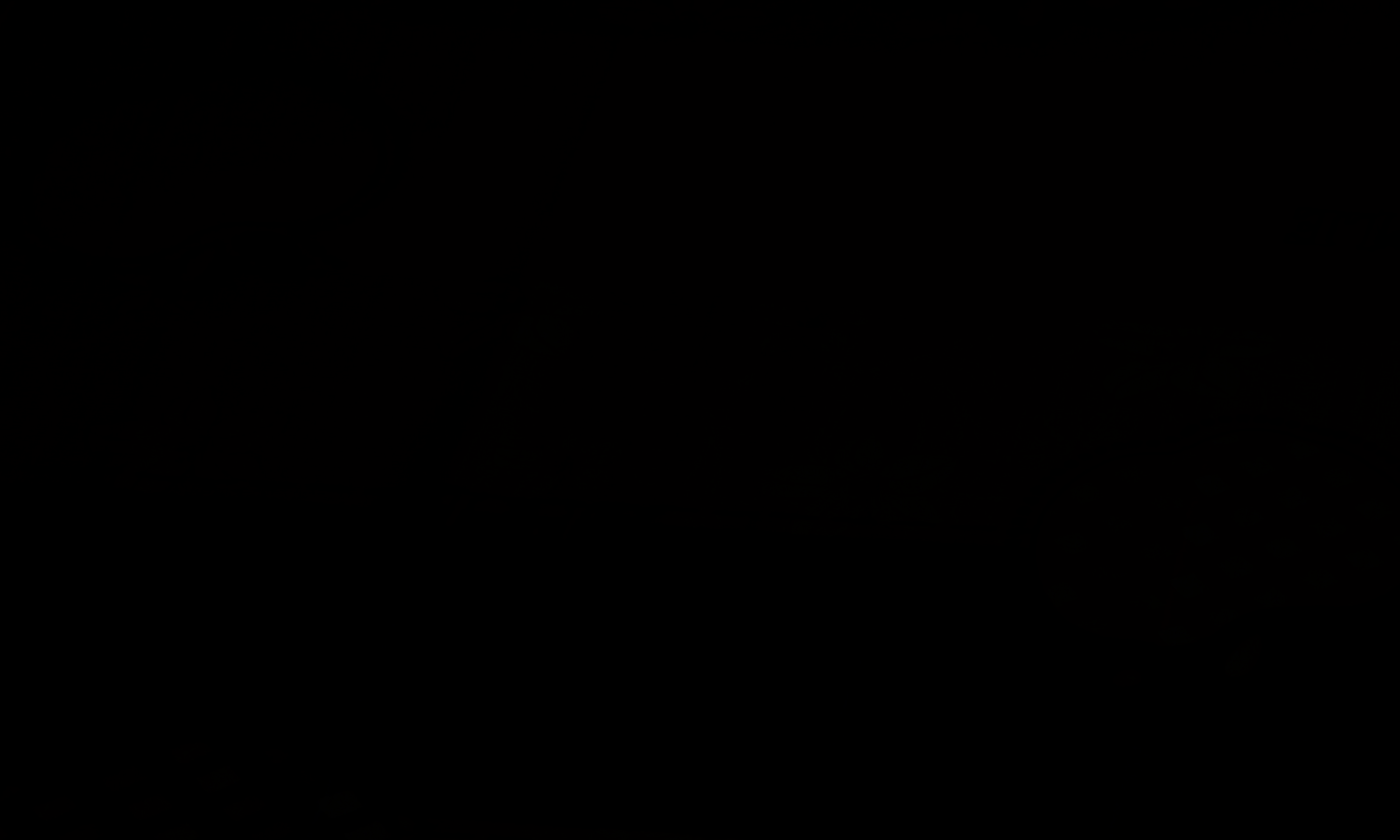} \\
         \includegraphics[keepaspectratio=true, scale=0.08, angle=180]{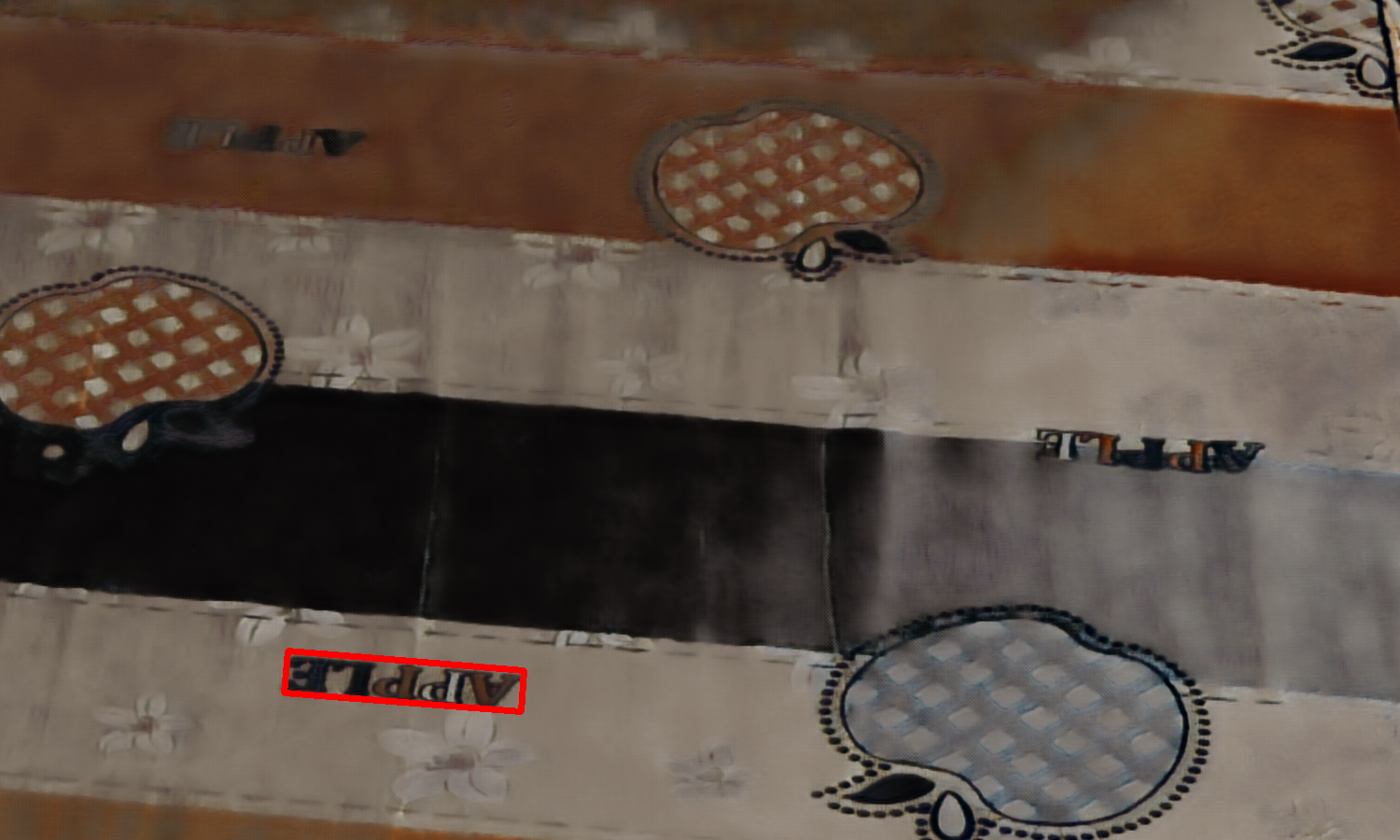} \\
         \centerline{\setstretch{3} \footnotesize \textcolor{green}{apple}, \textcolor{red}{nil}, \textcolor{red}{nil}} \\
         \end{tabular}
         \caption{Syn-SID-Sony-v2}
     \end{subfigure}
    \caption{Text detection results (CRAFT) and text recognition results (ASTER) of the enhanced images (genuine SID-\{Sony, Fuji\}, Syn-SID-\{Sony, Fuji\}-v1, and Syn-SID-\{Sony, Fuji\}-v2). (a) a genuine SID-Sony low-light image and its enhanced counterpart with detection and recognition results; (b) a Syn-SID-Sony-v1 low-light image and its enhanced counterpart with detection and recognition results; (c) a Syn-SID-Sony-v2 low-light image and its enhanced counterpart with detection and recognition results. Text recognition results of each detection box are listed from top-to-bottom then left to right in the captions.}
    \label{fig:genuine_synthetic_text_detection}
\end{figure*}

As shown in Figure~\ref{fig:genuine_synthetic_text_detection}, in comparison to synthetic v1-images, there are minor differences between genuine extremely low-light images and synthetic v2-images after being enhanced regarding text detection and recognition. This confirms that our proposed low-light synthesis model, Supervised-DCE, can synthesize more realistic extremely low-light images with similar intensity and noise distributions as the genuine extremely low-light images.

\section{Results on Synthesized IC15 Datasets}



In this assessment, we first cropped the texts based on the detection results using a constraint of Intersection over Union (IoU) greater than 0.5 with the ground truth bounding boxes. Subsequently, we performed text recognition (using TRBA and ASTER) on the cropped detected text regions. Given that the quality of text detection plays a crucial role in text spotting, our proposed method's success in outperforming the runner-up method, ELIE\_STR, by a significant margin, highlights the effectiveness of our approach in preserving fine details within text regions while enhancing low-light images. The complete quantitative results are presented in Table \ref{table:full_syn_results}.

\begin{figure*}[!htb]
    \centering
    \begin{subfigure}[b]{0.24\textwidth}    
    \begin{tabular}[b]{@{}l@{}}
        \includegraphics[width=\linewidth, height=1.8cm]{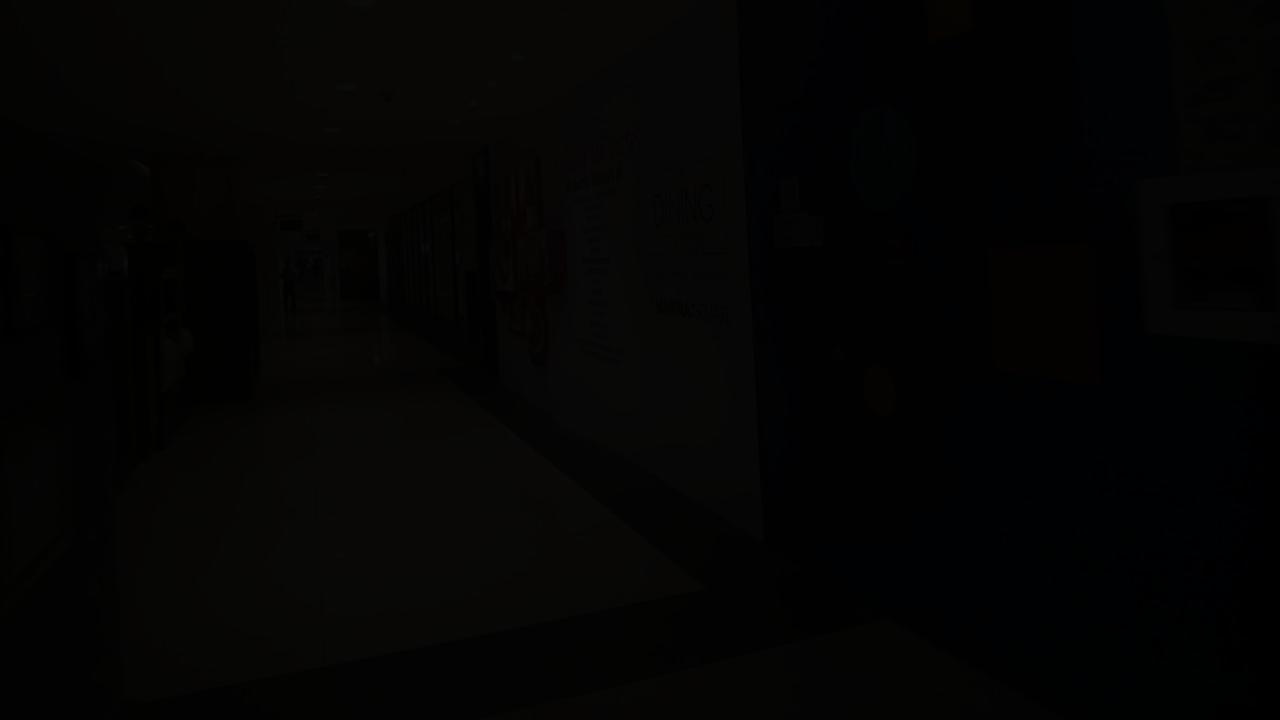}\\
        \includegraphics[width=\linewidth, height=0.8cm]{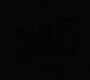}\\
        \centerline{\setstretch{3} \footnotesize \textcolor{red}{nil}} \\ 
        \includegraphics[width=\linewidth, height=1.8cm]{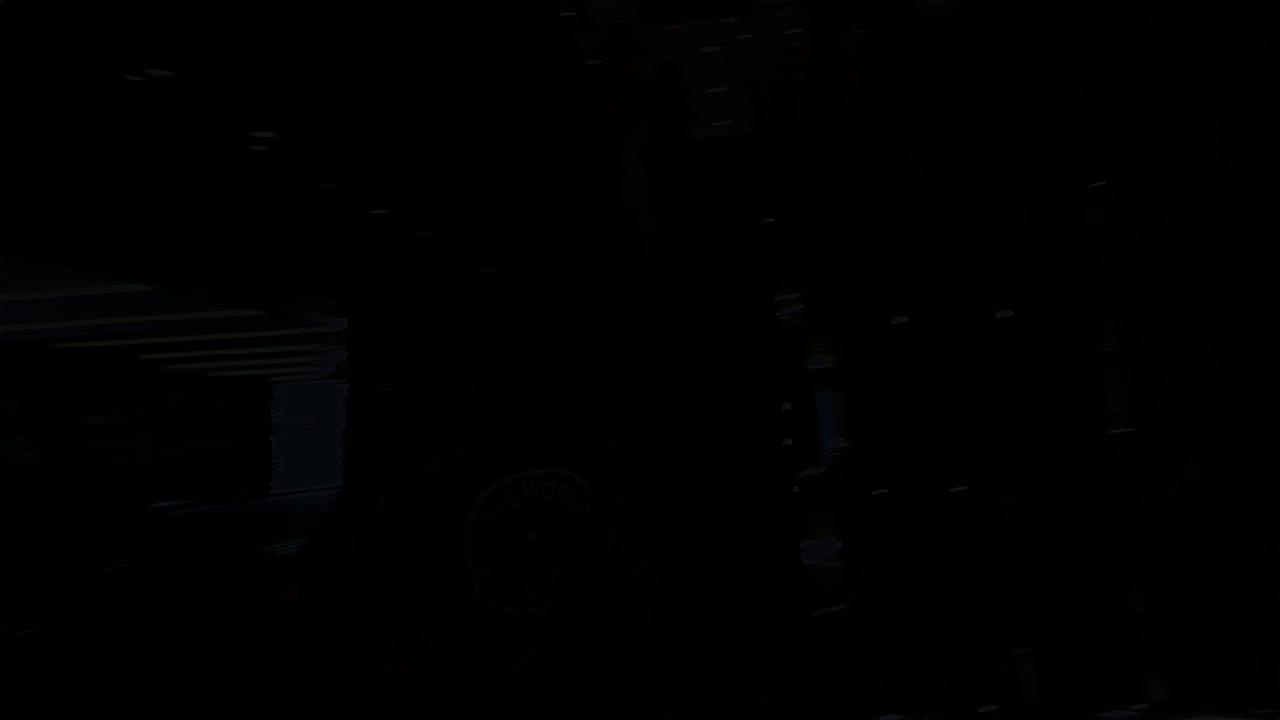}\\
        \includegraphics[width=\linewidth, height=0.8cm]{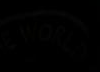}\\
        \centerline{\setstretch{3} \footnotesize \textcolor{red}{nil}} \\ 
    \end{tabular}
    \caption{Low-Light}
    \label{fig:syn_sony_v1_input}                  
    \end{subfigure}\hspace{-0.5mm}
    \begin{subfigure}[b]{0.24\textwidth}
    \begin{tabular}[b]{@{}l@{}}
        \includegraphics[width=\linewidth, height=1.8cm]{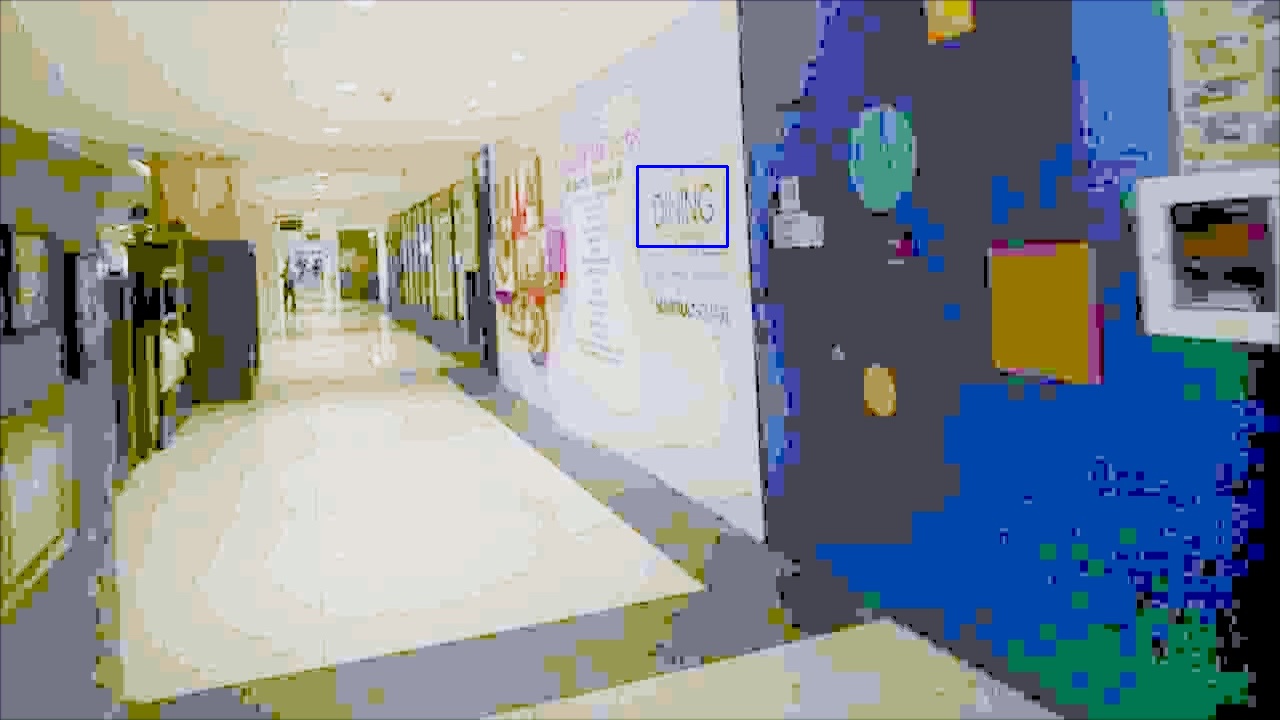}\\
        \includegraphics[width=\linewidth, height=0.8cm]{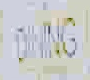}\\
        \centerline{\setstretch{3} \footnotesize \textcolor{red}{nil}} \\
        \includegraphics[width=\linewidth, height=1.8cm]{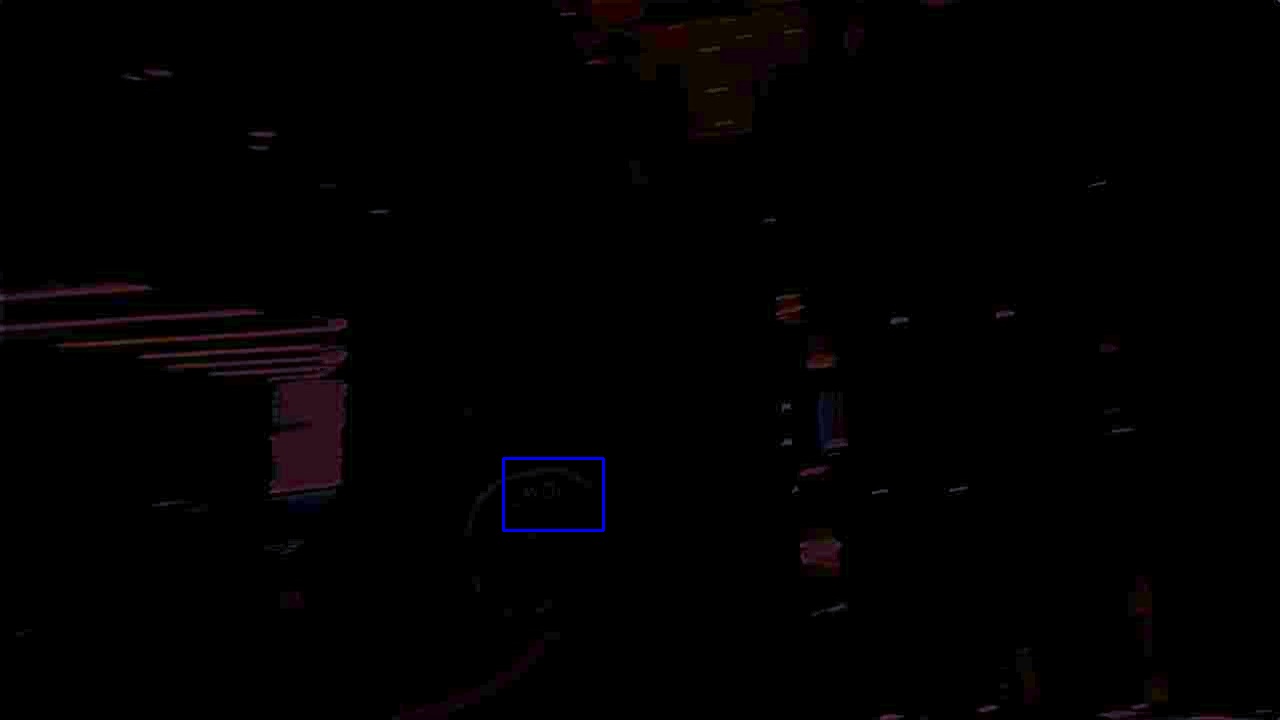}\\
        \includegraphics[width=\linewidth, height=0.8cm]{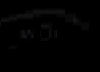}\\
        \centerline{\setstretch{3} \footnotesize \textcolor{red}{nil}} \\ 
    \end{tabular}
    \caption{Zero-DCE++~\cite{li2021learning}}    
    \label{fig:syn_sony_v1_zero_dce_pp}
    \end{subfigure}\hspace{-0.5mm}
    \begin{subfigure}[b]{0.24\textwidth}   
    \begin{tabular}[b]{@{}l@{}}
        \includegraphics[width=\linewidth, height=1.8cm]{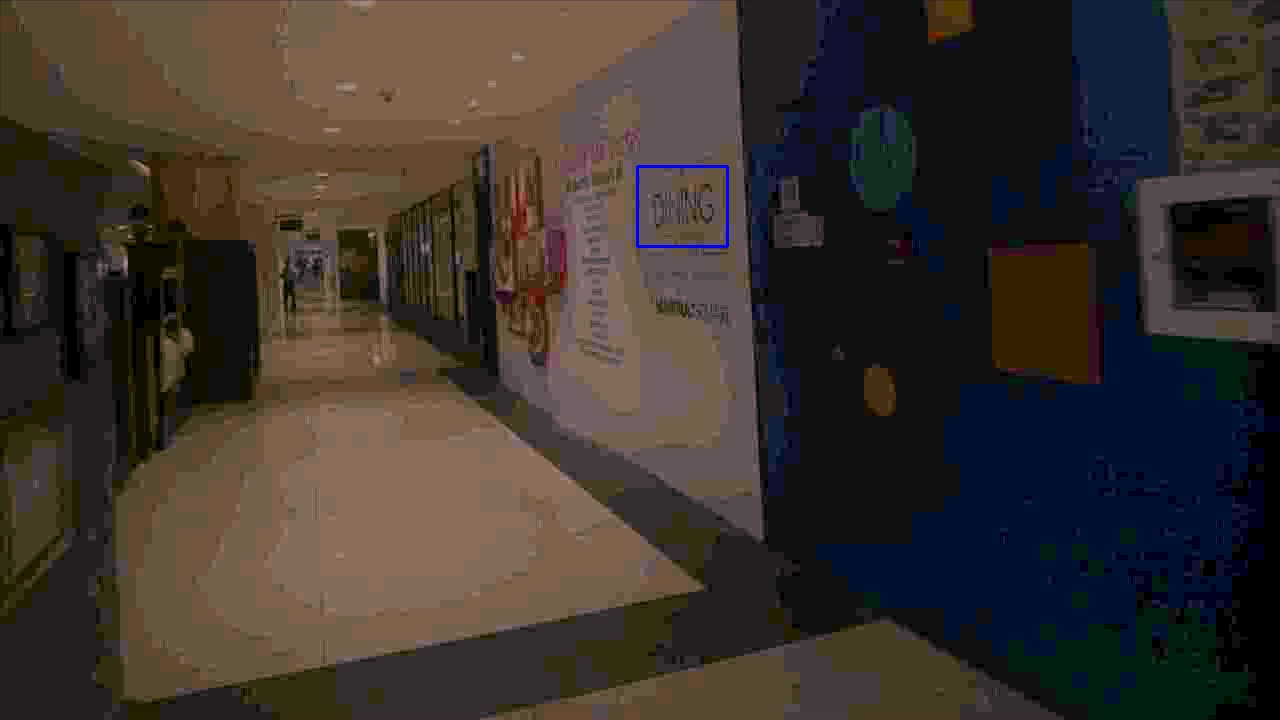}\\
        \includegraphics[width=\linewidth, height=0.8cm]{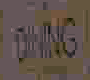}\\
        \centerline{\setstretch{3} \footnotesize \textcolor{red}{nil}} \\     
        \includegraphics[width=\linewidth, height=1.8cm]{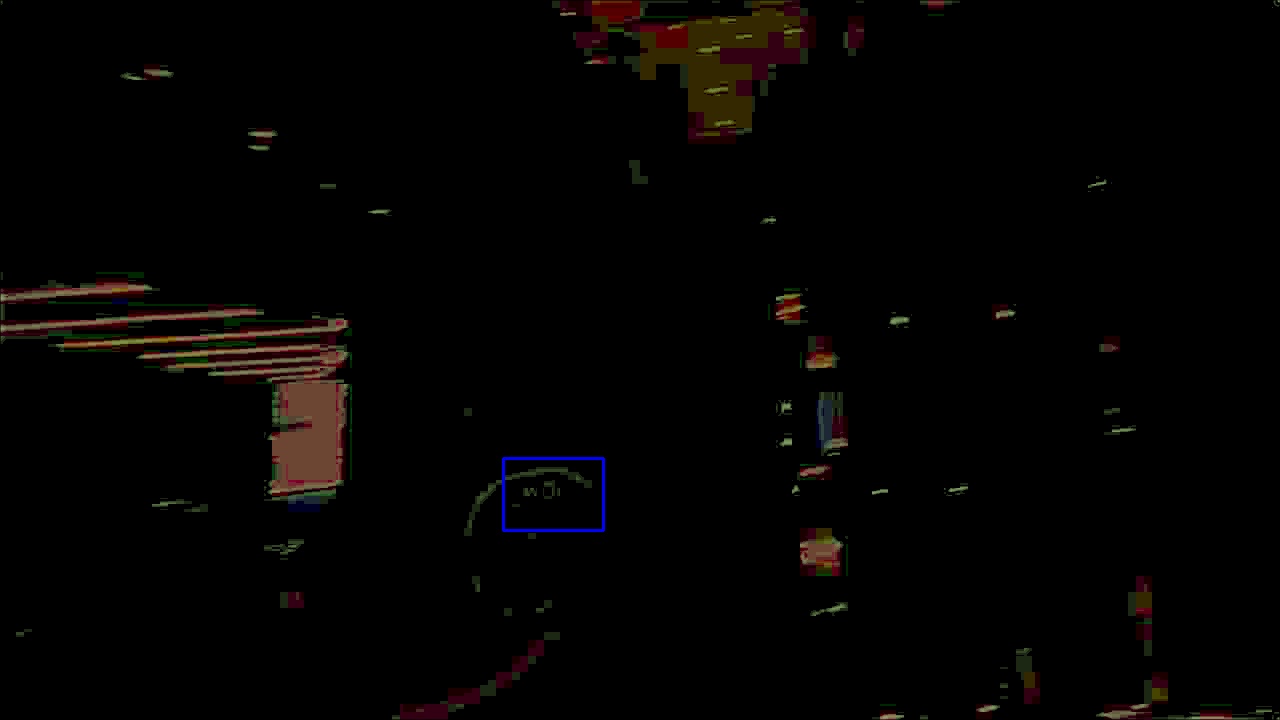}\\
        \includegraphics[width=\linewidth, height=0.8cm]{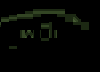}\\
        \centerline{\setstretch{3} \footnotesize \textcolor{red}{nil}} \\     
    \end{tabular}
    \caption{SCI~\cite{ma2022toward}}
    \label{fig:syn_sony_v1_sci}
    \end{subfigure}\hspace{-0.5mm}
    \begin{subfigure}[b]{0.24\textwidth}
    \begin{tabular}[b]{@{}l@{}}
        \includegraphics[width=\linewidth, height=1.8cm]{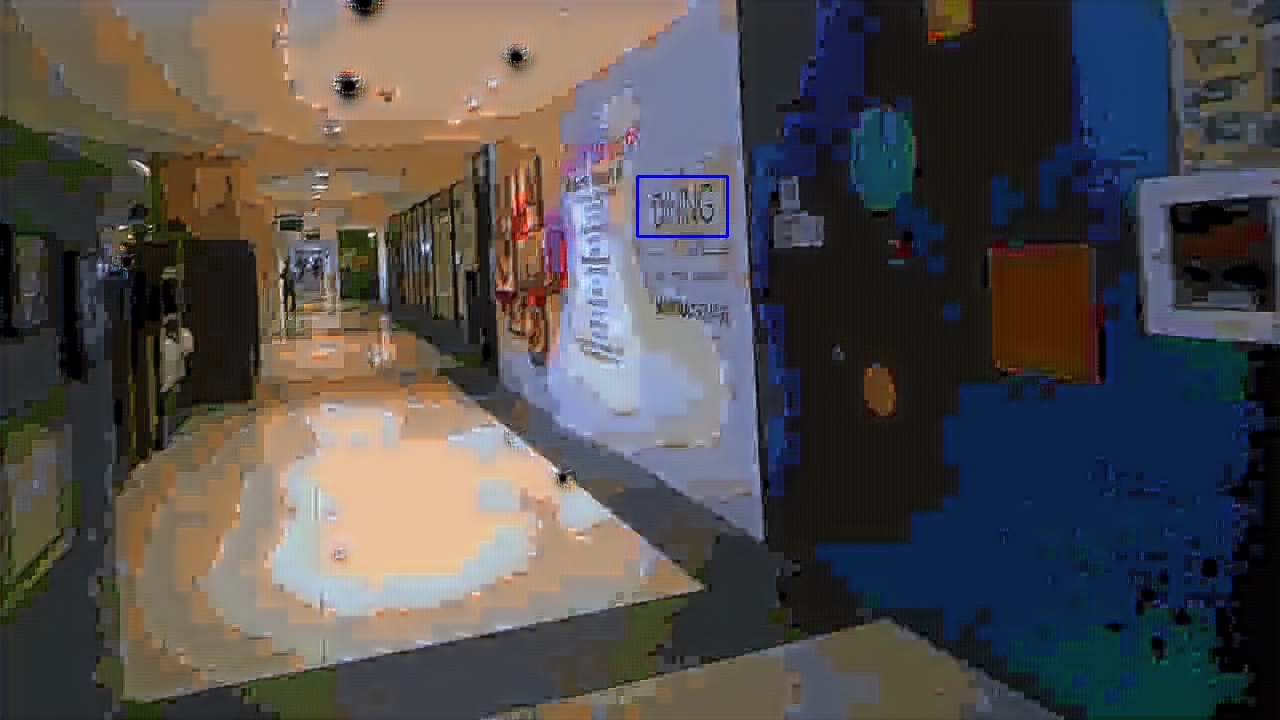}\\
        \includegraphics[width=\linewidth, height=0.8cm]{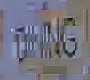}\\
        \centerline{\setstretch{3} \footnotesize \textcolor{red}{nil}} \\    
        \includegraphics[width=\linewidth, height=1.8cm]{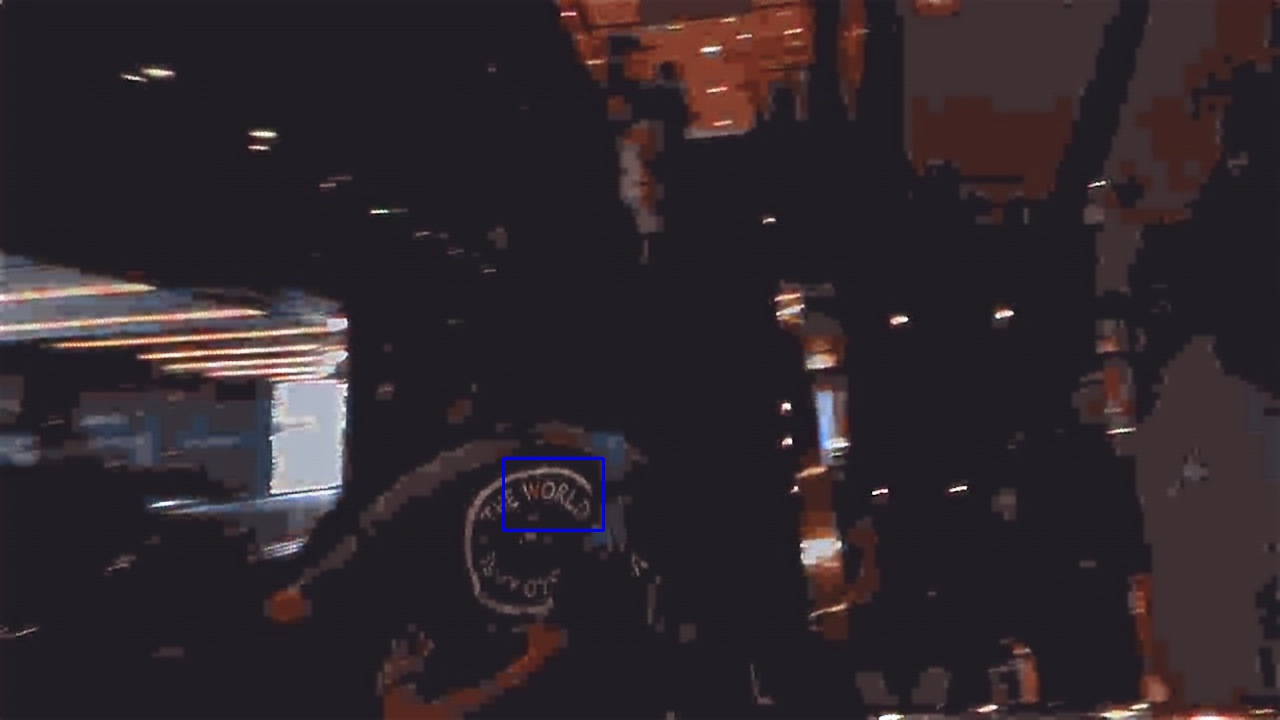}\\
        \includegraphics[width=\linewidth, height=0.8cm]{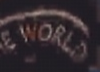}\\
        \centerline{\setstretch{3} \footnotesize \textcolor{red}{nil}} \\ 
    \end{tabular}
    \caption{CycleGAN~\cite{CycleGAN2017}}
    \label{fig:syn_sony_v1_cyclegan}
    \end{subfigure}\hspace{-0.5mm}
    \\[1ex]
    \begin{subfigure}[b]{0.24\textwidth} 
    \begin{tabular}[b]{@{}l@{}}
        \includegraphics[width=\linewidth, height=1.8cm]{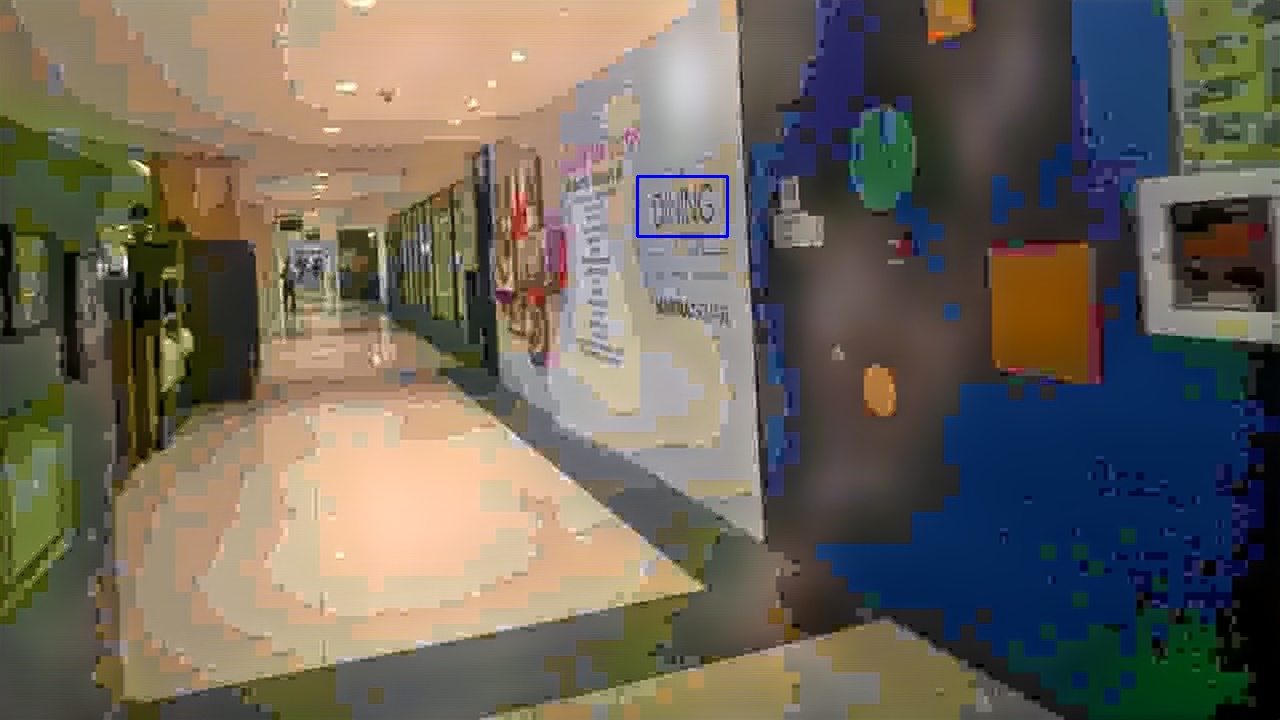}\\
        \includegraphics[width=\linewidth, height=0.8cm]{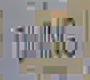}\\
        \centerline{\setstretch{3} \footnotesize \textcolor{red}{nil}} \\
        \includegraphics[width=\linewidth, height=1.8cm]{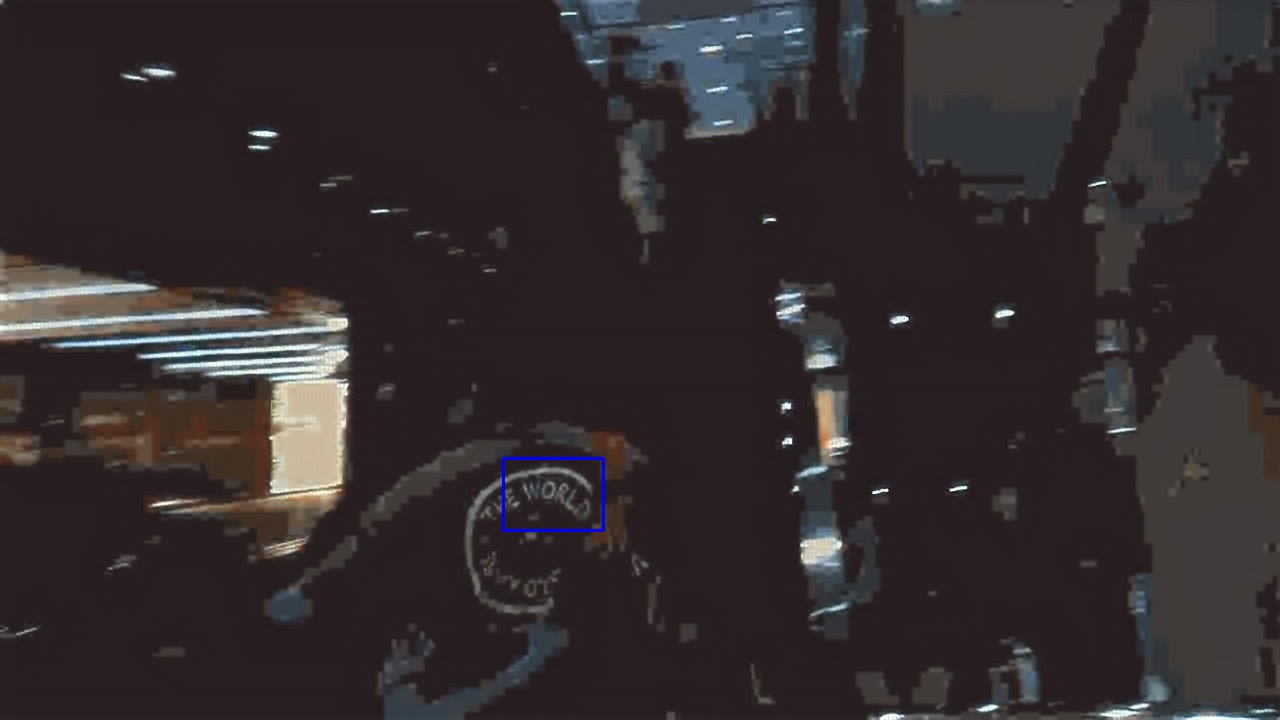}\\
        \includegraphics[width=\linewidth, height=0.8cm]{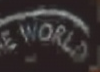}\\
        \centerline{\setstretch{3} \footnotesize \textcolor{red}{nil}} \\
    \end{tabular}
    \caption{EnlightenGAN~\cite{Jiang2019EnlightenGANDL}}
    \label{fig:syn_sony_v1_enlightengan}    
    \end{subfigure}
    \begin{subfigure}[b]{0.24\textwidth}
    \begin{tabular}[b]{@{}l@{}}
        \includegraphics[width=\linewidth, height=1.8cm]{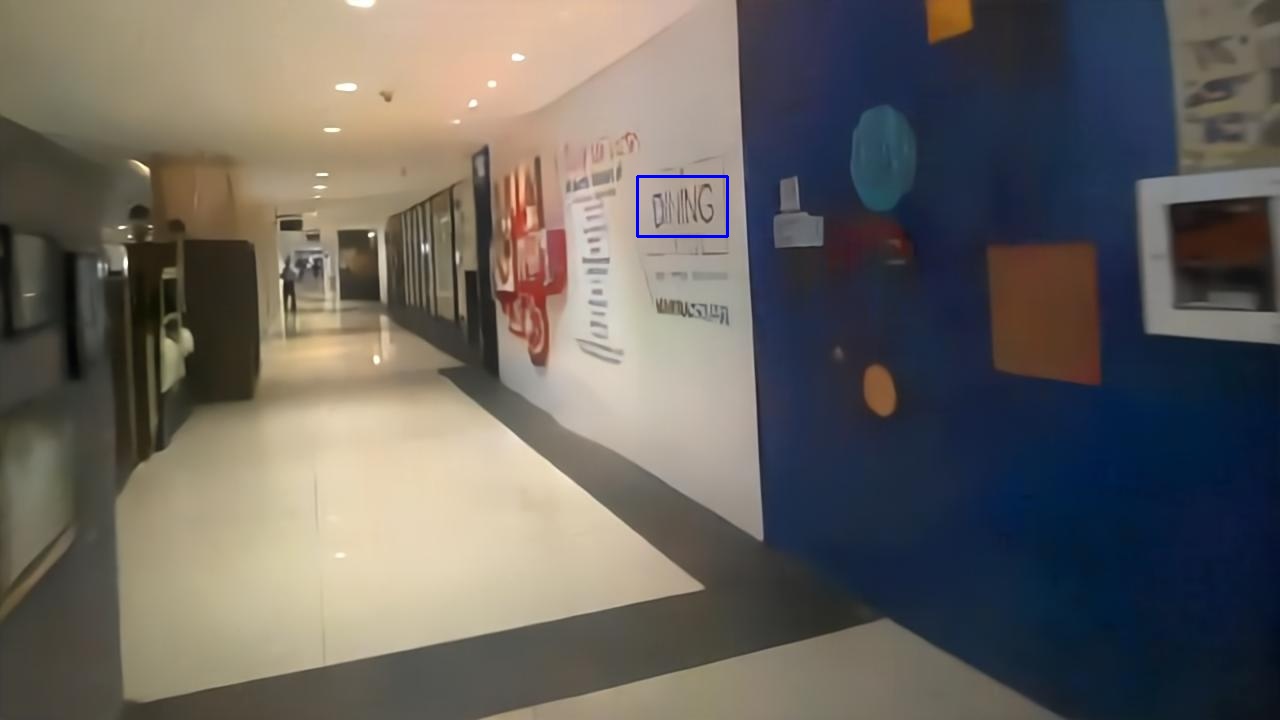}\\
        \includegraphics[width=\linewidth, height=0.8cm]{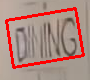}\\
        \centerline{\setstretch{3} \footnotesize \textcolor{red}{thing}} \\
        \includegraphics[width=\linewidth, height=1.8cm]{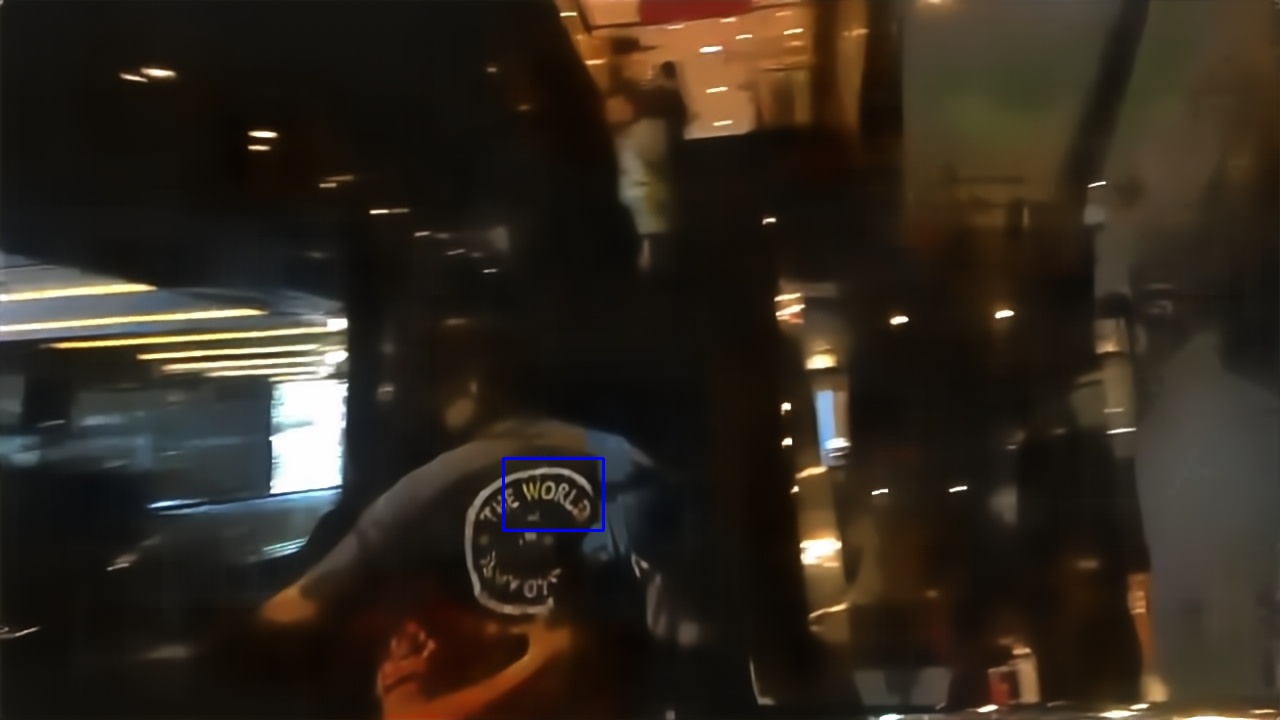}\\
        \includegraphics[width=\linewidth, height=0.8cm]{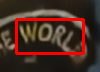}\\
        \centerline{\setstretch{3} \footnotesize \textcolor{red}{was}} \\ 
    \end{tabular}
    \caption{ELIE\_STR~\cite{hsu2022extremely}}    
    \label{fig:syn_sony_v1_elie}
    \end{subfigure}
    \begin{subfigure}[b]{0.24\textwidth}
    \begin{tabular}[b]{@{}l@{}}
        \includegraphics[width=\linewidth, height=1.8cm]{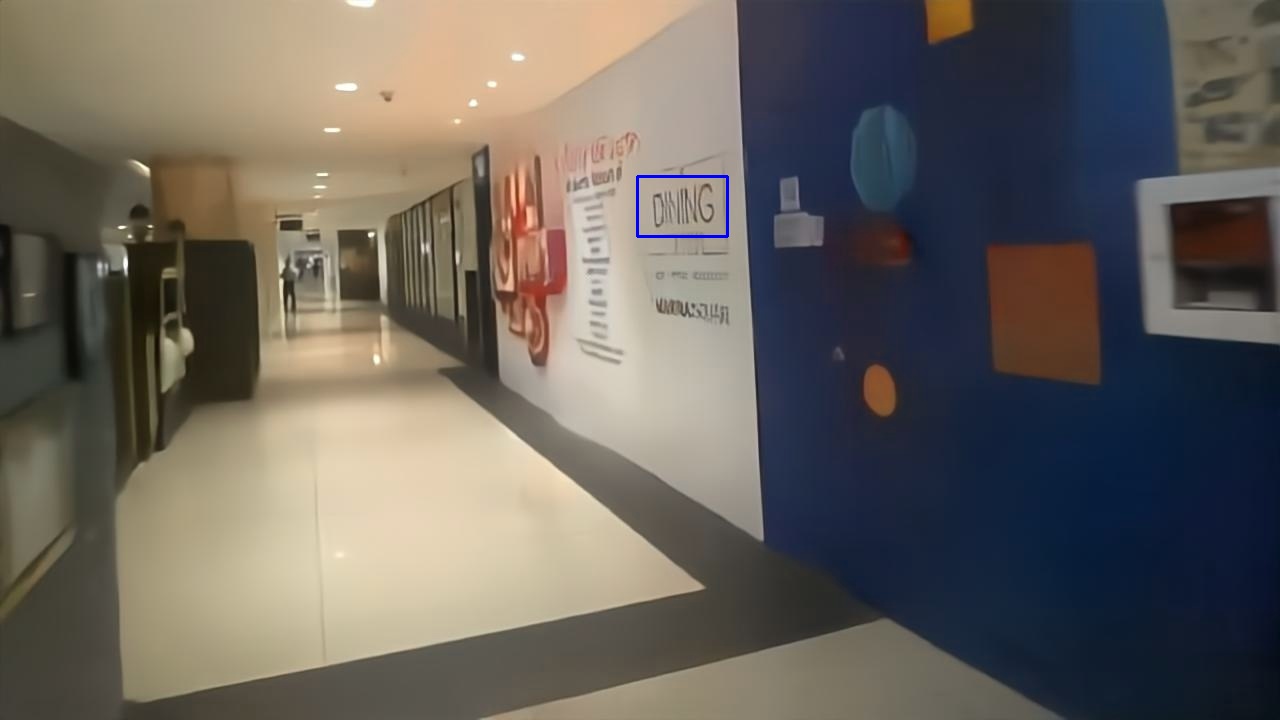}\\
        \includegraphics[width=\linewidth, height=0.8cm]{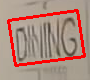}\\
        \centerline{\setstretch{3} \footnotesize \textcolor{green}{dining}} \\
        \includegraphics[width=\linewidth, height=1.8cm]{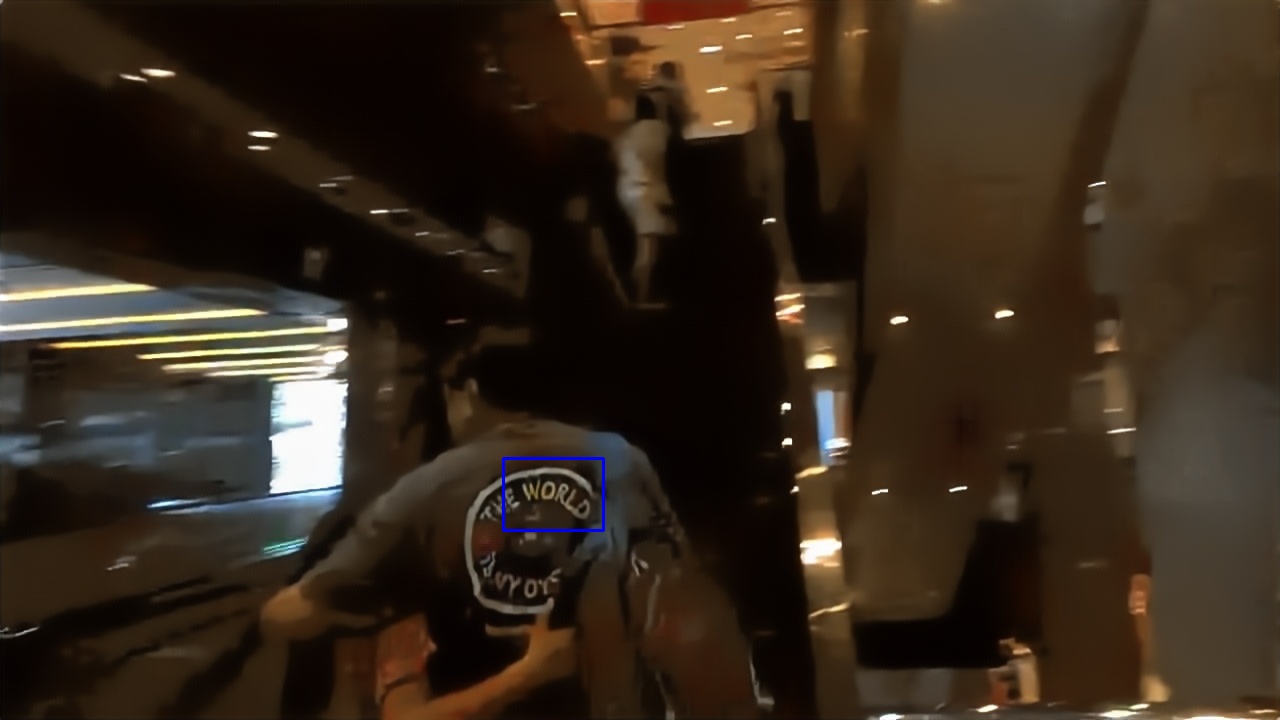}\\
        \includegraphics[width=\linewidth, height=0.8cm]{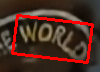}\\
        \centerline{\setstretch{3} \footnotesize \textcolor{green}{world}} \\
    \end{tabular}
    \caption{Ours}    
    \label{fig:syn_sony_v1_ours}
    \end{subfigure}
    \begin{subfigure}[b]{0.24\textwidth}
    \begin{tabular}[b]{@{}l@{}}
        \includegraphics[width=\linewidth, height=1.8cm]{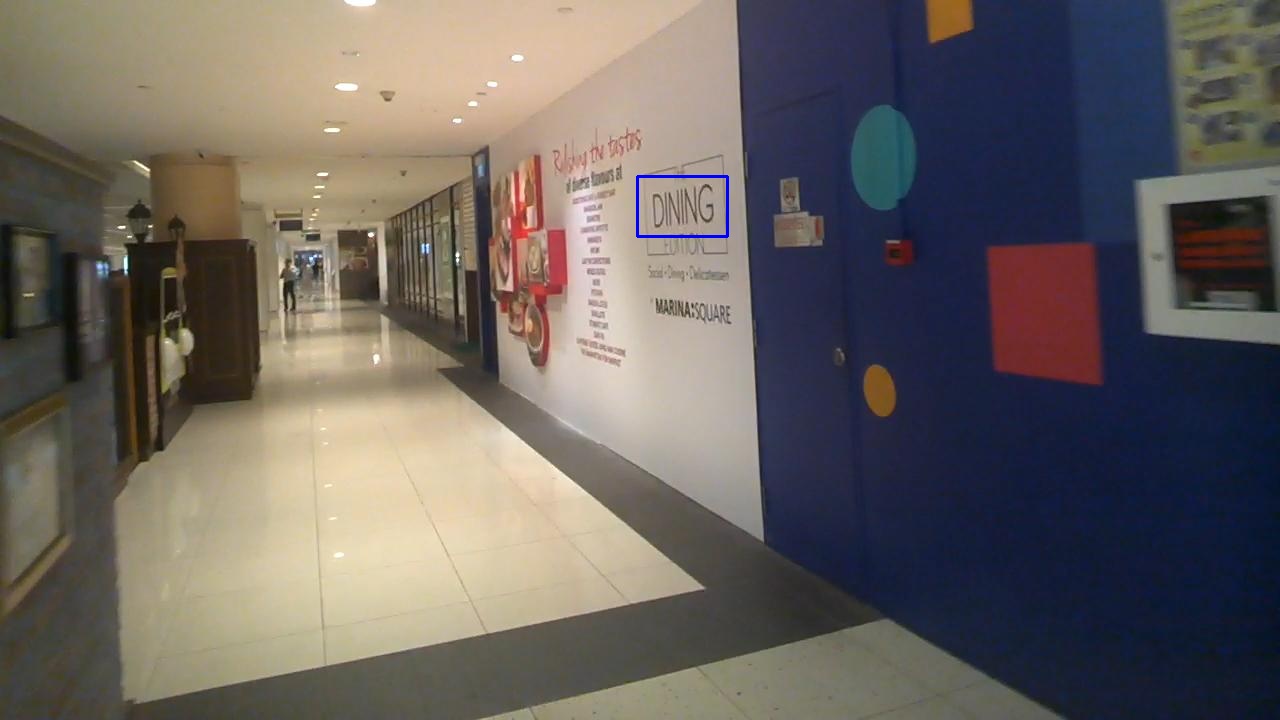}\\
        \includegraphics[width=\linewidth, height=0.8cm]{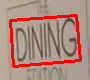}\\
        \centerline{\setstretch{3} \footnotesize \textcolor{green}{dining}} \\
        \includegraphics[width=\linewidth, height=1.8cm]{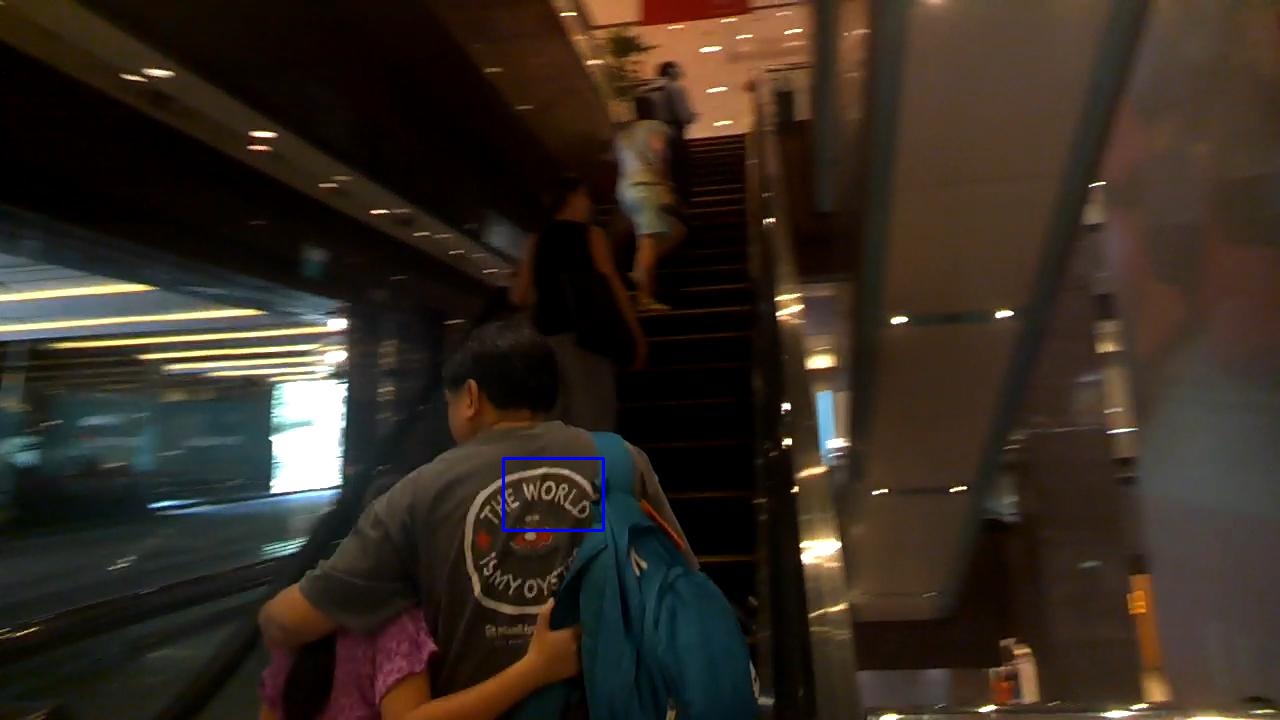}\\
        \includegraphics[width=\linewidth, height=0.8cm]{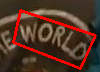}\\
        \centerline{\setstretch{3} \footnotesize \textcolor{green}{world}} \\
    \end{tabular}
    \caption{Ground Truth}
    \label{fig:syn_sony_v1_gt}     
    \end{subfigure}
    \caption{Comparison of our model with other state-of-the-art methods on the Syn-IC15-Sony-v1 (top) and Syn-IC15-Sony-v2 (bottom) dataset. PAN~\cite{wang2019efficient} is used as the text detector, while ASTER~\cite{Aster2019} is used for text recognition. Blue boxes represent the zoomed-in regions, while red boxes are drawn based on the predictions of PAN. Text recognition is performed on the word images cropped by using PAN's predictions (from column \ref{fig:syn_sony_v1_input} to column \ref{fig:syn_sony_v1_ours}) and ground truth text boxes (column \ref{fig:syn_sony_v1_gt}). \enquote{\textcolor{red}{nil}} stands for either no recognition results or invalid prediction boxes (with IoU $\leq$ 0.5) which are not considered for the subsequent recognition.}
    \label{fig:syn_IC15_by_sony_spotting}
\end{figure*}

\begin{figure*}[!htb]
    \centering
    \begin{subfigure}[b]{0.24\textwidth}
    \begin{tabular}[b]{@{}l@{}}
        \includegraphics[width=\linewidth, height=1.8cm]{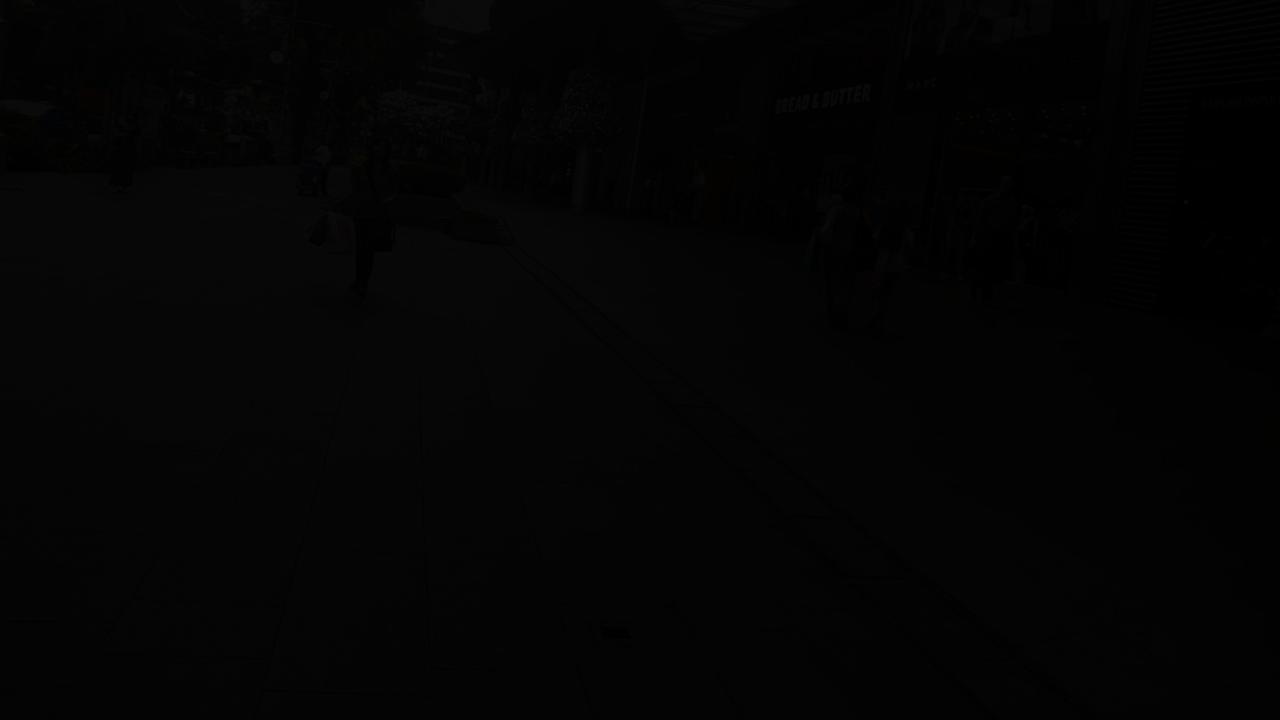}\\
        \includegraphics[width=\linewidth, height=1.3cm]{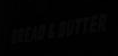}\\
        \centerline{\setstretch{3} \footnotesize \textcolor{red}{nil}} \\
        \includegraphics[width=\linewidth, height=1.8cm]{Images/ic15-fuji-v2/labelled_crop_356_input.jpg}\\
        \includegraphics[width=\linewidth, height=0.8cm]{Images/ic15-fuji-v2/cropped_356_input.png}\\
        \centerline{\setstretch{3} \footnotesize \textcolor{red}{nil}} \\
    \end{tabular}
    \caption{Low-Light}
    \label{fig:syn_fuji_v1_input}            
    \end{subfigure}\hspace{-0.5mm}
    \begin{subfigure}[b]{0.24\textwidth}   
    \begin{tabular}[b]{@{}l@{}}
        \includegraphics[width=\linewidth, height=1.8cm]{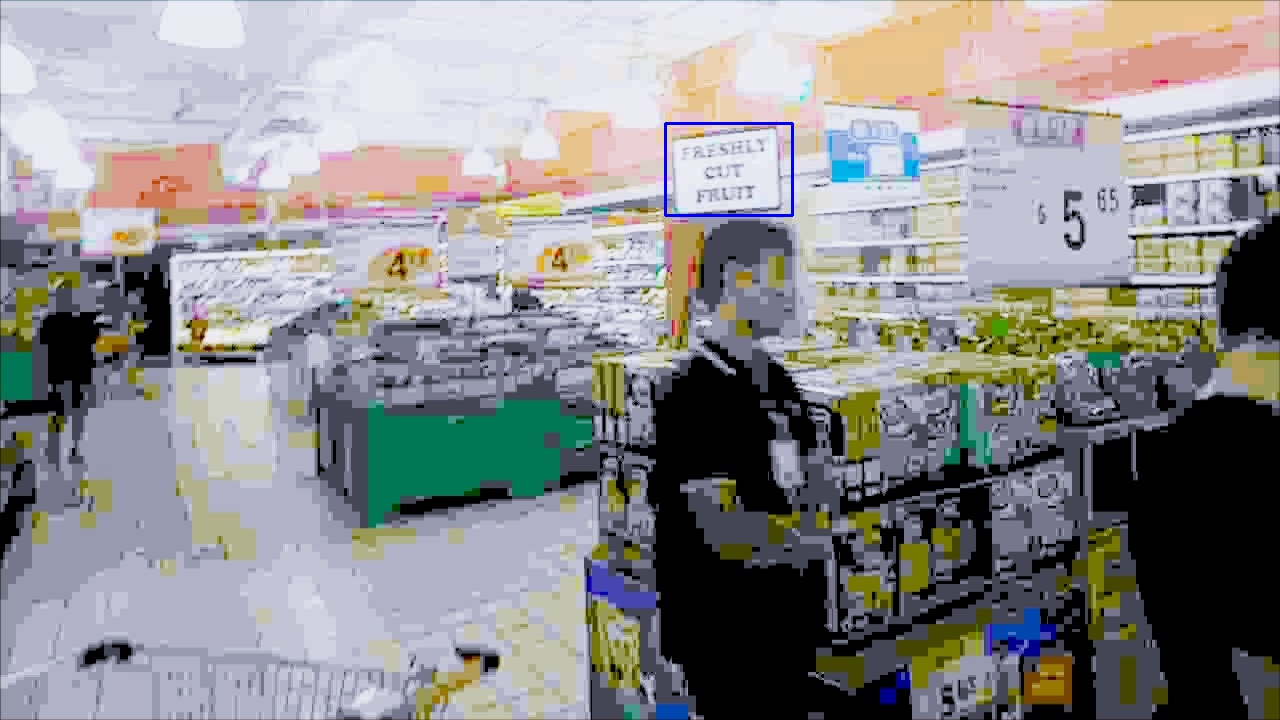}\\
        \includegraphics[width=\linewidth, height=1.3cm]{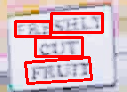}\\
        \centerline{\setstretch{3} \footnotesize \textcolor{red}{nil}, \textcolor{red}{the}, \textcolor{red}{for}} \\
        \includegraphics[width=\linewidth, height=1.8cm]{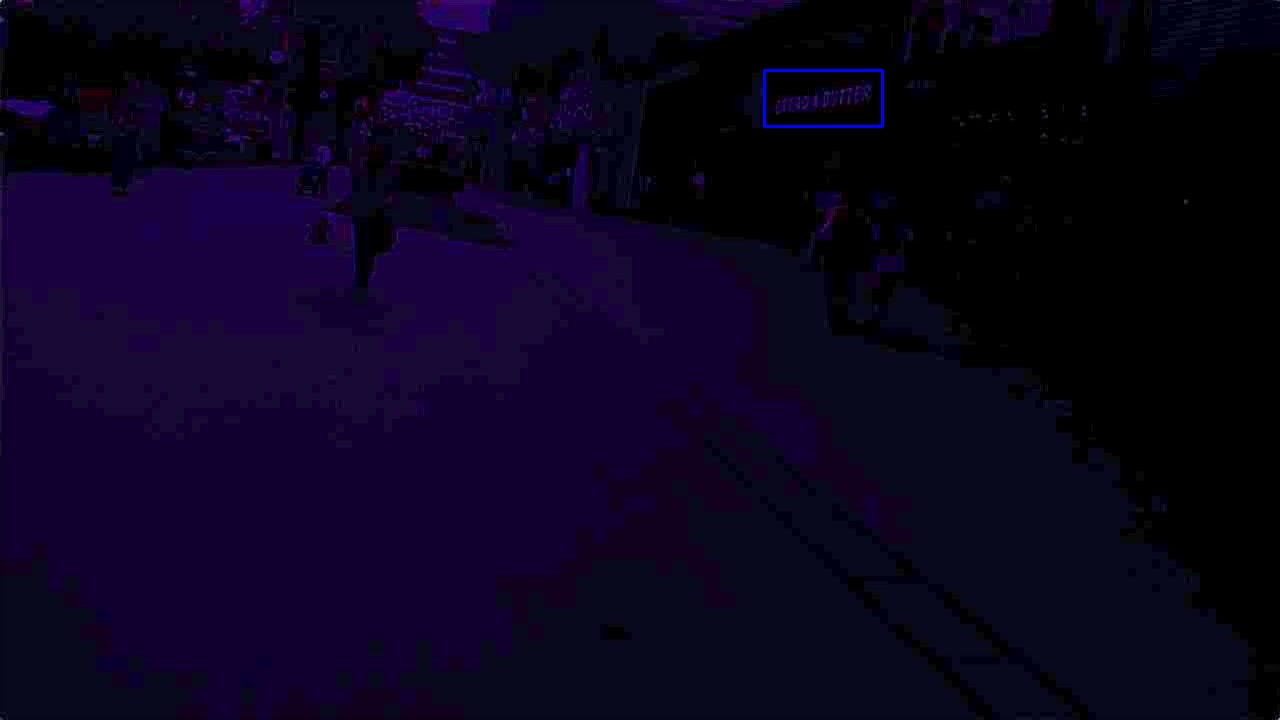}\\
        \includegraphics[width=\linewidth, height=0.8cm]{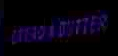}\\
        \centerline{\setstretch{3} \footnotesize \textcolor{red}{nil}} \\ 
    \end{tabular}
    \caption{Zero-DCE++~\cite{li2021learning}}    
    \label{fig:syn_fuji_v1_zero_dce_pp}
    \end{subfigure}\hspace{-0.5mm}
    \begin{subfigure}[b]{0.24\textwidth}
    \begin{tabular}[b]{@{}l@{}}
        \includegraphics[width=\linewidth, height=1.8cm]{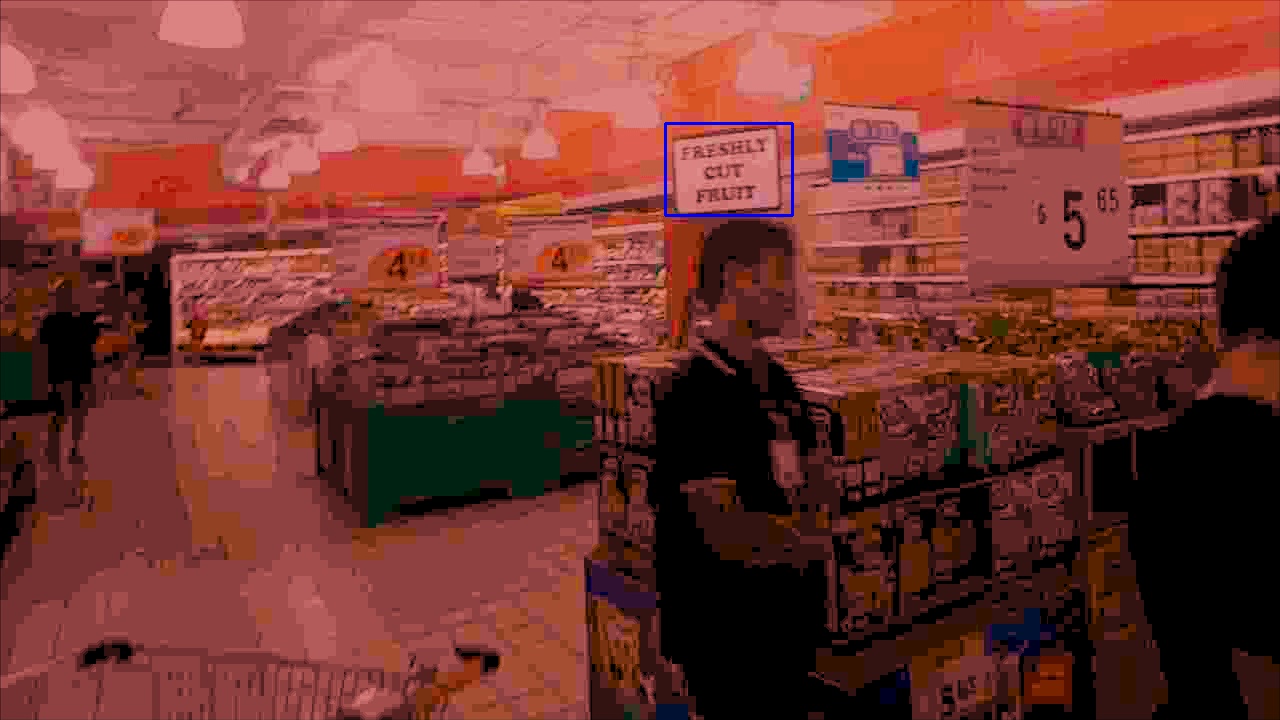}\\
        \includegraphics[width=\linewidth, height=1.3cm]{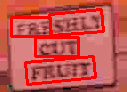}\\
        \centerline{\setstretch{3} \footnotesize \textcolor{red}{nil}, \textcolor{red}{out}, \textcolor{red}{from}} \\
        \includegraphics[width=\linewidth, height=1.8cm]{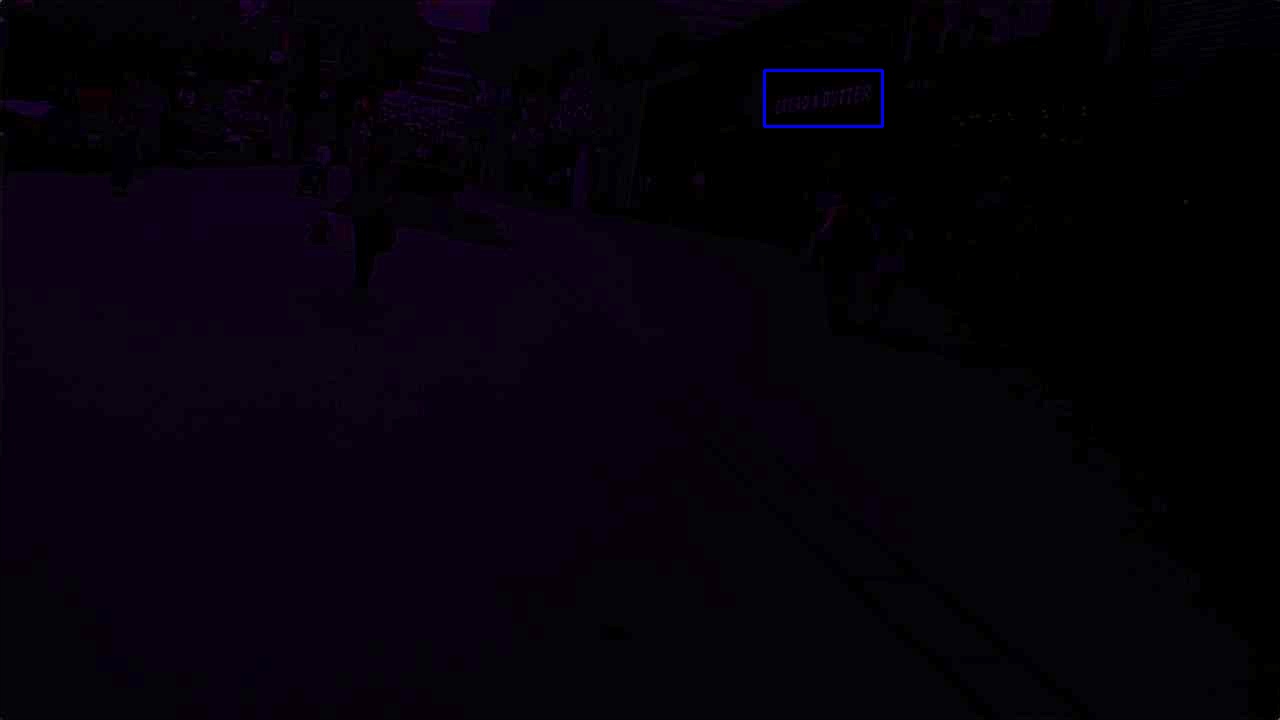}\\
        \includegraphics[width=\linewidth, height=0.8cm]{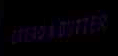}\\
        \centerline{\setstretch{3} \footnotesize \textcolor{red}{nil}} \\
    \end{tabular}
    \caption{SCI~\cite{ma2022toward}}    
    \label{fig:syn_fuji_v1_sci}
    \end{subfigure}\hspace{-0.5mm}
    \begin{subfigure}[b]{0.24\textwidth}
    \begin{tabular}[b]{@{}l@{}}
        \includegraphics[width=\linewidth, height=1.8cm]{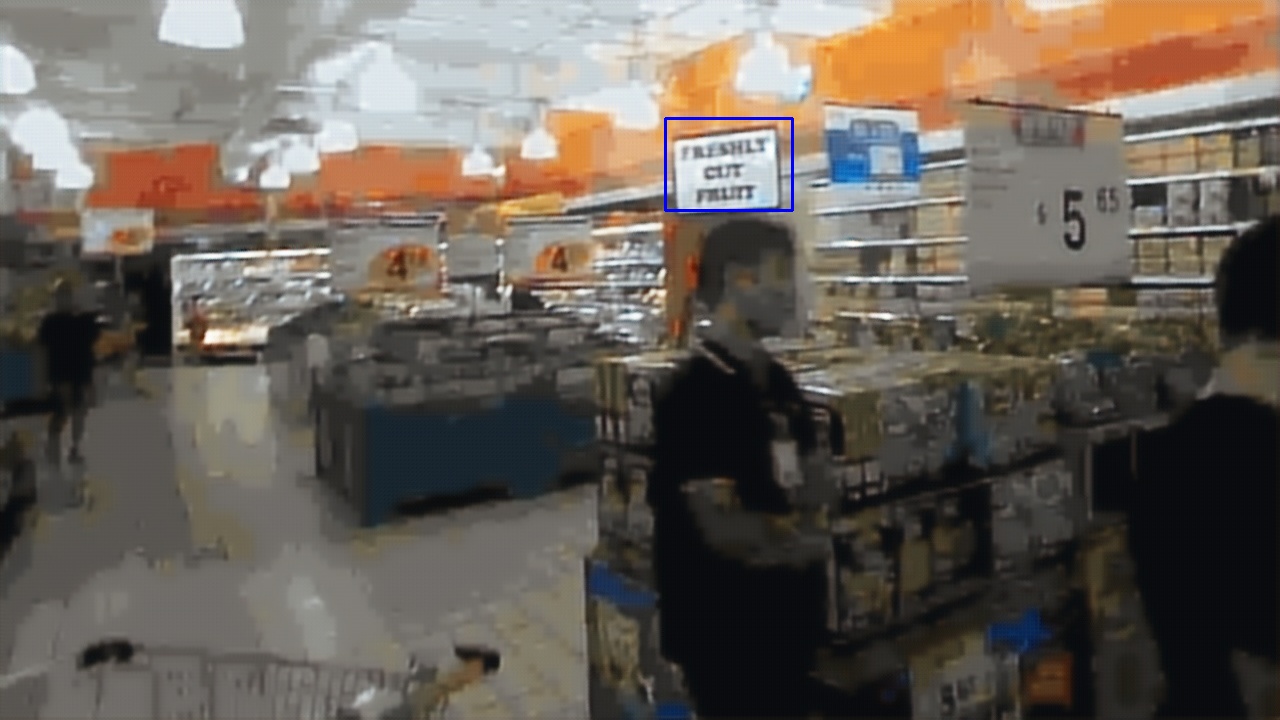}\\
        \includegraphics[width=\linewidth, height=1.3cm]{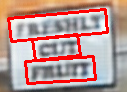}\\
        \centerline{\setstretch{3} \footnotesize \textcolor{red}{from}, \textcolor{red}{an}, \textcolor{red}{nil}} \\
        \includegraphics[width=\linewidth, height=1.8cm]{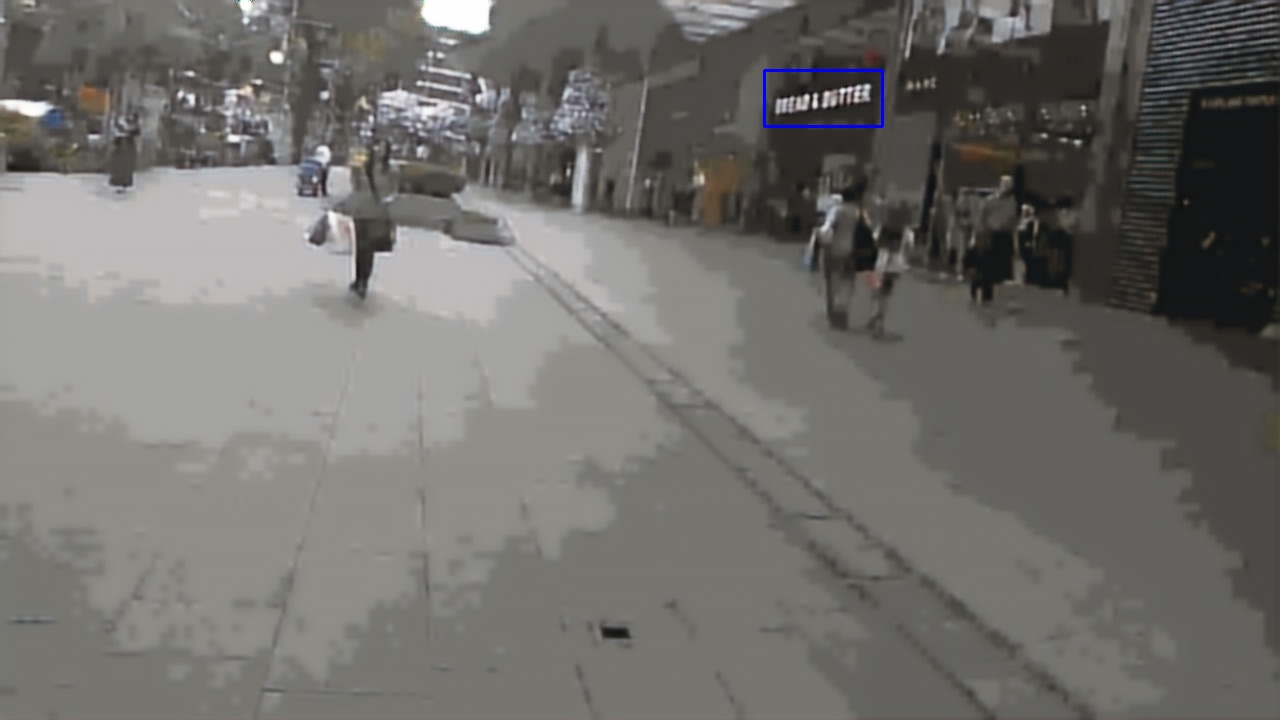}\\
        \includegraphics[width=\linewidth, height=0.8cm]{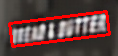}\\
        \centerline{\setstretch{3} \footnotesize \textcolor{red}{nil}} \\ 
    \end{tabular}
    \caption{CycleGAN~\cite{CycleGAN2017}}    
    \label{fig:syn_fuji_v1_cyclegan}
    \end{subfigure}\hspace{-0.5mm}
    \\[1ex]
    \begin{subfigure}[b]{0.24\textwidth}
    \begin{tabular}[b]{@{}l@{}}
        \includegraphics[width=\linewidth, height=1.8cm]{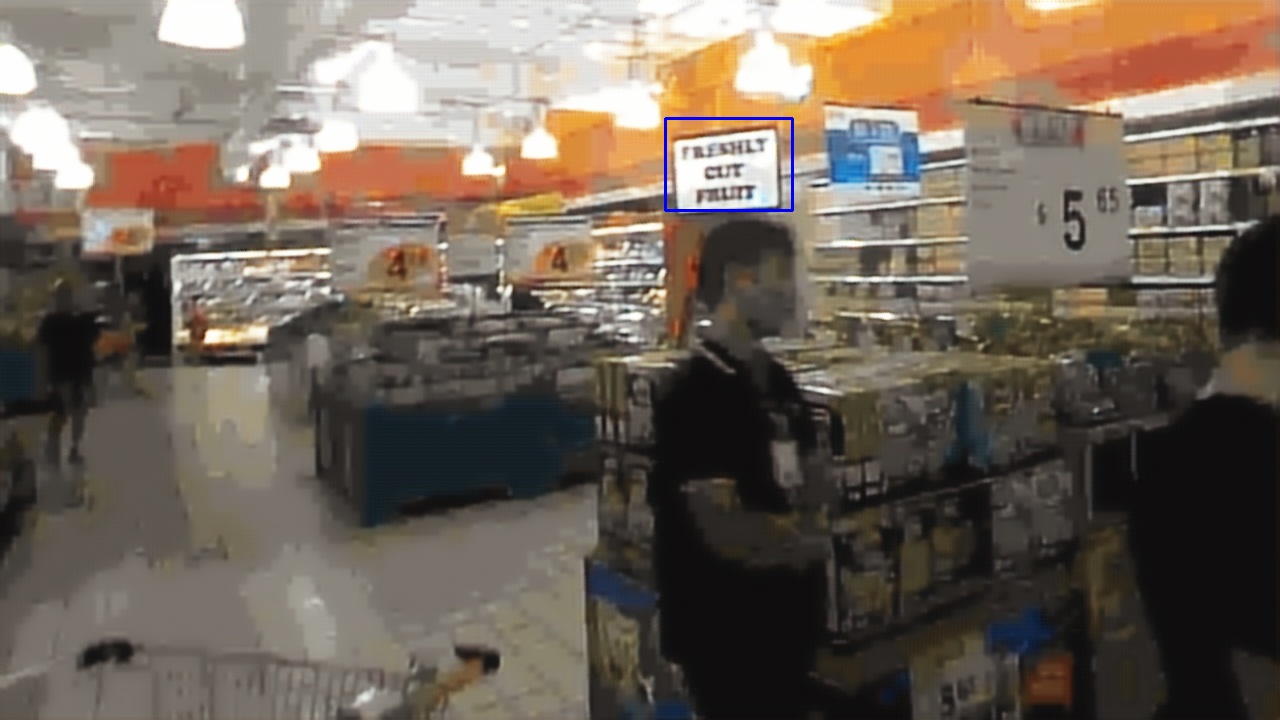}\\
        \includegraphics[width=\linewidth, height=1.3cm]{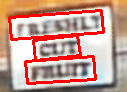}\\
        \centerline{\setstretch{3} \footnotesize \textcolor{red}{and}, \textcolor{green}{cut}, \textcolor{red}{but}} \\
        \includegraphics[width=\linewidth, height=1.8cm]{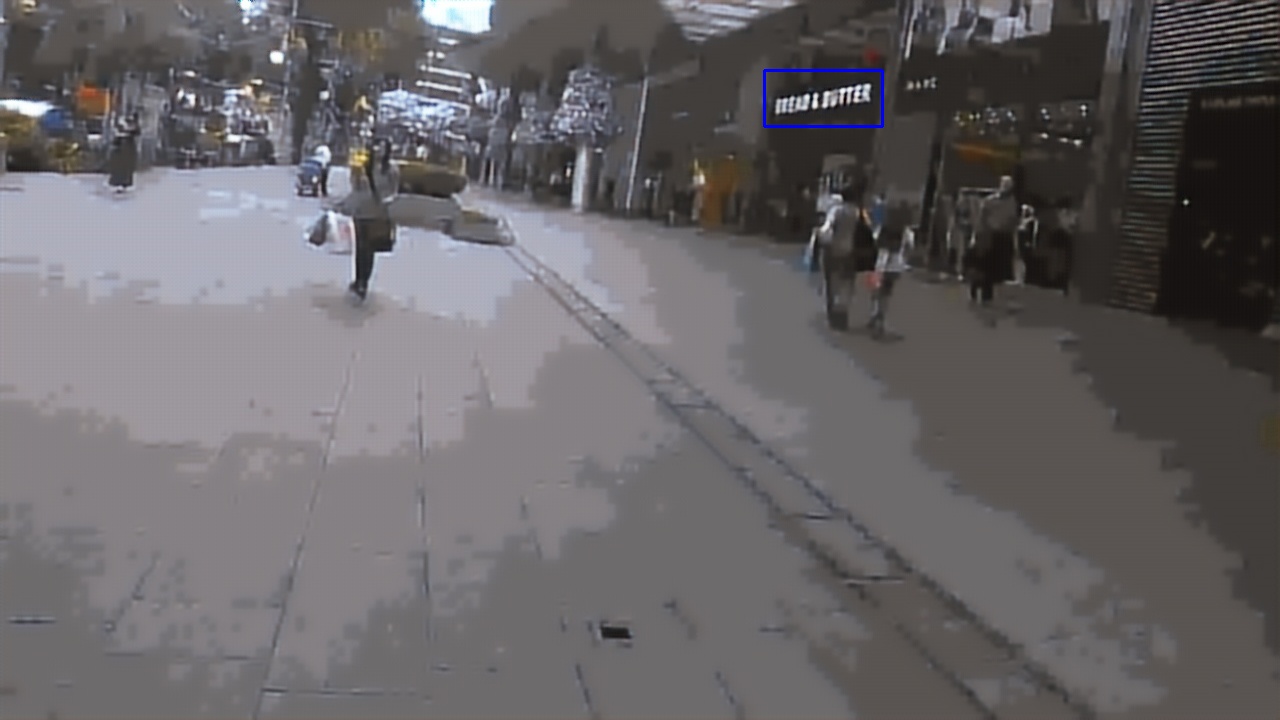}\\
        \includegraphics[width=\linewidth,height=0.8cm]{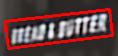}\\
        \centerline{\setstretch{3} \footnotesize \textcolor{red}{nil}} \\
    \end{tabular}
    \caption{EnlightenGAN~\cite{Jiang2019EnlightenGANDL}}    
    \label{fig:syn_fuji_v1_enlightengan}    
    \end{subfigure}
    \begin{subfigure}[b]{0.24\textwidth} 
    \begin{tabular}[b]{@{}l@{}}
        \includegraphics[width=\linewidth, height=1.8cm]{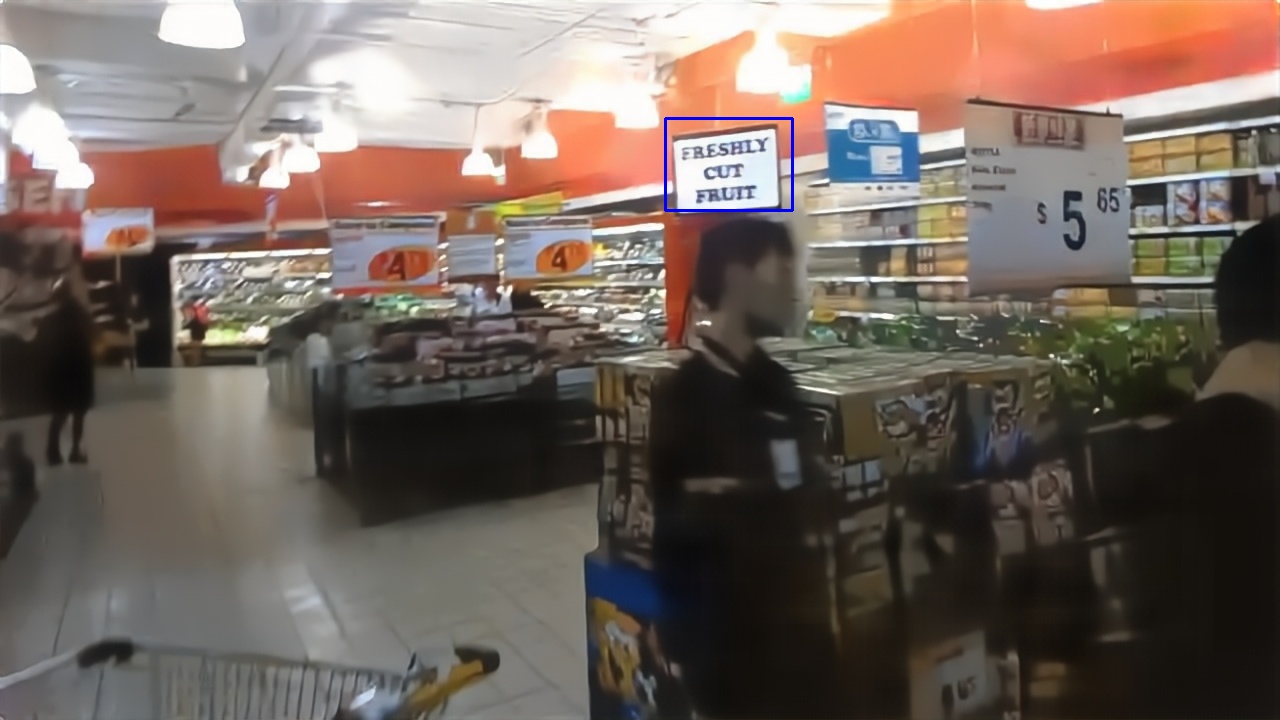}\\
        \includegraphics[width=\linewidth, height=1.3cm]{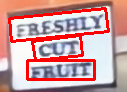}\\
        \centerline{\setstretch{3} \footnotesize \textcolor{red}{ereshly}, \textcolor{green}{cut}, \textcolor{red}{and} } \\
        \includegraphics[width=\linewidth, height=1.8cm]{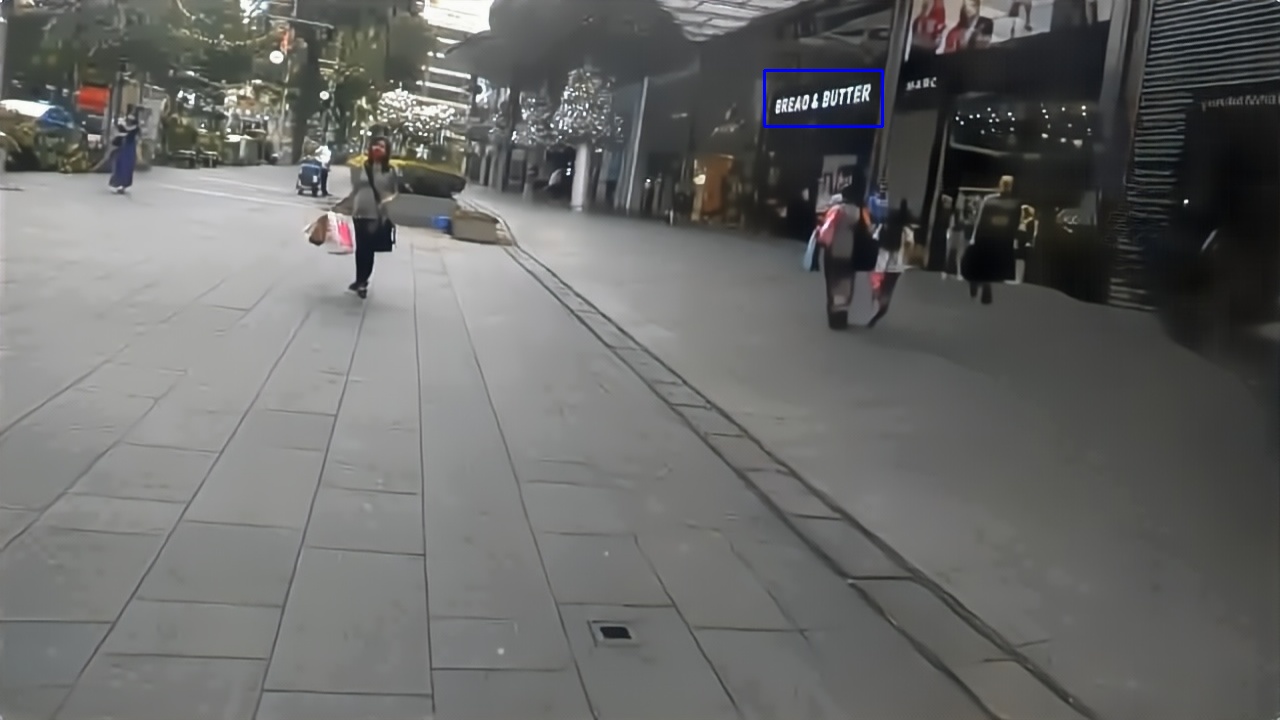}\\
        \includegraphics[width=\linewidth, height=0.8cm]{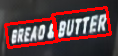}\\
        \centerline{\setstretch{3} \footnotesize \textcolor{red}{greads}, \textcolor{green}{butter}} \\
    \end{tabular}
    \caption{ELIE\_STR~\cite{hsu2022extremely}}
    \label{fig:syn_fuji_v1_elie}
    \end{subfigure}
    \begin{subfigure}[b]{0.24\textwidth}
    \begin{tabular}[b]{@{}l@{}}
        \includegraphics[width=\linewidth, height=1.8cm]{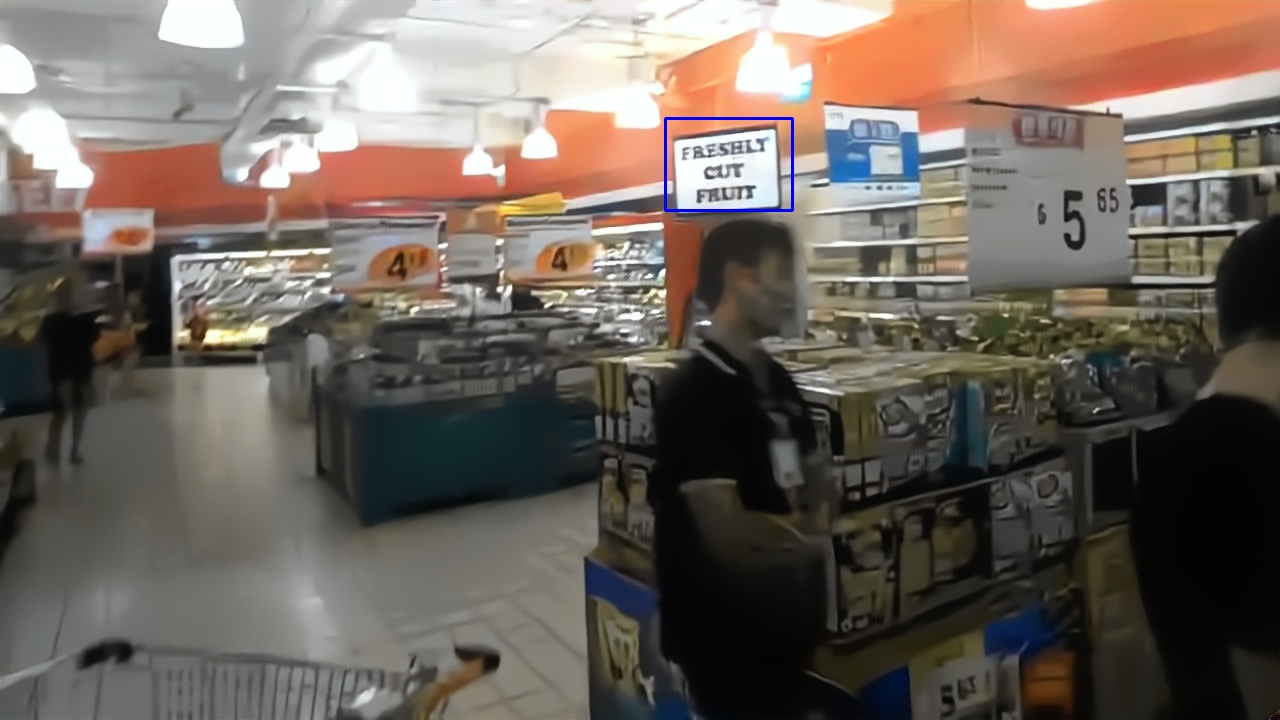}\\
        \includegraphics[width=\linewidth, height=1.3cm]{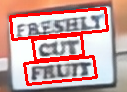}\\
        \centerline{\setstretch{3} \footnotesize \textcolor{green}{freshly}, \textcolor{green}{cut}, \textcolor{red}{from}} \\
        \includegraphics[width=\linewidth, height=1.8cm]{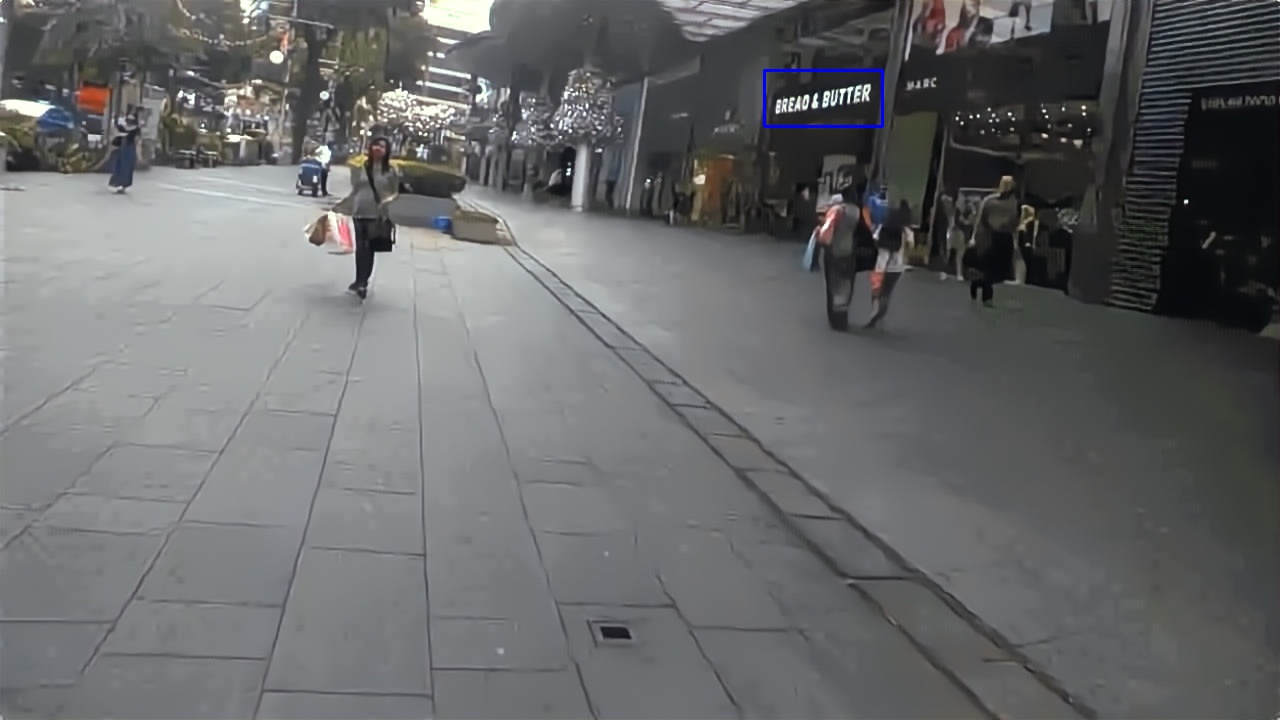}\\
        \includegraphics[width=\linewidth, height=0.8cm]{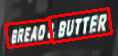}\\
        \centerline{\setstretch{3} \footnotesize \textcolor{green}{bread}, \textcolor{green}{butter}} \\
        \end{tabular}
    \caption{Ours}
    \label{fig:syn_fuji_v1_ours}    
    \end{subfigure}
    \begin{subfigure}[b]{0.24\textwidth}
    \begin{tabular}[b]{@{}l@{}}
        \includegraphics[width=\linewidth, height=1.8cm]{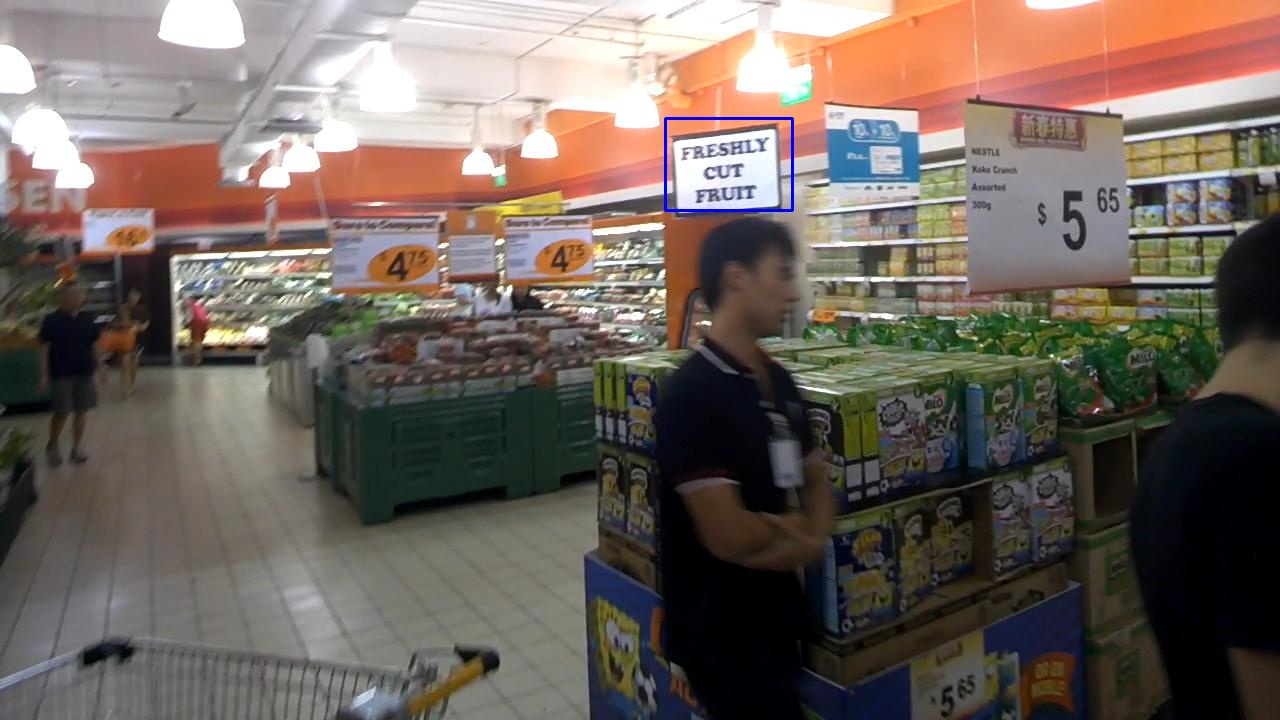}\\
        \includegraphics[width=\linewidth, height=1.3cm]{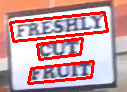}\\
        \centerline{\setstretch{3} \footnotesize \textcolor{green}{freshly}, \textcolor{green}{cut}, \textcolor{green}{fruit}} \\
        \includegraphics[width=\linewidth, height=1.8cm]{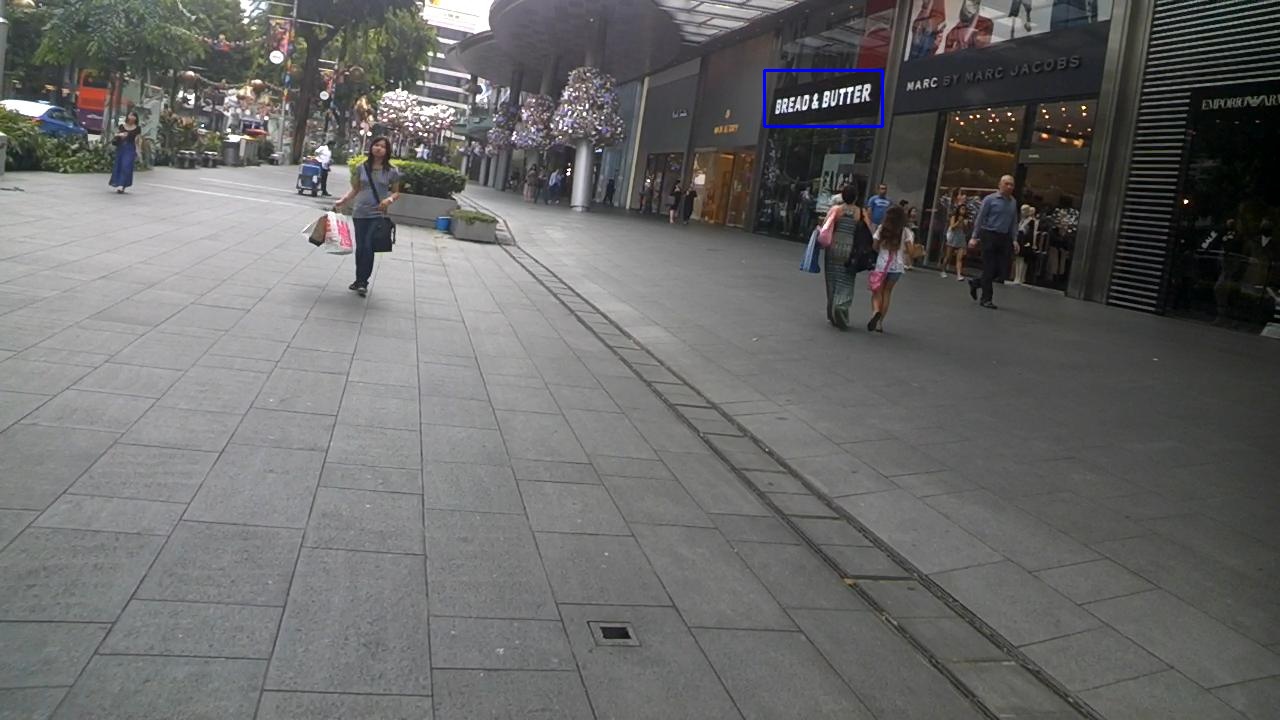}\\
        \includegraphics[width=\linewidth, height=0.8cm]{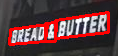}\\
        \centerline{\setstretch{3} \footnotesize \textcolor{green}{bread}, \textcolor{green}{butter}} \\
    \end{tabular}
    \caption{Ground Truth}
    \label{fig:syn_fuji_v1_gt}   
    \end{subfigure}
    \caption{Comparison of our model with other state-of-the-art methods on the Syn-IC15-Fuji-v1 (top) and Syn-IC15-Fuji-v2 (bottom) datasets. PAN~\cite{wang2019efficient} is used as the text detector, while ASTER~\cite{Aster2019} is used for text recognition. Blue boxes represent the zoomed-in regions, while red boxes are drawn based on the predictions of PAN. Text recognition is performed on the word images cropped by using PAN's predictions (from column \ref{fig:syn_fuji_v1_input} to column \ref{fig:syn_fuji_v1_ours}) and ground truth text boxes (column \ref{fig:syn_fuji_v1_gt}). \enquote{\textcolor{red}{nil}} stands for either no recognition results or invalid prediction boxes (with IoU $\leq$ 0.5) which are not considered for the subsequent recognition.}
    \label{fig:syn_IC15_by_fuji_spotting}
\end{figure*}

As for the qualitative results of the two-stage text spotting task, we present comprehensive results in Figure~\ref{fig:syn_IC15_by_sony_spotting} and Figure~\ref{fig:syn_IC15_by_fuji_spotting}. In the first row, we display the entire image to provide an overview of the scene. In the subsequent row, we zoom in on regions of interest that contain text instances, allowing for a closer examination of text spotting performance. Please note that text recognition is performed only when the IoU of detected boxes with respect to ground truth bounding boxes is larger than 0.5. Any recognition results below this threshold are considered failed text recognition and labeled as ``nil''. Additionally, correctly predicted words are colored in green, while incorrect text recognition results are indicated in red.

Existing methods often suffer from introducing excessive noise and artifacts when applying global and local image enhancement techniques simultaneously. As a result, enhanced images produced by these methods exhibit high pixelation and visual distortions. In contrast, our model generates enhanced images that are significantly less noisy and exhibit fewer artifacts, even in local text regions. While it is true that some text recognition results on our enhanced images may not be perfect, our results consistently exhibit the closest resemblance to ground truth compared to other methods.

In summary, our model can generate globally well-enhanced images given extremely low-light images while retaining important local details such as text features, leading to better text detection and recognition results.

\begin{table*}[!htbp]
\setlength\tabcolsep{1pt}
\centering
\begin{adjustbox}{width=\linewidth,center}
\begin{tabular}{c?{0.8mm}llccccccccc}
\cmidrule[2pt]{1-12}
\multicolumn{1}{c}{\multirow{2}{*}{}} & \multirow{2}[2]{*}{Type} & \multirow{2}[2]{*}{Method} & \multicolumn{3}{c}{Image Quality} & \multicolumn{2}{c}{H-Mean} & \multicolumn{4}{c}{Case-Insensitive Accuracy} \\ \cmidrule(lr){4-6} \cmidrule(lr){7-8} \cmidrule(lr){9-12}
\multicolumn{1}{c}{} & & & PSNR $\uparrow$ & SSIM $\uparrow$ & LPIPS $\downarrow$ & C $\uparrow$ & P $\uparrow$ & C+T $\uparrow$ & C+A $\uparrow$ & P+T $\uparrow$ & P+A $\uparrow$ \\ \cmidrule[2pt]{1-12}

\multirow{8}{*}{\rotatebox[origin=c]{90}{\textbf{Syn-IC15-Sony-v1}}} & & Input & - & - & - & 0.355 & 0.192 & 0.114 & 0.118 & 0.062 & 0.067 \\ \cline{2-12}
& \multirow{2}{*}{ZSL} & Zero-DCE ++~\cite{li2021learning} & 13.505 & 0.575 & 0.587 & 0.379 & 0.329 & 0.101 & 0.093 & 0.089 & 0.079 \\
& & SCI~\cite{ma2022toward} & 14.195 & 0.589 & 0.486 & 0.479 & 0.443 & 0.154 & 0.132 & 0.133 & 0.121 \\ \cline{2-12}
& \multirow{2}{*}{UL} & CycleGAN~\cite{CycleGAN2017} & 23.420 & 0.720 & 0.397 & 0.428 & 0.458 & 0.119 & 0.144 & 0.122 & 0.138 \\
& & EnlightenGAN~\cite{Jiang2019EnlightenGANDL} & 21.030 & 0.661 & 0.469 & 0.458 & 0.461 & 0.140 & 0.157 & 0.123 & 0.139 \\ \cline{2-12}
& \multirow{2}{*}{SL} & ELIE\_STR~\cite{hsu2022extremely} & 28.410 & 0.840 & 0.253 & 0.622 & 0.631 & 0.193 & 0.219 & 0.197 & 0.221 \\
& & Ours & \textbf{28.568} & \textbf{0.859} & \textbf{0.248} & \textbf{0.660} & \textbf{0.662} & \textbf{0.222} & \textbf{0.256} & \textbf{0.221} & \textbf{0.248} \\ \cline{2-12}
& & GT & - & - & - & 0.800 & 0.830 & 0.526 & 0.584 & 0.555 & 0.591 \\ \cmidrule[2pt]{1-12}

\multirow{8}{*}{\rotatebox[origin=c]{90}{\textbf{Syn-IC15-Sony-v2}}} & & Input & - & - & - & 0.198 & 0.068 & 0.046 & 0.047 & 0.018 & 0.019 \\ \cline{2-12}
& \multirow{2}{*}{ZSL} & Zero-DCE ++~\cite{li2021learning} & 8.189 & 0.063 & 0.753 & 0.259 & 0.238 & 0.059 & 0.049 & 0.051 & 0.044 \\
& & SCI~\cite{ma2022toward} & 9.055 & 0.111 & 0.711 & 0.266 & 0.278 & 0.066 & 0.052 & 0.066 & 0.055 \\ \cline{2-12}
& \multirow{2}{*}{UL} & CycleGAN~\cite{CycleGAN2017} & 20.212 & 0.768 & 0.404 & 0.443 & 0.429 & 0.113 & 0.114 & 0.104 & 0.107 \\
& & EnlightenGAN~\cite{Jiang2019EnlightenGANDL} & 19.610 & 0.753 & 0.440 & 0.477 & 0.456 & 0.116 & 0.114 & 0.099 & 0.104 \\ \cline{2-12}
& \multirow{2}{*}{SL} & ELIE\_STR~\cite{hsu2022extremely} & 22.724 & 0.792 & 0.339 & 0.570 & 0.566 & 0.126 & 0.136 & 0.122 & 0.130 \\
& & Ours & \textbf{22.938} & \textbf{0.809} & \textbf{0.318} & \textbf{0.593} & \textbf{0.577} & \textbf{0.138} & \textbf{0.157} & \textbf{0.134} & \textbf{0.143} \\ \cline{2-12}
& & GT & - & - & - & 0.800 & 0.830 & 0.526 & 0.584 & 0.555 & 0.591 \\ \cmidrule[2pt]{1-12}

\multirow{8}{*}{\rotatebox[origin=c]{90}{\textbf{Syn-IC15-Fuji-v1}}} & & Input & - & - & - & 0.347 & 0.175 & 0.100 & 0.102 & 0.097 & 0.095 \\ \cline{2-12}
& \multirow{2}{*}{ZSL} & Zero-DCE ++~\cite{li2021learning} & 13.124 & 0.569 & 0.591 & 0.379 & 0.323 & 0.099 & 0.092 & 0.094 & 0.083 \\
& & SCI~\cite{ma2022toward} & 15.426 & 0.615 & 0.599 & 0.468 & 0.463 & 0.144 & 0.129 & 0.139 & 0.124 \\ \cline{2-12}
& \multirow{2}{*}{UL} & CycleGAN~\cite{CycleGAN2017} & 24.476 & 0.797 & 0.377 & 0.406 & 0.371 & 0.109 & 0.113 & 0.097 & 0.104 \\
& & EnlightenGAN~\cite{Jiang2019EnlightenGANDL} & 25.923 & 0.803 & 0.364 & 0.420 & 0.406 & 0.115 & 0.121 & 0.104 & 0.106 \\ \cline{2-12}
& \multirow{2}{*}{SL} & ELIE\_STR~\cite{hsu2022extremely} & 27.528 & 0.837 & 0.264 & 0.646 & 0.645 & 0.199 & 0.216 & 0.193 & 0.200 \\
& & Ours & \textbf{28.437} & \textbf{0.856} & \textbf{0.248} & \textbf{0.668} & \textbf{0.668} & \textbf{0.218} & \textbf{0.234} & \textbf{0.211} & \textbf{0.217} \\ \cline{2-12}
& & GT & - & - & - & 0.800 & 0.830 & 0.526 & 0.584 & 0.555 & 0.591 \\ \cmidrule[2pt]{1-12}

\multirow{8}{*}{\rotatebox[origin=c]{90}{\textbf{Syn-IC15-Fuji-v2}}} & & Input & - & - & - & 0.133 & 0.028 & 0.023 & 0.022 & 0.006 & 0.009 \\ \cline{2-12}
& \multirow{2}{*}{ZSL} & Zero-DCE ++~\cite{li2021learning} & 8.793 & 0.134 & 0.708 & 0.193 & 0.137 & 0.040 & 0.033 & 0.028 & 0.025 \\
& & SCI~\cite{ma2022toward} & 8.071 & 0.069 & 0.789 & 0.213 & 0.128 & 0.048 & 0.042 & 0.034 & 0.030 \\ \cline{2-12}
& \multirow{2}{*}{UL} & CycleGAN~\cite{CycleGAN2017} & 23.920 & 0.733 & 0.367 & 0.397 & 0.386 & 0.070 & 0.074 & 0.066 & 0.073 \\
& & EnlightenGAN~\cite{Jiang2019EnlightenGANDL} & 24.600 & 0.740 & 0.389 & 0.415 & 0.396 & 0.089 & 0.091 & 0.081 & 0.084 \\ \cline{2-12}
& \multirow{2}{*}{SL} & ELIE\_STR~\cite{hsu2022extremely} & 25.536 & 0.819 & 0.301 & 0.556 & 0.546 & 0.121 & 0.133 & 0.117 & 0.127 \\
& & Ours & \textbf{25.999} & \textbf{0.830} & \textbf{0.283} & \textbf{0.598} & \textbf{0.583} & \textbf{0.143} & \textbf{0.152} & \textbf{0.128} & \textbf{0.144} \\ \cline{2-12}
& & GT & - & - & - & 0.800 & 0.830 & 0.526 & 0.584 & 0.555 & 0.591 \\ \cmidrule[2pt]{1-12}

\end{tabular}
\end{adjustbox}
\caption{Quantitative results of PSNR, SSIM, LPIPS, and text detection H-Mean for low-light image enhancement methods on Syn-IC15-Sony-v1, and Syn-IC15-Sony-v2, Syn-IC15-Fuji-v1, and Syn-IC15-Fuji-v2 datasets. Two-stage text spotting case-insensitive word accuracy are also reported, where C, P, T, and A are CRAFT, PAN, TRBA, and ASTER, respectively. Please note that TRAD, ZSL, UL, and SL stand for traditional methods, zero-shot learning, unsupervised learning, and supervised learning respectively. Scores in bold are the best of all.}
\label{table:full_syn_results}
\end{table*}

\section{Ablation Studies of Edge-Aware Attention and Text-Aware Copy-Paste Augmentation}
\label{sec:edge-aware-att-and-text-cap}
\begin{table*}[!ht]
\centering
\begin{adjustbox}{width=0.9\linewidth,center}
\begin{tabular}{lccccc}
\cmidrule[2pt]{1-6}
    \multirow{2}{*}{Model Variations} & \multicolumn{3}{c}{Image Quality} & \multicolumn{2}{c}{H-Mean} \\ \cmidrule(lr){2-4} \cmidrule(lr){5-6} 
    & PSNR $\uparrow$ & SSIM $\uparrow$ & LPIPS $\downarrow$ & CRAFT $\uparrow$ & PAN $\uparrow$ \\ \cmidrule[2pt]{1-6}
    Full Model (naive-PSA) & 23.768 & 0.725 & 0.774 & 0.354 & 0.291 \\
    Full Model (naive-Copy-Paste) & 22.822 & 0.714 & 0.787 & 0.356 & 0.284 \\
    Full Model & \textbf{25.596} & \textbf{0.751} & \textbf{0.751} & \textbf{0.368} & \textbf{0.298} \\ \cmidrule[2pt]{1-6}
\end{tabular}
\end{adjustbox}
\caption{Ablation study of our full model using different versions of PSA and Copy-Paste on SID-Sony-Text in terms of PSNR, SSIM, LPIPS, and text detection H-Mean. Scores in bold are the best of all.}
\label{table:det_ablation_study_cap_psa}
\end{table*}

In this experiment, we compared our design choices in the proposed framework against the original modules (i.e., original PSA vs. proposed Edge-Att and Copy-Paste vs. proposed Text-CP) while keeping other modules of our model unchanged. The quantitative results are reported in Table \ref{table:det_ablation_study_cap_psa}. In the original PSA module, the authors required the same feature maps to be fed into the individual channel and spatial attention submodules. In contrast, the proposed Edge-Att module takes in two heterogeneous inputs: low-light image feature maps for the channel attention part and edge feature maps for the spatial attention part. This allows the Edge-Att module to attend to rich high-level features from the low-light encoder while focusing on low-level features from the edge encoder. Such a design enables our full model to perform better than the naive-PSA model in terms of all metrics.

Furthermore, we studied the effectiveness of our proposed Text-CP compared to the original Copy-Paste augmentation. The results in Table \ref{table:det_ablation_study_cap_psa} show that our model trained with our proposed Text-CP scored better than the naive-Copy-Paste. By ensuring that texts are separated when pasted onto training image patches, our proposed method ensures that text detection models will not be confused by overlapping texts and can generate better text heatmaps for calculating text detection loss. This design also encourages our model to focus better on text instances since, without overlapping texts, text features will be clearer and more distinctive than other objects. 

\clearpage
\bibliographystyle{elsarticle-num} 
\bibliography{egbib}